\documentclass[10pt]{article}
\usepackage[utf8]{inputenc}
\usepackage{geometry}
\usepackage[dvipsnames]{xcolor}
\usepackage{amsfonts,amsmath,amssymb,float}
\usepackage{relsize}
\usepackage{fancyvrb}
\usepackage[colorlinks,linkcolor=MidnightBlue,citecolor=MidnightBlue,urlcolor=MidnightBlue,filecolor=black,backref=false]{hyperref}
\usepackage{geometry} 
\usepackage[defaultsans]{lato} 
\usepackage[T1]{fontenc}

\usepackage{graphicx}
\usepackage{natbib}
\usepackage{mathtools}
\usepackage{multirow}
\usepackage{bm}
\usepackage{algorithm}
\usepackage{algorithmic}
\usepackage{subcaption}
\usepackage{enumitem}
\usepackage{caption}
\usepackage{appendix}
\usepackage{setspace}

\makeatletter
\newcommand{\blu}[1]{{\color{black}{#1}}}
\newcommand{\R}{\mathbb{R}}
\newcommand{\E}{\mathbb{E}}
\newcommand{\V}{\mathbb{V}}
\newcommand{\Prob}{\mathbb{P}}
\DeclareMathOperator*{\argmax}{arg\,max}

\newcommand{\iidsim}{\overset{iid}{\sim}}
\newcommand{\indsim}{\overset{ind}{\sim}}
\newcommand{\eqd}{\overset{D}{=}}
\newcommand{\diag}{\text{diag}}
\newcommand{\Norm}{\text{N}} %Normal distribution
\newcommand{\Unif}{\text{Unif}} %Normal distribution
\newcommand{\Bern}{\text{Bern}} %Bernoulli distribution
\newcommand{\Gam}{\text{Gam}} %Gamma distribution
\newcommand{\GIG}{\text{GIG}} %GIG distribution
\newcommand{\PG}{\text{PG}} %Polya Gamma
\newcommand{\IG}{\text{IG}} %Inverse gamma distribution
\newcommand{\1}{\textbf{1}} %Indicator
\newcommand{\glob}{\text{glob}} %Indicator
\newcommand{\loc}{\text{loc}} %Indicator
\newcommand{\bW}{\mathbf{W}}
\newcommand{\bw}{\mathbf{w}}
\newcommand{\by}{\mathbf{y}}
\newcommand{\bx}{\mathbf{x}}

\newcommand{\bz}{\mathbf{z}}
\newcommand{\ba}{\mathbf{a}}

\newcommand{\bb}{\mathbf{b}}
\newcommand{\bD}{\mathbf{D}}
\newcommand{\bB}{\mathbf{B}}
\newcommand{\bS}{\mathbf{S}}
\newcommand{\bt}{\mathbf{t}}
\newcommand{\bM}{\mathbf{M}}
\newcommand{\bzero}{\mathbf{0}}
\newcommand{\bgamma}{\bm{\gamma}}
\newcommand{\bomega}{\bm{\omega}}
\newcommand{\btheta}{\bm{\theta}}
\newcommand{\btau}{\bm{\tau}}
\newcommand{\bpsi}{\bm{\psi}}
\newcommand{\bSigma}{\bm{\Sigma}}
\newcommand{\bmeta}{\bm{\eta}}
\newcommand{\Tau}{\mathrm{T}}

\DeclareMathOperator{\Tr}{Tr}
\DeclareMathOperator{\Var}{Var}

\DeclareTextFontCommand{\cmsspy}{\fontfamily{cmss}\selectfont}
\usepackage[noabbrev,capitalize,nameinlink]{cleveref}

\AtBeginDocument{%
  }

\allowdisplaybreaks

\begin{document}

\title{Variational Bayesian Bow tie Neural Networks with Shrinkage}


% \affiliation{%
%   \institution{University of Edinburgh}
%   \city{Edinburgh}
%   \country{UK}
% }
\title{Variational Bayesian Bow tie Neural Networks with Shrinkage}

\author{Alisa Sheinkman\thanks{School of Mathematics and Maxwell Institute for Mathematical Sciences, University of Edinburgh,  \href{mailto:a.sheinkman@sms.ed.ac.uk }{a.sheinkman@sms.ed.ac.uk}, \href{mailto:sara.wade@ed.ac.uk}{sara.wade@ed.ac.uk}} \, and \, Sara Wade$^{*}$}

% \thanks{School of Mathematics and Maxwell Institute for Mathematical Sciences, University of Edinburgh, \href{mailto:sara.wade@ed.ac.uk}{sara.wade@ed.ac.uk}}

\date{\today}

\maketitle


\begin{abstract}

Despite the dominant role of deep models in machine learning, limitations persist, including overconfident predictions, susceptibility to adversarial attacks, and underestimation of variability in predictions. The Bayesian paradigm provides a natural framework to overcome such issues and has become the gold standard for uncertainty estimation with deep models, also providing improved accuracy and a framework for tuning critical hyperparameters. However, exact Bayesian inference is challenging, typically involving variational algorithms that impose strong independence and distributional assumptions. Moreover, existing methods are sensitive to the architectural choice of the network. We address these issues by \blu{focusing on a stochastic relaxation of the standard feed-forward rectified neural network and using sparsity-promoting priors on the weights of the neural network for increased robustness to architectural design. Thanks to Polya-Gamma data augmentation tricks, which render a conditionally linear and Gaussian model, we derive a fast, approximate variational inference algorithm that avoids distributional assumptions and independence across layers. Suitable strategies to further improve scalability and account for multimodality are considered.}
%constructing a relaxed version of the standard feed-forward rectified neural network, and employing Polya-Gamma data augmentation tricks to render a conditionally linear and Gaussian model. Additionally, we use sparsity-promoting priors on the weights of the neural network for data-driven architectural design. To approximate the posterior, we derive a variational inference algorithm that avoids distributional assumptions and independence across layers and is a faster alternative to the usual Markov Chain Monte Carlo schemes.
\end{abstract}






\section{Introduction}
% While Deep Learning is known for its expressive and powerful models, these models are often overconfident and tend to overfit. The Bayesian approach to deep models is a natural framework to overcome such obstacles by creating ensembles of deep models: Bayesian neural networks provide improved uncertainty quantification with better calibration and robustness to gradient-based attacks.  We study a deep generative model generalizing rectified linear networks, where we employ a stochastic relaxation of the activation function and a Polya-Gamma data augmentation trick to render the model which is conditional linear and Gaussian. Additionally, we consider sparsity-inducing global-local Normal-generalized inverse Gaussians (N-GIG) priors on the weights of the network known to provide improvement in the prediction performance of Bayesian deep models.

% don't provide reliable uncertainty estimates, are often overconfident even when predictions are incorrect and are susceptible to adversarial attacks 

%

Neural networks (NNs) are effective deep models that play a dominant role in machine learning and have achieved remarkable success across various domains including medicine and biological sciences \citep{jumper2021highly, YU2021}, natural language processing \citep{mikolov2013, touvron2023llamaopenefficientfoundation}, computer vision and image analysis \citep{dosovitskiy2021imageworth16x16words}, data privacy and security \citep{yang2019} and beyond. 
However, modern machine learning applications often lack reliable, if any, uncertainty estimates \citep{guo17, Gal2016Uncertainty, ashukha2020pitfalls}. Classical deep models are easily fooled and are susceptible to adversarial attacks \citep{szegedy2013intriguing, nguyen2015deep, Foolllmandvision}, and even when the adversarial attacks fail,  
the saliency interpretations of deep neural networks (DNNs) are rather brittle \citep{carbone2021}. When data variations leading to out-of-distribution (OOD) shifts occur, neural networks often fail to generalize well \citep{hein2019relu,  zhang2024featurecontaminationneuralnetworks, ashukha2020pitfalls}. Moreover, standard neural networks usually lack intuitive interpretation and explainability
and so are regarded as black boxes \citep{lipton2018mythos}.
To address these challenges, Bayesian neural networks (BNNs) have emerged as a compelling extension of conventional neural networks (for a review, see e.g. \citep{jospin2022hands,arbel2023primer}). While finite (non-Bayesian) deep ensembles of independent neural networks have been shown to improve prediction and uncertainty estimates \citep{LakshminarayananDeepEns}, the Bayesian approach creates infinite ensembles of deep neural networks. 
The advantage of this approach is that it controls the model complexity and builds regularization into the model by marginalizing the parameters.
Indeed, BNNs have become the gold standard for uncertainty estimation in the context of data-driven decision-making and in safety-critical applications, where robustness and calibration are crucial \citep{mcallister2017concrete, carbone2020,gruver2023protein, yang2023sneakypromptjailbreakingtexttoimagegenerative, klarner2023}. 

A core problem of Bayesian machine learning lies in performing inference; in practice, the posterior distribution of the model's parameters given observations is not available in closed form, and direct sampling from the posterior is computationally expensive, meaning one has to employ approximate Bayesian inference. Markov chain Monte Carlo (MCMC) is a gold standard solution since it produces draws, which are asymptotically exact samples from the posterior, but for large data sets or complex models with multimodal posteriors, it can be prohibitively slow.  Variational inference \cite[VI][]{ Jordan1999,  blei2017variational} instead utilizes optimization rather than sampling making it a more computationally effective method suitable for high-dimensional, large-scale problems. VI approximates the posterior with the closest (most commonly, in terms of the Kullback–Leibler divergence) member of some tractable variational family of distributions taken as close as possible to the true posterior.

\blu{In this work, we focus on advancing approximate inference for BNNs. Specifically, we improve the bow tie neural network of \citep{smithbayesian} by introducing sparsity-inducing priors, constructing a fast, approximate variational inference algorithm, and  
exploring further strategies to improve computational gains and performance in terms of both accuracy and uncertainty estimation. Below, we review related research and provide an outline of the paper and its contributions.}

 \blu{\paragraph{Related work.}
Sparsity-inducing priors are known for their ability to improve robustness to overparametrization in Bayesian modeling;  
%the predictive performance of Bayesian deep models; 
they also lead to better calibrated uncertainty and, in certain settings, may recover the sparse structure of the target function \cite[e.g.][]{castillo2015bayesian,song2017nearly, song2020bayesian}. Such priors generally fall within two classes: 1) the two-group discrete mixture priors with a point mass at zero (referred to as spike-and-slab priors) \citep{george1993variable,mitchell1988bayesian} or 2) shrinkage priors, which employ a single distribution to approximate the spike-and-slab shape, yet are more computationally attractive, as they avoid exploring the space of all possible models. Both types of priors have become widely used in Bayesian deep modeling,  due to their high-dimensionality and overparametrization, and are further supported by theoretical guarantees (\citep{polson2018posterior,sun2022learning} for BNNs with spike-and-slab priors and \citep{castillo2024posterior, lee2022asymptotic} for BNNs with heavy-tailed shrinkage priors). 

To scale with the size of the data and model complexity, various variational algorithms and methods have been proposed for sparse BNNs \cite[for an overview, see][]{zhang2018advances}.  
Specifically, variational inference techniques for BNNs with spike-and-slab priors include \citep{blundell2015weight, bai2020efficient, bai2019adaptive, jantre2024spike}. Within the class of shrinkage priors, horseshoe priors \citep{carvalho09a, PiironenHorseshoe} on the BNN weights combined with variational approximations have been shown to provide competitive empirical results \citep{louizos2017, ghosh2018, ghosh2019}. 
Dropout regularization in neural networks has also been shown to be closely connected to sparse BNNs with suitable variational approximations \citep{KingmaVD2015, hron18a,louizos2017, liu2019variational,nalisnick2019dropout}. 
% Recently, \citep{castillo2024posterior} shows that a suitably rescaled heavy-tailed prior on the neural network weights achieves automatic adaptation, simultaneously to both the intrinsic dimension and smoothness of the underlying function, and near-optimal minimax contraction rates of the fractional posterior distribution and its mean-field variational approximation.
In a slightly different direction, \citep{li2024training} consider BNNs with Gaussian priors and introduce a variational inference methodology which enforces sparsity in the weights during the training. Finally, we mention the adaptive variational Bayes method of \citep{ohn2024adaptive}, which was successfully applied to Bayesian deep learning scenarios; the framework operates on a collection of models, considers "sieve" priors \citep{ArbelSieve2014} to combine several variational approximations.

\paragraph{Contributions and outline.}
\citep{smithbayesian} introduced a bow tie neural network, where a stochastic relaxation of the rectified linear unit (ReLU) activation function leads to a model amenable to the Polya-Gamma (PG) data augmentation trick \citep{polson2013bayesian} and results in conditionally linear and Gaussian stochastic activations. In this paper, we improve bow tie neural networks in several ways. First, to improve robustness with increasing width and depth of the networks, we place sparsity-inducing global-local normal-generalized inverse Gaussians (N-GIG) priors \citep{polson2010shrink}  on the weights of the network. \cref{sec:themodel} describes the bow tie model with shrinkage priors and implementation of PG data augmentation.  Second, while \citep{smithbayesian} focus on MCMC, in \cref{sec:inference} we propose a (block) structured mean-field family for the approximate variational posterior, which is flexible and doesn't require parametric assumptions on the distributional form of each component as well as on independence across layers. For the chosen family,  fast coordinate ascent variation inference (CAVI) \citep{bishop2016pattern} can be performed, with all variational updates available in the closed form.  Third, to improve the scalability of the algorithm, we consider two strategies: a stochastic variant that employs subsampling to cope with large data and a post-process node selection algorithm to obtain a sparse posterior that eases the storage and computational burden of predictions. 
%whilst continuous shrinkage priors result in more tractable computations than their spike-and-slab counterparts, they do not incur exact zeros on the neural network's weights. To obtain a sparse posterior that eases the storage and computational burden of predictions, we implement a simple post-process node selection algorithm controlled by the empirical Bayesian false discovery rate (FDR).  
Fourth, we propose improving accuracy and uncertainty estimation by considering ensembles of variational approximations obtained by running several parallel variational algorithms with different random starting points. In this way, our approach accounts for the multimodality of the posterior distributions arising in Bayesian deep models. 
\cref{sec:vb,sec:VIwithEM} derive the inference algorithm and in \cref{sec:svivbnn} a stochastic variant \citep{hoffman2013stochastic} of the algorithm is developed. Further, we derive a variable selection procedure in \cref{sec:selectvariables} for faster prediction (\cref{sec:prediction}) and propose ensembles in \cref{sec:ensembles}. We evaluate our method\footnote{We provide a Python implementation of our model on \href{https://github.com/sheinkmana/V_bowtie_NN}{GitHub}.} on a range of classical regression tasks as well as synthetic regression tasks and demonstrate its competitiveness compared to alternative, well-known Bayesian algorithms in  \cref{sec:experiments}.}

\section{Bayesian Augmented Bow Tie Neural Network with Shrinkage} \label{sec:themodel}
\subsection{Bow tie Neural Networks}\label{sec:bowtie}
 We begin by describing the class of recently proposed \textit{bow tie networks} \citep{smithbayesian}, which are deep generative models that generalize feed-forward rectified linear neural networks with stochastic activations. Let $\bx_n \in \R^{D_0}$ be the inputs, $\by_n \in \R^{D_{L+1}}$ be the outputs and $\ba_n = \{\ba_{n,l}\}_{l=1}^{L}$ with $\ba_{n,l} \in \R^{D_l}$ be the latent activations at each of the $L$ intermediate layers. 
For notational purposes, assume $\ba_{n,0} = \bx_{n}$.
The model assumes:
\begin{align*}
\by_n \mid \ba_n, \bx_n, \btheta  \sim \Norm \left(\by_n | \bz_{n,L+1}, \bSigma_{L+1} \right) \quad \text{ for } n=1, \ldots, N,
\end{align*}
where 
\begin{align}
&\ba_{n,l} |\bz_{n,l}, \btheta \sim \Norm \left( f(\bz_{n,l}), \bSigma_{l} \right), \quad \text{with} \quad \bz_{n,l} =  \bW_l \ba_{n,l-1} + \bb_l \quad   \text{for} \quad l=1,\ldots, L+1.
\label{eq:bnn_stochastic}
\end{align}
% \begin{align*}
% \bz_{n,l} &= \left\lbrace\begin{array}{cc} \bW_l \ba_{n,l-1} + \bb_l & \text{ for } l=2,\ldots, L+1, \\
% \bW_1 \bx_{n} + \bb_1 & \text{ for } l=1,
% \end{array} \right.\\
% \ba_{n,l} |\bz_{n,l}, \btheta &\sim \Norm \left( f(\bz_{n,l}), \bSigma_{l} \right) \quad \text{ for } l=1, \ldots, L,
% \end{align*}
Here $f(\bz)$ is a nonlinear activation function applied elementwise and the parameters  $\btheta = (\bW_l, \bb_l, \bSigma_l)_{l=1}^{L+1}$ consist of the weights $\bW_l \in \R^{D_l \times D_{l-1}}$, biases $\bb_l \in \R^{D_l}$ and covariance matrices $\bSigma_l \in \R^{D_l \times D_l}$. %Denote $\btheta = (\bW_l, \bb_l, \bSigma_l)_{l=1}^{L+1}$.

Note that \cref{eq:bnn_stochastic} is a stochastic relaxation of the standard feed-forward NN, which is recovered in the limiting case when $\bSigma_l \rightarrow \bzero $ for $l=1,\ldots, L$. Instead of relying on local gradient-based algorithms, \citep{smithbayesian} introduces another relaxation of the model and employs a \textit{Polya-Gamma data augmentation trick} \citep{polson2013bayesian} to render the model conditionally linear with Gaussian activations. Specifically, consider the ReLU activation function $f(z) = \max(0,z)$. It can alternatively be written as
a product of $z$ and a binary function $\gamma$, i.e.  
$f(z) = \gamma z$ where $\gamma = \1(z>0)$.
In this way, $\gamma$  determines whether the node is activated ($\gamma =1$) or not ($\gamma = 0$).
%In a similar fashion, the additional stochastic relaxation replaces $f(\bz_{n,l})$ with $\bgamma_{n,l} \odot \bz_{n,l}$, where $\odot$ represents the elementwise product. 
\blu{The additional stochastic relaxation assumes that $\gamma$ is not deterministic, but a binary random variable whose success probability depends on $z$. Specifically, the dependence is constructed through the logistic function $\sigma(x) = \exp(x)/(1+\exp(x)) $ and a temperature parameter $\Tau \geq 0$:}
\begin{align*}
\ba_{n,l} |\bz_{n,l}, \bgamma_{n,l},\btheta &\sim \Norm \left( \bgamma_{n,l} \odot \bz_{n,l} , \bSigma_{l} \right),\\
\gamma_{n,l,d} &\indsim Bern\left(\sigma\left( z_{n,l,d}/\Tau\right) \right) \text{ for } T>0,
% \begin{cases}
%\Bern\left(\sigma\left( z_{n,l,d}/\Tau\right) \right) \text{ for } T>0,\\
%\Bern\left(\text{ReLU}\left( z_{n,l,d}\right) \right) \text{ for } T=0,\\
%\end{cases}
\end{align*}
where $\odot$ represents the elementwise product. 
Thus, the nodes are turned off or on stochastically depending on their input. \blu{Note that in the limit as the temperature $T \rightarrow 0$, we have that $\gamma_{n,l,d} =  \1(z_{n,l,d}>0)$ and $\ba_{n,l} |\bz_{n,l}, \bgamma_{n,l},\btheta \sim \Norm \left( \1(\bz_{n,l}>0) \odot \bz_{n,l} , \bSigma_{l} \right)$.}

\begin{figure}[!t]
  \centering  
  \includegraphics[height=5cm]{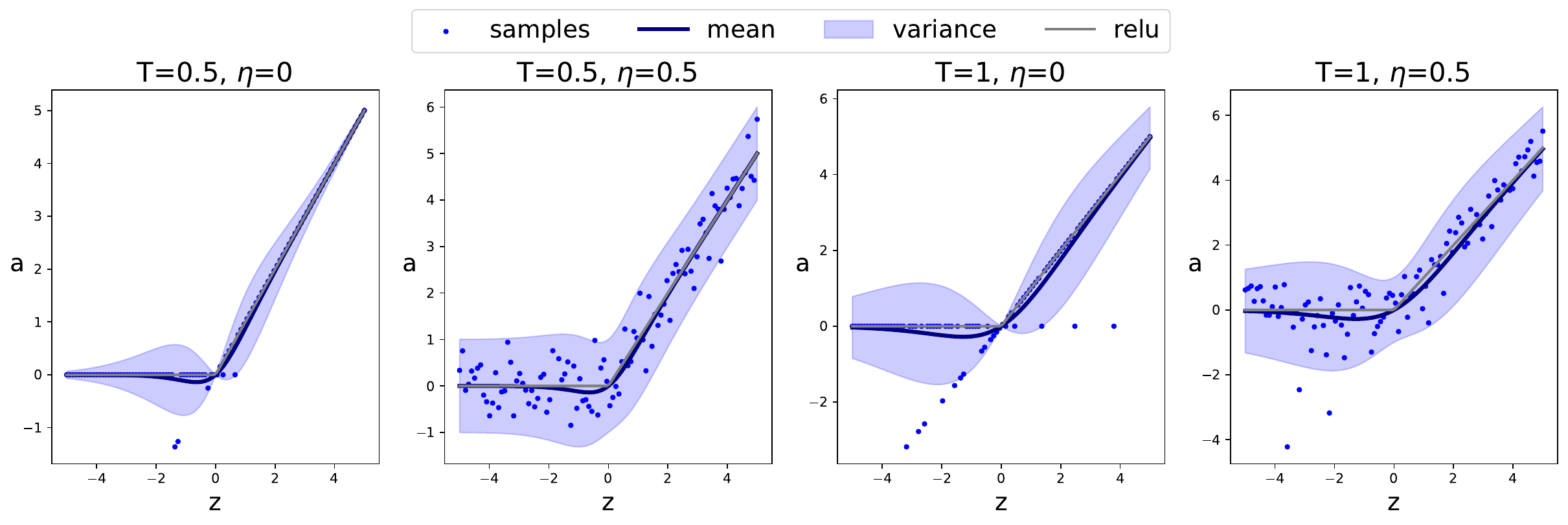}
        \caption{Conditional distribution of $a$ given the input $z$  for various settings of the temperature $T$ and noise $\eta$, with the conditional mean in \cref{eq:cmean} (solid line), conditional variance in \cref{eq:cvar} (shaded region) and samples from the conditional distribution in \cref{eq:amarginal} (points).}
          \label{fig:bowtie}
\end{figure}

For $T>0$, after marginalizing the binary activations, the stochastic activations $\ba_n$ are distributed as a mixture of two normals:
\begin{align}
\blu{\mathrm{a}_{n,l,d}} |z_{n,l,d},\btheta &\sim \sigma(z_{n,l,d}/\Tau)\Norm \left( z_{n,l,d} , \eta_{l,d}^2 \right) + \left(1-\sigma(z_{n,l,d}/\Tau)\right)\Norm \left( 0 , \eta_{l,d}^2 \right) ,
\label{eq:amarginal}
\end{align}
where the variance $\eta_{l,d}^2$ is the $(d,d)$th element of $\bSigma_l$, and 
\begin{align}
\E[\blu{\mathrm{a}_{n,l,d}} |z_{n,l,d},\btheta] &= \E[\E[\blu{\mathrm{a}_{n,l,d}} | z_{n,l,d},\gamma_{n,l,d},\btheta]]= \E[ \gamma_{n,l,d}  z_{n,l,d} ] \nonumber \\
&= \sigma(z_{n,l,d}/\Tau)z_{n,l,d}, \label{eq:cmean}
\end{align}
\begin{align}
\V(\blu{\mathrm{a}_{n,l,d}} |z_{n,l,d},\btheta) &= \E[ \V(\blu{\mathrm{a}_{n,l,d}} | z_{n,l,d},\gamma_{n,l,d},\btheta)] + \V(\E[\blu{\mathrm{a}_{n,l,d}} | z_{n,l,d},\gamma_{n,l,d},\btheta]) \nonumber\\
&= \E[\eta_{l,d}^2] + \V( \gamma_{n,l,d}  z_{n,l,d}) \nonumber\\
&= \eta_{l,d}^2 + z_{n,l,d}^2\sigma(z_{n,l,d}/\Tau)\left(1-\sigma(z_{n,l,d}/\Tau) \right). \label{eq:cvar}
\end{align}
The conditional distribution of $\blu{\mathrm{a}_{n,l,d}}$ is displayed in \cref{fig:bowtie}, for different combinations of the temperature parameter $T$ and variance $\eta_{l,d}^2$. The ReLU activation is recovered in the case of $T=0$ and  $\eta_{l,d}^2 = 0$, while other choices of $T$ and  $\eta_{l,d}^2$ generalize the ReLU. \blu{The stochastic activations have smoother conditional mean functions for non-zero values of $T$, with potential bimodality (density resembles a bow tie, hence the name) for larger $T$ relative to $\eta$. This may be relevant in settings with multimodal predictive distributions, although our experiments focus on small temperature values for more similarity to the ReLU.} 
%The density resembles a bow tie, hence the name of the model. 
%\AS{Reviewer 3 asked 'The authors should motivate the role of the temperature and discuss
%how it leads to 0,1 variables." 0,1 variables of what? I don't know what to say. What can say: as $T$ approach $0$ from the above $\mathrm{a}_{n,l,d} \to \Norm \left( z_{n,l,d} , \eta_{l,d}^2 \right)$; as $T$ approaches $0$ from below $\mathrm{a}_{n,l,d} \to \Norm \left( z_{n,l,d} , \eta_{l,d}^2 \right)$, As the temperature $T$ increases, the variance of  $\mathrm{a}_{n,l,d}$ and the overall difference between the mean of $\mathrm{a}_{n,l,d}$ and the original ReLU function increases.}
% \begin{figure}[!t]
%   \centering  
%   \includegraphics[height=3cm]{pictures/figure_1cut.pdf}
%   \includegraphics[height=3cm]{pictures/figure_1cutone.pdf}
%         \caption{Conditional distribution of $a$ given the input $z$  for various settings of the temperature $T$ and noise $\eta$, with the conditional mean in \cref{eq:cmean} (solid line), conditional variance in \cref{eq:cvar} (shaded region) and samples from the conditional distribution in \cref{eq:amarginal} (points).}
%         \Description{For zero values of temperature and noise the plot of the conditional distribution of stochastic activation coincides with the plot of ReLU function. For nonzero values of the temperature parameter and noise, the conditional distribution looks like a bow tie. This explains the name of the model}
%           \label{fig:bowtie}
% \end{figure}

\subsection{Shrinkage Priors} \label{sec:shrinkagepriors}

Prior elicitation in Bayesian neural networks is challenging, as understanding how the high-dimensional weights map to the functions implemented by the network is not trivial. Standard Gaussian priors are often a default choice, also due to their link with $\ell_2$ regularization in maximum a posteriori (MAP) inference; indeed, such priors were used in \citep{smithbayesian}. For an overview and discussion on priors in Bayesian neural networks, see \citep{fortuin2022priors}. 

 We take an alternative approach to the Gaussian priors of \citep{smithbayesian} in order to sparsify our model. Sparsity-inducing priors %ease the problem of storage and computational costs, 
 have been shown to provide improvement in the predictive performance of deep models and can provide a data-driven approach to selecting the width and depth, easing the difficult task of specifying the network architecture. 
In this work, we focus on a class of continuous shrinkage priors, namely, global-local normal scale-mixtures with generalized inverse Gaussian shrinkage priors on the scale parameters, referred to as \textit{global-local normal-generalized inverse Gaussian priors} \citep{griffin2021bayesian}. Global-local scale-mixtures aim to shrink less important weights whilst leaving large ones, which is achieved through a global parameter controlling the overall shrinkage, with the local parameters allowing deviations at the level of individual nodes  %that local priors should have heavy tails and global priors should have substantial mass near zero 
\citep{polson2010shrink,bhadra2019lasso}. 
This choice of priors is also motivated by the theoretical guarantees for high-dimensional regression
\citep{song2017nearly, griffinbrownInfnormgamma, polson2013bayesian}, for a survey on global-local shrinkage methods we refer to \citep{griffin2021bayesian}. 

 The N-GIG priors on the weights have the following hierarchical structure:
 \begin{align}
\Prob(\bW_l | \bpsi_{l} , \tau_l) &= \prod_{d=1}^{D_l} \prod_{d'=1}^{D_{l-1}}\Norm\left(W_{l,d,d'} |  0, \tau_l \psi_{l,d,d'} \right), \label{eq:normalGIG} \\
\Prob(\bpsi_{l} ) &=\prod_{d=1}^{D_l} \prod_{d'=1}^{D_{l-1}} \GIG\left( \psi_{l,d,d'} \mid \nu_{\loc, l}, \delta_{\loc, l}, \lambda_{\loc, l} \right), \label{eq:lGIG}\\
\Prob(\tau_{l} ) &=\GIG\left( \tau_l \mid \nu_{\glob}, \delta_{\glob}, \lambda_{\glob} \right), \label{eq:glGIG}
 \end{align}
 where $\tau_{l}$ is the global shrinkage parameter for layer $l$ and  $\psi_{l,d,d'}$ is the local shrinkage parameter for each weight. The generalized inverse Gaussian (GIG) prior is given by:
 \begin{align*}
 \GIG\left( \psi \mid \nu, \delta, \lambda \right) \propto \psi^{\nu-1} \exp \left( -\frac{1}{2} (\delta^2/\psi + \lambda^2\psi) \right),
 \end{align*}
with parameters $\nu$, $\delta$, and $\lambda$; for a proper prior,  $\nu>0$ if $\delta= 0$ or $\nu<0$ if $\lambda=0$. In \cref{eq:lGIG}, we allow the GIG parameters for the local scale parameters $\psi_{l,d,d'}$ to vary across layers to adjust local shrinkage for wider layers. 
Furthermore, to encourage more shrinkage for larger depth and width, we scale the global parameters $\tau_{l}$ with respect to $L$ and the local parameters $\psi_{l,d,d'}$ with respect to $D_{l}$ (details of our approach are provided \cref{sec:initschemes}). 

\begin{figure}[!t]
     \centering
     \subcaptionbox{Marginal prior for the weights.\label{plot}}[.36\textwidth]{  \includegraphics[width = \linewidth]{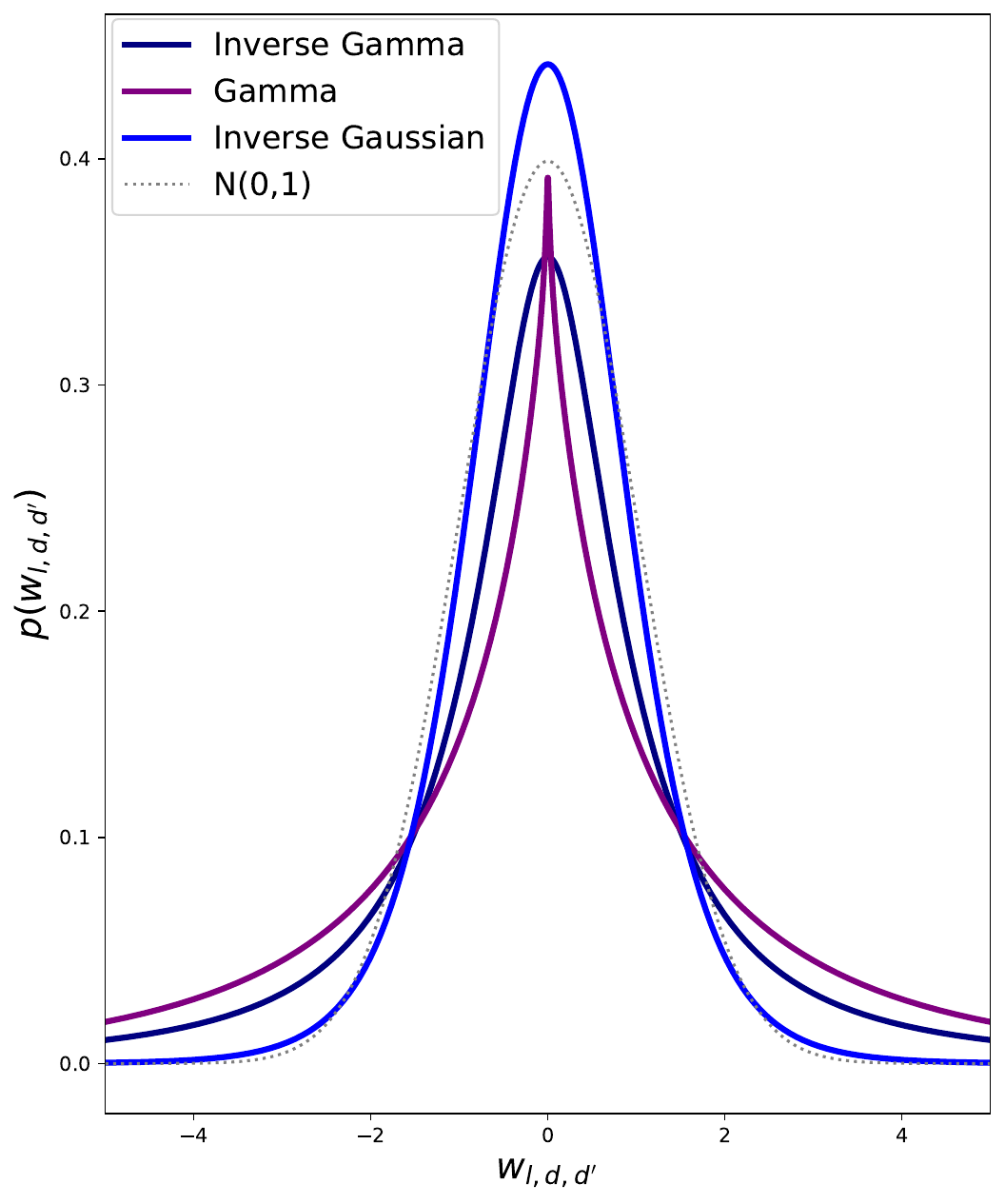}}
      \subcaptionbox{Joint prior for the weights. \label{plot2}}[.63\textwidth]{ \includegraphics[scale =0.27]{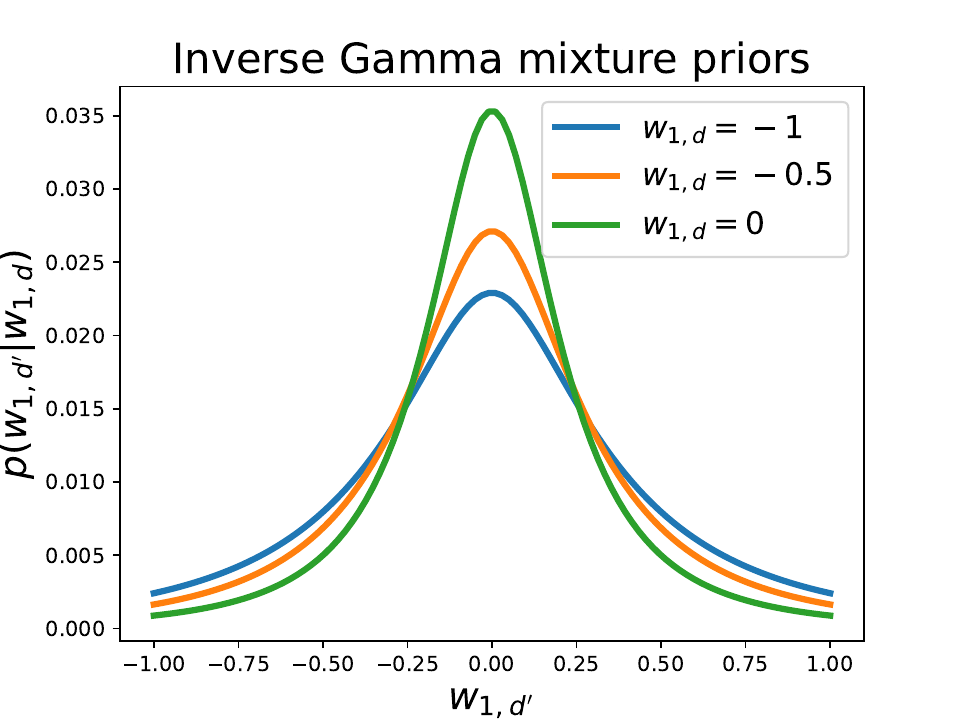}     \includegraphics[scale =0.27]{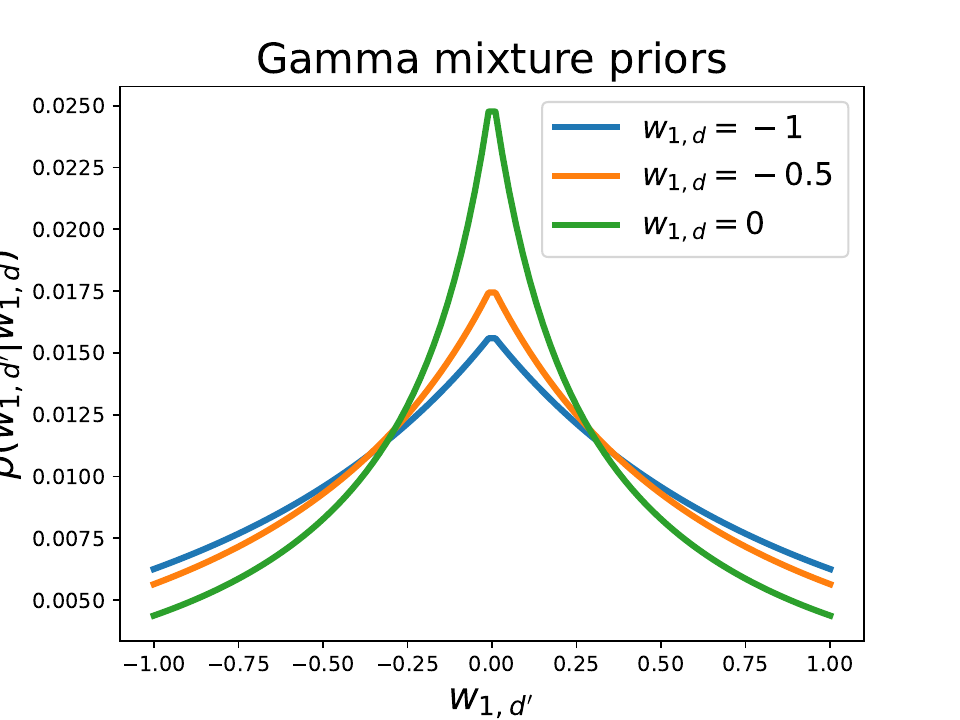} \medskip
     \includegraphics[scale = 0.29]{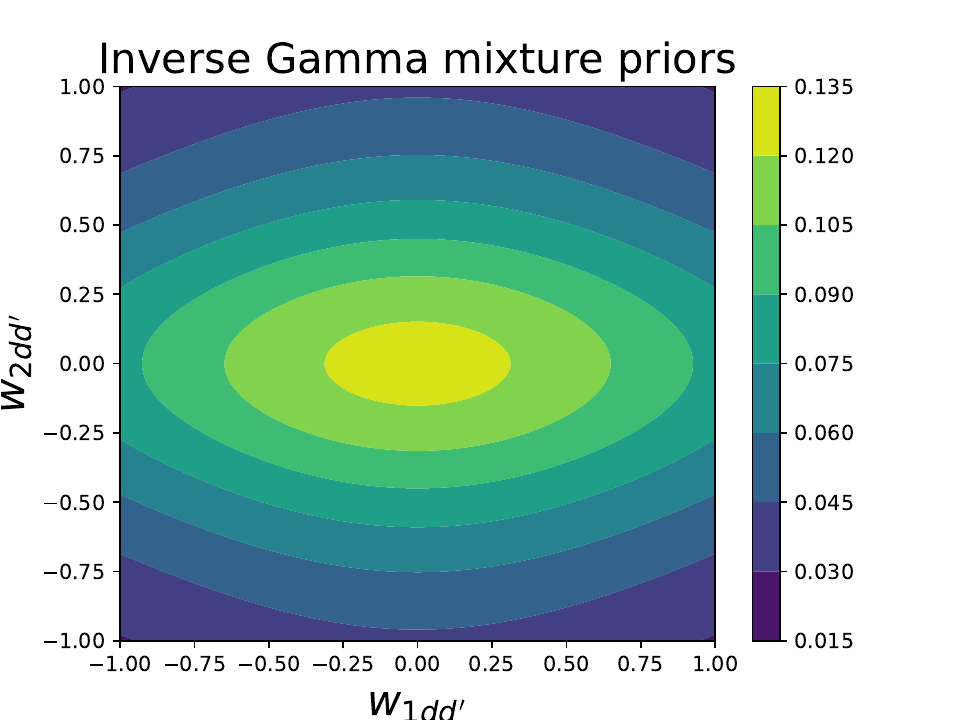} 
     \includegraphics[scale = 0.29]{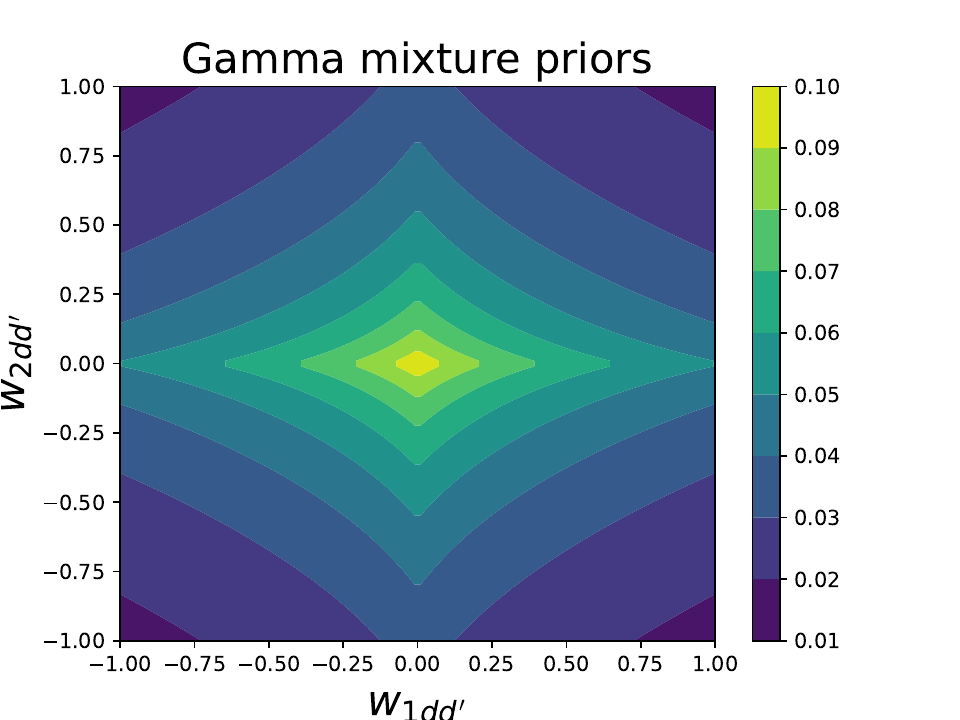}}
     \caption{
     %Density estimates for the probability distribution of weights in different cases of global-local mixture priors. 
     Illustration of the prior on the weights. (a) the marginal density of the weights for different choices within the GIG family. (b) the \blu{conditional prior of the weights within the same layer (top) and joint prior of the weights across layers (bottom)} for two choices of IG (left) and Gamma (right) mixing priors.}\label{fig:priors} 
   \end{figure}

When the global shrinkage parameter $\tau_l$ is fixed, examples of the marginal distribution for $w_{l,d,d'}$ include Laplace \citep{parkcasellalasso}, Student-t (ST) \citep{TippingSparse}, Normal-Gamma (NG) \citep{caronsparse, griffinbrownInfnormgamma}, Normal inverse Gaussian (NIG) \citep{caronsparse}. Each example has a different tail behavior, inducing different forms of shrinkage (see \cref{tab:examples} for an overview and \cref{fig:priors} (a)  for a visualization). Note that if the prior is polynomial-tailed,  then for large signals the amount of shrinkage is mitigated even given small $\tau_l$ \citep{polson2010shrink}.
\begin{table}[!b]
\begin{center}
\caption{Examples with the class of N-GIG priors.}
\label{tab:examples}
\resizebox{0.99\linewidth}{!}{
\begin{tabular}{| p{2.5cm}|p{3cm}|p{3cm}|p{3cm}|p{3.5cm}|}
 \hline
 \multicolumn{5}{|c|}{Marginal for $w_{l,d,d'}$ when $\tau_l$ is fixed} \\
 \hline
  & Student-T & Laplace & NG & NIG  \\
 \hline
 Mixing distribution & Inverse Gamma (IG) & Gamma  & Gamma & Inverse Gaussian (IGauss) \\
Parameters  &   $\nu<0,\delta>0,\lambda =0$ & $\nu =1,\delta=0,\lambda$ & $\nu,\delta=0,\lambda$  & $\nu = \frac{1}{2},\delta,\lambda$ \\
Tail behavior & polynomial-tailed & exponential-tailed &  exponential-tailed & exponential-tailed \\
 \hline
\end{tabular}}
\end{center}
\end{table}
 The global shrinkage parameter $\tau_l$ leads to a non-separable penalty for the weights within the same layer, i.e. after integrating out $\tau_l$, the weights within the same layer are dependent.  %\cref{fig:priors} (a) illustrates the marginal density of the weights for different choices of mixing priors within the GIG family. 
 This is illustrated in \blu{the top row of \cref{fig:priors} (b), which shows how the conditional prior density of one weight varies given different values of another weight within the same layer, for two choices of inverse Gamma (IG) and Gamma mixing priors. Instead,  across layers the weights are independent, as illustrated by the contour plots for the joint density in the bottom row of  \cref{fig:priors} (b). The bottom row of  }\cref{fig:priors} (b) also highlights how the variance depends on the width of the layer, with more hidden units and smaller variance for the second layer compared to the first. 
 %Specifically, for IG priors with $\lambda = 0 $ and $\nu = -1.5$ (equivalently $df = 3$) we consider $\delta_{\glob}= 1, \delta_{loc, 1} = 1, \delta_{\loc, 2} = 2$  and for plots obtained with Gamma priors ($\delta = 0, \nu = 1$) we use $\lambda_{\glob } = \sqrt{0.5}, \lambda_{loc, 1} = \sqrt{0.5},  \lambda_{loc, 2} = 0.5$. 

The opposite effects of varying width and depth in deep neural networks are studied in \citep{vladimirova2021}; while depth accentuates a model’s non-Gaussianity, the width makes models increasingly Gaussian. 
Indeed, infinitely wide BNNs are closely related to Gaussian processes (GPs), typically relying on appropriately scaled i.i.d. Gaussian weights \citep{neal,lee18,matthews} and relaxing these assumptions, e.g. through ordering, constraints, heavy tails, or bottlenecks, results in non-Gaussian limits, such as stable processes \citep{peluchetti2020}, deep GPs \citep{agrawal2020wide}
or more exotic processes \citep{sell2020dimension,chada2022multilevel}. The sparsity-promoting priors in \cref{eq:normalGIG,eq:lGIG,eq:glGIG} provide a framework for the data to inform on the width and depth of the network.

\subsection{Polya-Gamma Data Augmentation}

As in \citep{smithbayesian}, we employ \textit{Polya-Gamma data augmentation} \citep{polson2013bayesian} to render the model conditionally linear and Gaussian. 
First, recall the definition of the Polya-Gamma distribution with parameters $b>0$ and $c \in \mathbb{R}$, denoted $\PG(b,c)$. The random variable $X \sim \PG(b,c)$ if 
\begin{align*}
    X \eqd \frac{1}{2\pi^2} \sum_{k=1}^\infty \frac{g_k}{(k-1/2)^2 + c^2/4\pi^2}, \quad \text {where } g_k \iidsim \Gam(b,1).
\end{align*}
The key identity that we use is:
\begin{align}
    \frac{\exp(z)^a}{(1+\exp(z))^b} = 2^{-b} \exp(\kappa z) \int_0^\infty \exp(-\frac{\omega z^2}{2} ) p(\omega) d\omega, \label{eq:pgidentity}
\end{align}
where $\kappa = a-b/2$ and $p(\omega) = PG(\omega| b,0)$. The integral is a Gaussian kernel, thus if $z=\bw^T\bx$, conditioned on the latent variable $\omega$, $\bw$ has a Gaussian distribution and conditioned on $\bw$, $\omega$ has a PG distribution. While to sample from the PG distribution, one can use the alternating series method of \citep{devroye2006nonuniform}, all finite moments of the PG random variables are available in closed form, and that becomes useful for expectation-maximization or variational Bayes algorithms. Specifically, for $c>0$
\begin{align}
    \E[\omega] = \frac{b}{2c} \frac{\exp(c)-1}{1+\exp(c)}. \label{eq:pgmean}
\end{align}
Moreover, the PG distribution is closed under convolution with the same scale parameter; if $\omega_1 \sim \PG(b_1,c)$ and $\omega_2 \sim \PG(b_2,c)$, then $\omega_1 + \omega_2 \sim \PG(b_1+b_2,c)$.  

\subsection{Augmented Model} \label{sec:modelposterior}

The model  in \cref{sec:bowtie} augmented with stochastic activations $\ba =(\ba_{n,l})$ and  binary activations $\bgamma = (\bgamma_{n,l})$ is:
\begin{align*}
    p(\by, \ba, \bgamma |\btheta) &= \prod_{n=1}^N \Norm\left(\by_n \mid \bz_{n,L+1}, \bSigma_{L+1}\right) \prod_{n=1}^N \prod_{l=1}^L \Norm \left(\ba_{n,l} \mid \bgamma_{n,l} \odot \bz_{n,l} , \bSigma_{l} \right)\\ & \times \prod_{d=1}^{D_l}\frac{\exp(z_{n,l,d}/T )^{\gamma_{n,l,d}}}{1+\exp(z_{n,l,d}/T )}.
\end{align*}
Then using the \cref{eq:pgidentity}, the last term can be written as:
\begin{align*}
  \frac{\exp(z_{n,l,d}/T )^{\gamma_{n,l,d}}}{1+\exp(z_{n,l,d}/T)} =  2^{-1} \exp \left (\frac{\kappa_{n,l,d} z_{n,l,d}}{T}\right ) \int_0^\infty \exp \left (-\frac{\omega_{n,l,d}z_{n,l,d}^2}{2T^2} \right ) p(\omega_{n,l,d}) d\omega_{n,l,d}, 
\end{align*}
where $\omega_{n,l,d} \sim \PG(1,0)$ and $\kappa_{n,l,d} = \gamma_{n,l,d} - 1/2$. 
Thus, introducing the additional augmented variables $\bomega = (\omega_{n,l,d} )$, we arrive at the augmented model:
\begin{align*}
    p(\by, \ba, \bgamma,\bomega |\btheta) &\propto \prod_{n=1}^N \Norm\left(\by_n \mid \bz_{n,L+1}, \bSigma_{L+1}\right) \prod_{n=1}^N \prod_{l=1}^L \Norm \left( \ba_{n,l} \mid \bgamma_{n,l} \odot \bz_{n,l} , \bSigma_{l} \right) \\
    & \times\prod_{d=1}^{D_l} \exp \left (\frac{\kappa_{n,l,d} z_{n,l,d}}{T}\right ) \exp \left (-\frac{\omega_{n,l,d}z_{n,l,d}^2}{2T^2} \right ) p(\omega_{n,l,d}).
\end{align*}

The covariance matrices are assumed to be diagonal $\bSigma_l = \diag(\eta_{l,1}^2,\ldots \eta_{l,D_l}^2)$, with  variances  denoted by $\bmeta_l=(\eta_{l,1}^2,\ldots \eta_{l,D_l}^2)$. Additionally, we assume conjugate priors for the variances $\eta_{l,d}^2 \iidsim \IG(\alpha^h_0,\beta^h_0)$ for $l=1, \ldots, L$ and $\eta_{L+1,d}^2 \iidsim \IG(\alpha_0,\beta_0)$ and for the biases $b_{l,d} \iidsim \Norm(0, s^2_0)$. Here, we consider different prior parameters $\alpha_0^h, \beta_0^h$ for the variance terms associated to the hidden layers in comparison to the parameters $\alpha_0, \beta_0$ for the final layer. In particular,  $\alpha_0, \beta_0$ are chosen to reflect prior knowledge in the noise, while $\alpha_0^h, \beta_0^h$ are chosen so that the prior concentrates on small values and realizations of the stochastic activation function are more similar to the ReLU.  

\begin{figure}[!t]
\centering 
\includegraphics[width=0.8\linewidth]{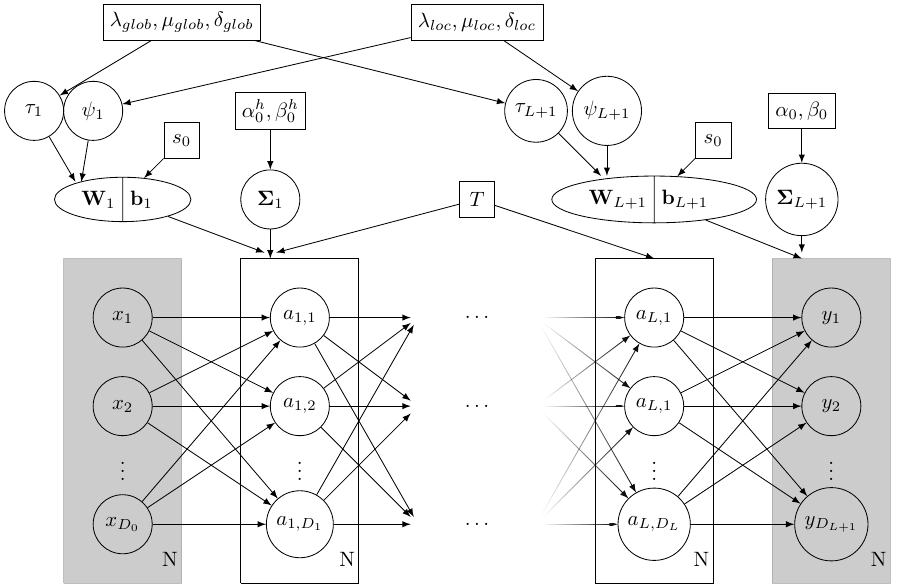}
\caption{Directed Acyclic Graph of the model.}
\label{fig:DAG}
\end{figure}

A graphical model of the bow tie BNN with stochastic relaxation and shrinkage priors  is displayed in \cref{fig:DAG}, and the posterior distribution over both the model parameters and latent variables is:
\begin{align*}
    p(\ba, \bgamma,\bomega,\bW, \bb, \bmeta,\bpsi,\btau) &\propto\prod_{n=1}^N \Norm\left(\by_n \mid \bz_{n,L+1}, \bSigma_{L+1}\right) \prod_{n=1}^N \prod_{l=1}^L \Norm \left( \ba_{n,l} \mid \bgamma_{n,l} \odot \bz_{n,l} , \bSigma_{l} \right) \\
    & \times\prod_{d=1}^{D_l} \exp \left (\frac{\kappa_{n,l,d} z_{n,l,d}}{T}\right ) \exp \left (-\frac{\omega_{n,l,d}z_{n,l,d}^2}{2T^2} \right ) p(\omega_{n,l,d}) \\  
    & \times  \prod_{d=1}^{D_{L+1}}\IG(\eta^2_{L+1,d} \mid \alpha_0, \beta_0) \times \prod_{l=1}^L  \prod_{d=1}^{D_l} \IG( \eta^2_{l,d} \mid \alpha_0^h, \beta_0^h) \times \prod_{n=1}^N \prod_{l=1}^L \prod_{d=1}^{D_l} \Bern \left(\gamma_{n,l,d} \mid \sigma \left (\frac{z_{n,l,d}}{T} \right ) \right) \\
    & \times  \prod_{l=1}^L \left( \prod_{d=1}^{D_l} \left( \Norm(b_{l,d} \mid 0, s_0^2) \times \prod_{d'=1}^{D_{l-1}}\Norm\left(W_{l,d,d'} |  0, \tau_l \psi_{l,d,d'} \right)\right)\right) \\
& \times \prod_{l=1}^L \left( \GIG\left( \tau_l \mid \nu_{\glob}, \delta_{\glob}, \lambda_{\glob} \right) \prod_{d=1}^{D_l} \prod_{d'=1}^{D_{l-1}} \GIG\left( \psi_{l,d,d'} \mid \nu_{\loc,l}, \delta_{\loc,l}, \lambda_{\loc,l} \right) \right).
\end{align*}

\section{Inference} \label{sec:inference}
\subsection{Variational Bayes}\label{sec:vb}

While Markov chain Monte Carlo is considered the gold-standard tool for approximating posterior distributions in Bayesian modeling due to its asymptotic guarantees, MCMC algorithms can be prohibitively slow when the model dimension and sample size are large. Instead, we focus on variational inference, an alternative fast, approximate Bayesian inference method that has gained popularity in the literature \citep{Ormerod2010, zhang2018advances}, due to both the explosion in the amount of data collected and the use of highly parametrized models for increased flexibility. 
VI has been shown to yield reasonably accurate approximations in several problems as well as desirable frequentist properties. Namely, \blu{consistency and asymptotic normality of the variational posterior expectation are established in \citep{wang2019frequentist}}, and theoretical guarantees for optimal contraction rates of variational posteriors under certain assumptions appear in several recent works \citep{zhang2020convergence,alquier2020concentration, bhattacharya2023convergence, yang2020alpha, szaborayvbforsparse}. %\blu{In the sparse deep learning settings, several variational algorithms were developed for Bayesian neural networks with spike-and-slab priors \citep{blundell2015weight, bai2020efficient}, horseshoe \citep{ghosh2018, ghosh2019} and in the context of dropout \citep{louizos2017,liu2019variational, nalisnick2019dropout, dropoutGal}. Convergence rates of variational approximations arising on BNNs with spike-and-slab priors are studied in \citep{sun2022learning,cherief2020convergence,bai2019adaptive,bai2020efficient}; consistency of variational Bayes for BNNs with horseshoe priors is considered in \citep{jantre2024spike} and for neural networks with heavy-tails in \citep{lee2022asymptotic,castillo2024posterior}. 
\blu{We refer readers interested in the caveats of VI to \citep{gelmanyesbutdidit,barber2011bayesian,hron18a}.}

Consider fitting a model parameterized by $\btheta$ to the observed data $\mathcal{D}$. In variational inference, the true posterior $p(\btheta \mid \mathcal{D})$ is approximated by a density $q(\btheta)$ taken from a family of distributions $\mathcal{F}$ that minimizes the Kullback-Leibler divergence between the approximate and true posterior, or equivalently maximizes the \textit{evidence lower bound} (ELBO)
		\begin{equation}
		\label{eqn:ELBO}
		\text{ELBO} = \E_{q(\btheta) } \left [ \log \left( \frac{p(\btheta,  \mathcal{D})  }{q(\btheta)}  \right)\right ].
\end{equation}

A common choice for $\mathcal{F}$ is the nonparametric, mean-field family on a partition $\{ \btheta_1, \ldots, \btheta_J \}$ of $\btheta$, assuming that the variational posterior factorizes over (blocks) of latent variables:
	$
	q(\btheta) = \prod_{j=1}^J q_j(\btheta_j),
	$
	where $J \le \text{dim} (\btheta)$. Without any further parametric assumptions, it has been shown \citep{hinton1993keeping,mackay1995probable,Jordan1999} that the optimal choice for each product component $q_j$ is 
	\begin{equation}
	\label{eqn::CAVIupdate}
	q_j (\btheta_j) \propto \exp \left [ \E_{- \btheta_j} \log \left(  p(\btheta,  \mathcal{D}) \right)  \right ],
	\end{equation}
	where the above expectation is taken with respect to $\prod_{j^\prime \neq j} q_{j^\prime} (\btheta_{j^\prime})$. %This choice of product component is
	The process of sweeping through the components of the 
	partition and updating one at a time via \cref{eqn::CAVIupdate} is known as \textit{coordinate ascent variational inference} (CAVI). \citep{wand2011mean} studies CAVI's performance and improvement techniques in the case of elaborate distributions. Limitations of the mean-field assumption can be found in \citep{wand2011mean, Neville_2014, coker2022wide}.

 We specify the mean-field family for the approximate variational posterior:
\begin{align}
    q(\ba, \bgamma,\bomega,\bW, \bb, \bmeta,\bpsi,\btau) = q(\ba) q(\bgamma) q(\bomega)q(\bW,\bb)q(\bmeta)q(\bpsi)q(\btau).\label{eq:vbnnmfamily}
\end{align}
Note that the assumption on the family above could be referred to as the \blu{\textit{structured mean-field assumption}}. Importantly, unlike existing variational algorithms for BNNs, we do not make any assumptions on independence of parameters between layers.  
The calculation of each component of the variational posterior is given in the  \cref{appendix:vb}, where using \cref{eqn::CAVIupdate} we obtain the following update steps.

\paragraph{Global shrinkage parameters:} the parameters $\btau$ are independent across layers (and can be updated in parallel) with a GIG variational posterior:
\begin{align}
     q(\btau) &= \prod_{l}^{L+1} \GIG \left (\tau_l \; \mid \; \hat{\nu}_{\glob,l}, \hat{\delta}_{\glob,l}, \lambda_{\glob} \right ),\label{eq:vbtau} 
\end{align}
where for $l=1,\ldots,L+1$,
\begin{align*}
 \hat{\nu}_{\glob,l} &= \nu_{\glob}-\frac{D_lD_{l-1}}{2} \quad \text{and} \quad
     \hat{\delta}_{\glob,l}  = \sqrt{\delta_{\glob}^2 + \sum_{d}^{D_l} \sum_{d'}^{D_{l-1}} \E \left [ \frac{1}{\psi_{l,d,d'}} \right] \E \left[ W_{l,d,d'}^2  \right]}.
\end{align*}
\paragraph{Local shrinkage parameters:} the parameters $\bpsi$ are independent across and within layers (and can be updated in parallel) with a GIG variational posterior:
\begin{align}
    q(\bpsi) & = \prod_{l}^{L+1} \prod_{d}^{D_l} \prod_{d'}^{D_{l-1}} \GIG \left (\psi_{l,d,d'} \; \mid \; \hat{\nu}_{\loc,l,d,d'},\hat{\delta}_{\loc,l,d,d'}, \lambda_{\loc,l} \right ), \label{eq:vbpsi}
\end{align}
where for $l=1, \dots, L+1, \; d=1,\ldots, D_l, \; d'=1,\ldots, D_{l-1}$,
\begin{align*}
    \hat{\nu}_{\loc,l,d,d'} &= \nu_{\loc,l}-\frac{1}{2} \quad \text{and} \quad 
    \hat{\delta}_{\loc,l,d,d'} =  \sqrt{ \E \left [ \frac{1}{\tau_l} \right ] \E \left[ W_{l,d,d'}^2  \right]  + \delta_{\loc,l}^2}.
\end{align*}
\paragraph{Covariance matrix:} the diagonal elements of the covariance matrix $\bmeta_l$ are independent across and within layers (and can be updated in parallel) with an inverse-Gamma variational posterior:
\begin{align}
 q(\bmeta) =  \prod_l^{L+1} \prod_d^{D_l} \IG(\eta_{l,d}^2 \mid \alpha_{l,d}, \beta_{l,d}), \label{eq:vbeta}
 \end{align}
 where for the hidden layers $l=1, \ldots, L$, the updated variational parameters for $d=1, \ldots, D_l$ are given by
 \begin{align*}
     \alpha_{l,d} &= \alpha^h_0+ \frac{N}{2},\\ 
    \beta_{l,d} & = \beta^h_0  + \frac{1}{2}\sum_{n}^N  \left(  \E \left [\blu{\mathrm{a}_{n,l,d}} \right ] - \E \left [\gamma_{n,l,d} \right ]   \E \left [\widetilde{\bW}_{l,d} \right ]   \E \left [\widetilde{\ba}_{n,l-1} \right ]\right )^2  + \E \left [a^2_{n,l,d} \right ] -  \E \left [\blu{\mathrm{a}_{n,l,d}} \right ]^2 \\
&  +  \frac{1}{2}\sum_{n}^N\E \left [\gamma_{n,l,d} \right ] \Tr \left ( \E \left [ \widetilde{\bW}_{l,d}^T \widetilde{\bW}_{l,d} \right ]  \E \left [\widetilde{\ba}_{n,l-1} \widetilde{\ba}_{n,l-1}^T \right ] \right ) - \E \left [\gamma_{n,l,d} \right ]^2  \Tr \left ( \E \left [\widetilde{\bW}_{l,d}^T \right ] \E \left [\widetilde{\bW}_{l,d} \right ]  \E \left [\widetilde{\ba}_{n,l-1} \right ] \E \left [ \widetilde{\ba}_{n,l-1}^T \right ]  \right ).
\end{align*}
And for the final layer, the updated variational parameters for $d=1, \ldots, D_{L+1}$ are given by
 \begin{align*}
   \alpha_{L+1,d} &= \alpha_0  + \frac{N}{2}, \\
    \beta_{L+1,d}&  = \beta_0 +  \frac{1}{2}\sum_{n}^N  \left( y_{n,d} -  \E \left [  \widetilde{\bW}_{L+1,d} \right ]  \E [ \widetilde{\ba}_{n,L}] \right)^2 \\
    & +\frac{1}{2} \sum_n^N  \Tr \left ( \E \left [  \widetilde{\bW}_{L+1,d}^T \widetilde{\bW}_{L+1,d} \right ] \E \left [ \widetilde{\ba}_{n,L} \widetilde{\ba}_{n,L}^T\right] \right) - \Tr \left (\E [  \widetilde{\bW}_{L+1,d}]^T \E[\widetilde{\bW}_{L+1,d} ] \E [ \widetilde{\ba}_{n,L}] \E[\widetilde{\ba}_{n,L}^T]\right ).
\end{align*} 
Note that in the above, $\bW_{l,d}$ represents the $d$-th row of the weight matrix $\bW_{l}$. Additionally, for $l = 1, \ldots, L+1$ we introduce the notation $\widetilde{\bW}_{l,d}= (b_{l,d}, \bW_{l,d})$ and $\widetilde{\bW} = (\bb, \bW)$, and let the vector $\widetilde{\ba}_{n,l}$ represent the stochastic activation augmented with an entry of one, i.e. $\widetilde{\ba}_{n,l} = (1, \ba_{n,l}^T)^T$.

\paragraph{Weights and biases:} the weights and biases are independent across layers and within layers, independent across the $D_l$ regression problems, with a Gaussian variational posterior:
\begin{align}
    & q( \bb, \bW) = \prod_{l}^{L+1} \prod_d^{D_l} \Norm \left (\widetilde{\bW}_{l,d}\mid \mathbf{m}_{l,d}, \mathbf{B}_{l,d}\right ),\label{eq:vbWb}
\end{align}    
 where for the hidden layers $l=1, \ldots, L$, the updated variational parameters for $d=1, \ldots, D_l$ are given by
 \begin{align*}
    \mathbf{B}_{l,d}^{-1} &= \mathbf{D}^{-1}_{l,d}+\sum_{n}^N   \left ( \frac{1}{T^2} \E \left [\omega_{n,l,d}\right ] + \E \left [ (\eta_{l,d})^{-2}\right ] \E \left [\gamma_{n,l,d}\right ]  \right) \E \left [\widetilde{\ba}_{n, l-1} \widetilde{\ba}_{n, l-1}^T \right ],   \\
    \mathbf{m}_{l,d}^T  &= \mathbf{B}_{l,d}\left (\sum_{n}^N \E \left [  (\eta_{l,d})^{-2} \right ] \E \left [   \gamma_{n,l,d}\right ] \E \left [ \blu{\mathrm{a}_{n,l,d}} \widetilde{\ba}_{n,l-1}\right ] + \frac{1}{T} \E \left [\widetilde{\ba}_{n, l-1} \right ] \left ( \E \left [\gamma_{n,l,d} \right ]  - \frac{ 1}{2}\right)  \right),
    \end{align*}
     and for the final layer, the updated variational parameters for $d=1, \ldots, D_{L+1}$ are given by
\begin{align*}
    \mathbf{B}^{-1}_{L+1,d} &=\bD^{-1}_{L+1,d} +\E \left [(\eta_{L+1,d})^{-2} \right ] \sum_{n}^N \E \left [\widetilde{\ba}_{n,\blu{L}} \widetilde{\ba}_{n,\blu{L}}^T \right], \\
      \mathbf{m}_{L+1,d}^T  &=\mathbf{B}_{L+1,d}\E \left [(\eta_{L+1,d})^{-2} \right ] \left ( \sum_{n}^N y_n \E \left [\widetilde{\ba}_{n,\blu{L}} \right]  \right ), 
\end{align*}
where for $l=1, \ldots, L+1$ and $d=1, \ldots, D_{l}$,
\begin{align*}
\bD_{l,d}^{-1} &= \diag \left ( s_0^{-2}, \E \left [\tau_l^{-1}\right ] \E \left [ \psi_{l,d,1}^{-1} \right ], \ldots, \E \left [\tau_l^{-1}  \right ] \E \left [ \psi_{l,d,D_{l-1}}^{-1} \right ]\right ). %\\
  %D_{L+1,d}^{-1}& = \diag \left ( s_0^{-2}, \E \left [\frac{1}{\tau_{L+1}} \right ] \E \left [ \frac{1}{\psi_{L+1,d,1}} \right ], \ldots, \E \left [\frac{1}{\tau_{L+1}} \right ] \E \left [ \frac{1}{\psi_{L+1,d,D_{L}}} \right ]\right ),
\end{align*}

\begin{figure}
    \centering
    \includegraphics[width = 0.68\linewidth]{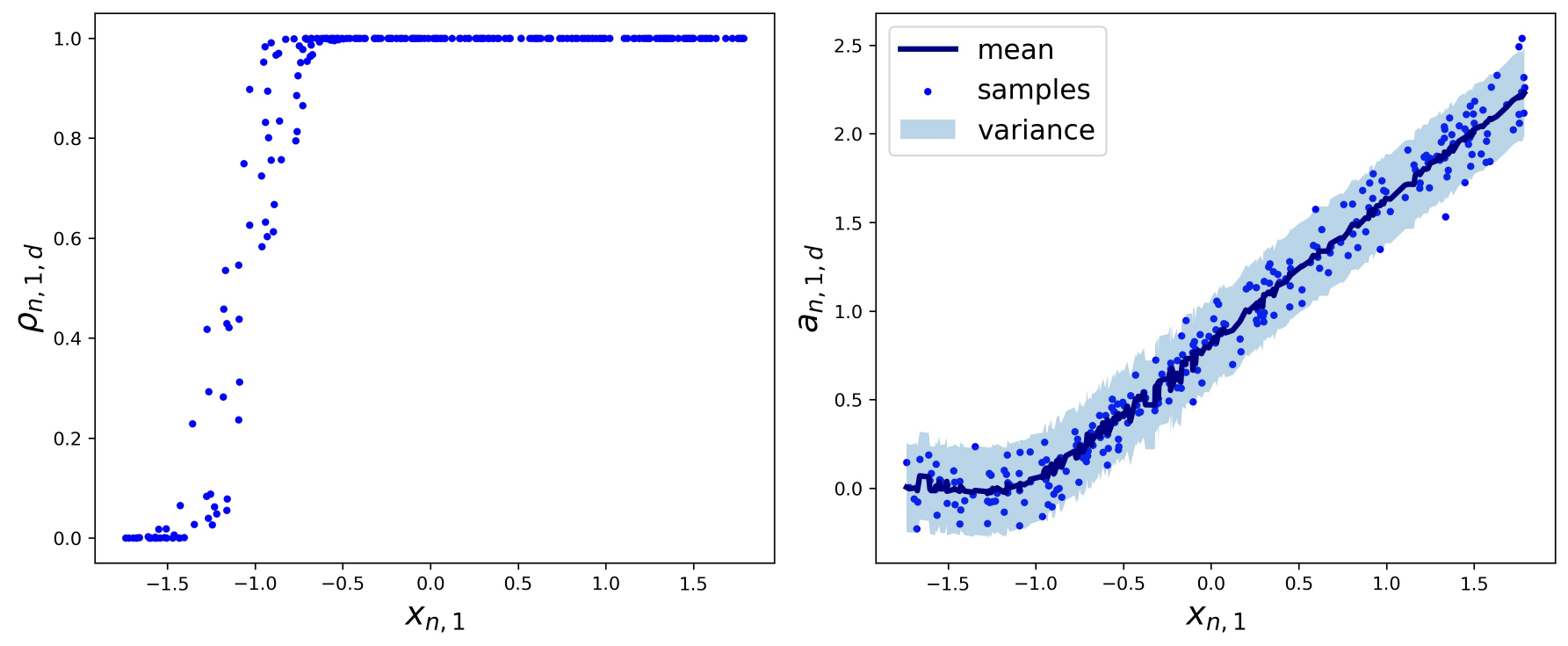}
    \includegraphics[width = 0.3\linewidth]{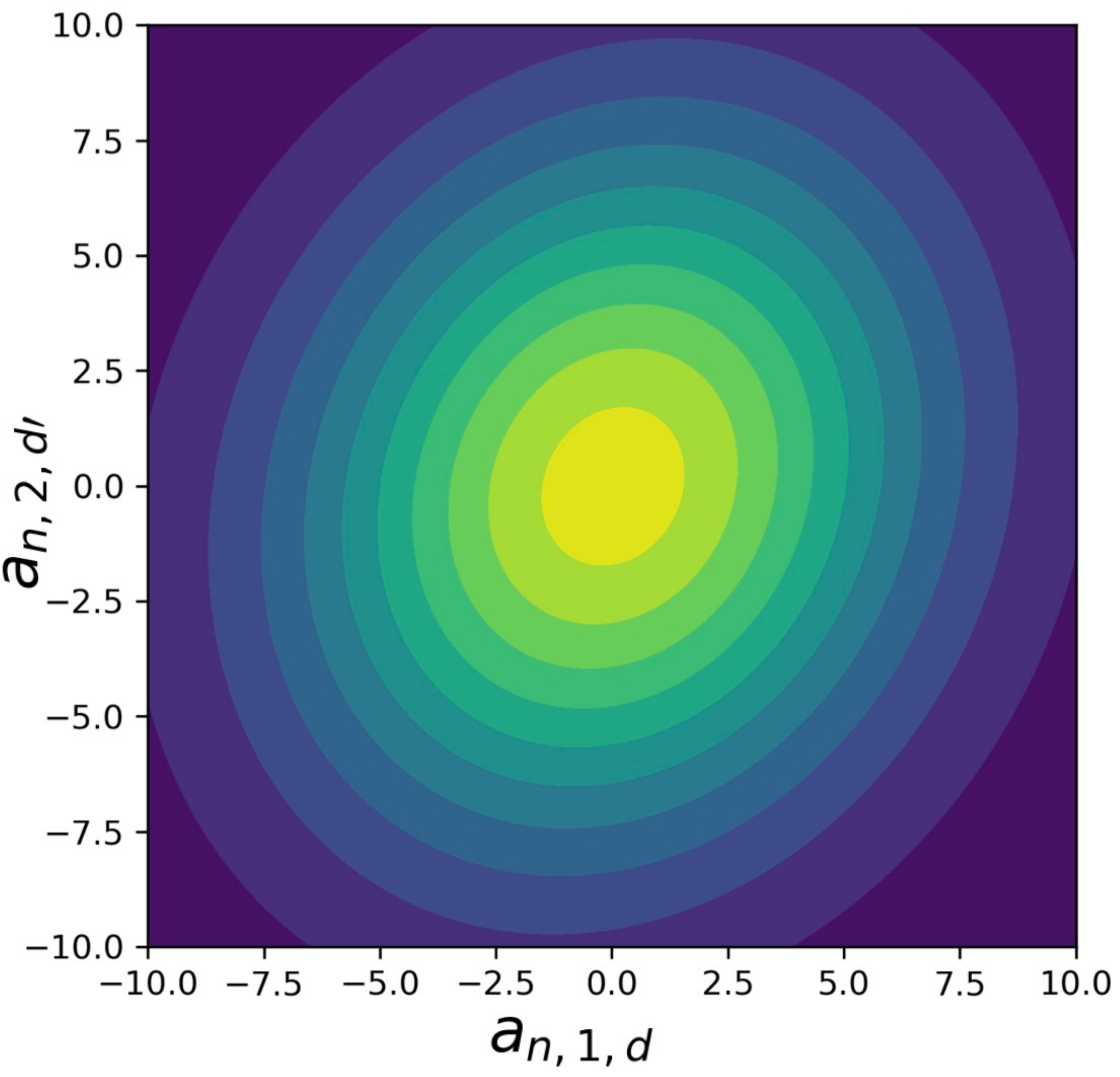}
    \caption{ \blu{ An illustration of the variational posterior of the binary and stochastic activations. The variational posterior of  $\gamma_{n,1,d}$ (on the left) and $a_{n,1,d}$ (in the middle), both as a function of $x_{n,1}$ across all observations, along with the joint distribution of $(a_{n,1,d}, a_{n,2,d'})$ (on the right) in the case of the toy example of \cref{sec:experiments} for particular values of $d, d', n$. }}
    \label{fig:varposterioraandrho}
\end{figure}

\paragraph{Polya-Gamma augmented variables:}  $\bomega$ are independent across observations $n=1,\ldots, N$, layers $l=1,\ldots, L$, and width $d=1, \ldots, D_l$, with a Polya-Gamma variational posterior:
\begin{align}
q(\bomega) &=  \prod_{n}^N \prod_{l}^L \prod_{d}^{D_l} \PG(  \omega_{n,l,d} \mid 1, A_{n,l,d}), \label{eq:vbomega} 
\end{align}
with updated variational parameters: 
\begin{align*}
    A_{n,l,d} &= \frac{1}{T} \sqrt{\left ( \Tr \left ( \E \left [ \widetilde{\bW}_{l,d}^T \widetilde{\bW}_{l,d} \right ] \E \left [ \widetilde{\ba}_{n,l-1} \widetilde{\ba}_{n,l-1}^T\right ] \right ) \right )}.
\end{align*}
Note that simulating from or evaluating the density of the PG is not necessary, and the CAVI updates of the other parameters only require computing the expectation of $\bomega$ with respect to the variational posterior in \cref{eq:vbomega}, which is straightforward to compute (\cref{eq:pgmean}). 
\paragraph{Binary activations:} $\bgamma$ are independent across observations $n=1,\ldots, N$, layers $l=1,\ldots, L$, and width $d=1, \ldots, D_l$, with a Bernoulli variational posterior:
\begin{align}
  q(\bgamma)  &= \prod_{n}^N\prod_{l}^L \prod_{d}^{D_l}\Bern\left (\gamma_{n,l,d} \mid \rho_{n,l,d} \right ), \label{eq:vbgamma}
  \end{align}
  with 
  \begin{align*}
  \rho_{n,l,d} =    &  \sigma \left (-\frac {\E \left [ \eta_{l,d}^{-2} \right ] }{2}   \Tr \left (\E \left[\widetilde{\bW}_{l,d}^T \widetilde{\bW}_{l,d}\right ] \E \left [\widetilde{\ba}_{n, l-1} \widetilde{\ba}_{n, l-1}^T \right ] \right )  + \E \left [ \eta_{l,d}^{-2} \right ] \E \left [ \widetilde{\bW}_{l,d} \right ] \E \left [\widetilde{\ba}_{n, l-1} \blu{\mathrm{a}_{n,l,d}} \right ] + \frac{1}{T} \E \left [  \widetilde{\bW}_{l,d}  \right ] \E \left [\widetilde{\ba}_{n, l-1}\right ] \right).
\end{align*}
%\SW{To illustrate the variational posterior, for a simple example (e.g. the toy example), can you plot $\rho_{n,1,d}$ vs $x_{n,1}$ for some choice of $d$ that gives an interesting plot?}. \AS{\cref{fig:varposterioraandrho}}
\blu{The parameters $\rho_{n,l,d} $ represent the posterior probability that the node is active and are illustrated} for the toy example of Section \ref{sec:toy} in \cref{fig:varposterioraandrho}. 

%where $\sigma(x) = \exp(x)/(1+\exp(x)) $ is the logistic function. 
\paragraph{Stochastic activations:} $\ba$ are independent across observations $n=1,\ldots, N$ and conditionally Gaussian given the previous layer with variational posterior: 
\begin{equation}
    q(\ba) = \prod_{n=1}^N \prod_{l=1}^{L} \Norm \left (\ba_{n,l} \mid  \mathbf{t}_{n,l}+ \bM_{n,l} \ba_{n,l-1}, \mathbf{S}_{n,l} \right ), \label{eq:vba}
\end{equation}
where denote $\ba_{n,0} \coloneqq \bx_{n}, \mathbf{S}_{n,L} \coloneqq \mathbf{S}_{L}$ and the updated variational parameters for $n=1,\ldots,N$ and $l = 1, \ldots, L-1$ are 
\begin{align*}
   \mathbf{S}^{-1}_{n, l} &  = \hat{\mathbf{\Sigma}}_l^{-1} - \mathbf{M}_{n, l+1}^T \mathbf{S}_{n,l+1}^{-1}\mathbf{M}_{n, l+1} + \sum_{d=1}^{D_{l+1}} \left ( \E \left [\frac{1}{\eta_{l+1,d}^2} \right ] \E \left[ \gamma_{n,l+1,d}\right ] + \frac{1}{T^2}  \E \left [ \omega_{n,l+1,d}\right ] \right )\E \left [ \bW_{l+1,d}^T \bW_{l+1,d} \right ], \\
   \mathbf{t}_{n, l}  & = \mathbf{S}_{n,l} \left ( \mathbf{M}_{n, l+1}^T \mathbf{S}_{n, l+1}^{-1} \mathbf{t}_{n,l+1} + \hat{\mathbf{\Sigma}}_l^{-1}  \E \left[ \bgamma_{n,l}\right ] \odot  \E \left[ \bb_{l}\right ] + \frac{1}{T} \sum_{d=1}^{D_{l+1}}\E \left [ \bW_{l+1,d}^T \right ]  \left ( \E \left [ \gamma_{n,l+1,d} \right ]  - \frac{1}{2}\right )  \right. - \\
    & \left. - \sum_{d=1}^{D_{l+1}} \left ( \E \left [\frac{1}{\eta_{l+1,d}^2} \right ] \E \left[ \gamma_{n,l+1,d}\right ] +  \frac{1}{T^2} \E \left [ \omega_{n,l+1,d}\right ]\right)
\E \left [ \bW_{l+1,d} b_{l+1,d}\right ] \right ),  \\
 \mathbf{M}_{n,l} &= \mathbf{S}_{n,l} \hat{\mathbf{\Sigma}}_l^{-1} \E \left[ \bgamma_{n,l}\right ]  \mathbf{1}^T_{D_{l-1}} \odot \E \left [  \bW_{l}\right ], \\
 \hat{\mathbf{\Sigma}}_{l}^{-1} &=\diag \left ( \E \left[ \eta_{l,1}^{-2} \right ], \dots, \E \left[\eta_{l,D_l}^{-2}\right ]\right ). 
\end{align*}
And for the final layer with $n=1,\ldots,N$ and $l = L$,
\begin{align*}
    & \mathbf{S}^{-1}_{L} = \hat{\mathbf{\Sigma}}_{L}^{-1} + \sum_{d=1}^{D_{L+1}}\E \left [  \frac{1}{\eta_{L+1,d}^2}\right ]  \E \left [ \bW_{L+1,d}^T \bW_{L+1,d} \right], \\
    & \mathbf{t}_{n,L} = \mathbf{S}_{L} \left ( \left (\sum_{d=1}^{D_{l+1}} \E \left [\frac{1}{\eta_{L+1,d}^2} \right ] 
   \left ( -  \E \left [ \bW_{L+1,d}^T b_{L+1,d} \right] +  \E \left [ \bW_{L+1,d}^T \right] y_{n,d} \right ) \right )  + \hat{\mathbf{\Sigma}}_{L}^{-1}  \E \left[ \bgamma_{n,L}\right ] \odot \E \left [ \bb_{L}\right ] \right ), \\
   & \mathbf{M}_{n,L} = \mathbf{S}_{L} \hat{\mathbf{\Sigma}}_{L}^{-1} \E \left[ \bgamma_{n,L}\right ]  \mathbf{1}^T_{D_{L-1}} \odot \E \left [  \bW_{L}\right ], \\
   &\hat{\mathbf{\Sigma}}_{L}^{-1}=\diag \left ( \E \left[\eta_{L,1}^{-2} \right ], \dots, \E \left[\eta_{L,D_L}^{-2}\right ]\right ). 
\end{align*}
\cref{fig:varposterioraandrho} illustrates on the toy example of \cref{sec:toy} how the variational posterior of the stochastic activations (middle) resembles a smoothed, noisy ReLU. Due the independence assumption between the stochastic and binary activations, the potentially bimodal bow tie distribution (\cref{eq:amarginal}) is approximated with a unimodal Gaussian in the variational framework, which may better approximate the true posterior when the temperature is not too large relative to the noise (see \cref{fig:bowtie}). In addition, the proposed approximation has the advantage of avoiding explicit assumptions of independence between layers, \blu{enabling it to capture dependencies between stochastic activations across} layers, as illustrated for the toy example in \cref{fig:varposterioraandrho} (right).  
%\SW{To illustrate the variational posterior, for a simple example (e.g. the toy example), can you plot samples, mean, and shaded region for $a_{n,1,d}$ vs $x_{n,1}$ for some choice of $d$ that gives an interesting plot?} \AS{How is that? I added folder "options" with possible choices}

The corresponding optimization objective, i.e. the ELBO in \cref{eqn:ELBO}, is available in closed form and provided in the \cref{appendix:ELBO}.

\subsection{VI with EM} \label{sec:VIwithEM}
% \SW{
% In this, case, we considering the variational parameters as updated parameters, optimizing the ELBO, simply corresponds to find the parameters $h$ , which maximize the expected log prior:
% \begin{align*}
%     h_{\text{glob}} = \argmax \E\left[\sum_{l=1}^{L+1} \log(\GIG(\tau_l \mid \nu_{\glob}, \delta_\glob, \lambda_\glob) \right].
% \end{align*}
% Thus, we need a few extra terms in the formulas below (but it may not matter for the inverse gamma case, since I think the missing terms don't depend on $\delta$). 
% } 
\begin{algorithm}[t]
\caption{VI with EM} \label{alg:CAVIwithEM}
\begin{algorithmic}
\REQUIRE Initialize hyperparameters
\WHILE{ELBO has not converged}
\FOR{${l = 1, \ldots, L}$} 
\STATE update $\nu_{\glob, l}$ and $ \delta_{\glob, l} \quad$ \COMMENT{parameters of $\tau_l$}
\STATE update $\nu_{\loc, l, d, d'} $ and $ \delta_{\loc, l, d, d'} \quad$ \COMMENT{parameters of $\psi_{l,d,d'}$ for $d = 1 \ldots D_{l},\; d' = 1 \ldots D_{l-1}$}
\STATE update $\alpha_{l,d}$ and $ \beta_{l,d}\quad$ \COMMENT{parameters of $\eta_{l,d}$ for $d = 1 \ldots D_{l}$}
\STATE update $A_{n,l,d}\quad$ \COMMENT{parameter of $\omega_{n, l,d}$  for $d = 1 \ldots D_{l},\; n = 1 \ldots N$}
\ENDFOR
\STATE update $\nu_{\glob, L+1} $ and $\delta_{\glob, L+1}  \quad$ \COMMENT{parameters of $\tau_{L+1}$}
\STATE update $\nu_{\loc, L+1, d, d'} $ and $\delta_{\loc, L+1, d, d'}\quad$ \COMMENT{parameters of $\psi_{L+1,d,d'}$ for $d = 1 \ldots D_{y},\; d' = 1 \ldots D_{L}$}
\STATE update $\alpha_{L+1,d} $ and $\beta_{L+1,d}$ for $d = 1 \ldots D_{y}$
\FOR{${l = 1, \ldots, L}$}
\STATE update $\mathbf{S}_{n, l}, \; \mathbf{M}_{n, l} $ and $\mathbf{t}_{n, l} \quad$ \COMMENT{parameters of $\ba_{n,l}$   for $ n = 1 \ldots N$ }
\ENDFOR
\FOR{${l = 1, \ldots, L}$}
\STATE update $\mathbf{B}_{l,d} $ and $\mathbf{m}_{l,d} \quad$ \COMMENT{parameters of $(b_{l,d}, \bW_{l,d})$ for $d = 1 \ldots D_{l}$}
\STATE update $\rho_{n,l,d}\quad$ \COMMENT{parameter of $\gamma_{n, l,d}$  for $d = 1 \ldots D_{l},\; n = 1 \ldots N$}
\ENDFOR 
\STATE update $\mathbf{B}_{L+1,d} $ and $\mathbf{m}_{L+1,d} \quad$ \COMMENT{parameters of $(b_{L+1,d}, \bW_{L+1,d})$  for $d = 1 \ldots D_{y}$}
% \STATE Compute $\E_{\delta_{\glob}}[\text{ELBO}]$
\STATE update $h_{\glob} \quad$ \COMMENT{EM for global hyperparameter}
% \FOR{${l = 1, \ldots, L}$}
% \STATE update $h_{\loc, l} \quad$ \COMMENT{EM for local hyperparameters}
% \ENDFOR 
\ENDWHILE
\end{algorithmic}
\end{algorithm}
The hyperparameters can play a crucial role in Bayesian neural networks. When dealing with the sparsity-inducing priors setting an excessively large scale parameter weakens the shrinkage effects, whilst choosing a scale parameter that is too small may wipe out the effects of the important hidden nodes.
Manually picking suitable values is challenging, and instead, we seek a more efficient strategy, utilizing the similarity between the variational and \textit{expectation-maximization (EM) algorithms}. 
Specifically, we investigate the hybrid scheme combining VI with an EM step \citep{osborne2022latent} so that the steps of the CAVI algorithm proceed with the EM update to set the hyperparameter for global shrinkage variable $\btau$. Due to weak identifiability, we do not jointly update global and local hyperparameters. Let $h_{\glob}$ represent $\delta_{\glob}$ or $\lambda_{\glob}$ and consider the ELBO treated as a function of $h_{\glob}$, then the optimal values as approximate MAP estimates are: 
% Note that using the independence across layers and shrinkage parameters, we can solve $L+1$ separate optimization tasks to obtain local shrinkage hyperparameters and one task to obtain global hyperparameter.
\begin{align*}
h_{\glob}= \argmax   \E_{\glob}[\text{ELBO}],
\end{align*}
where
\begin{align*}
 \E_{\glob}[\text{ELBO}] = &\; \E\left[\sum_{l=1}^{L+1} \log(\GIG(\tau_l \mid \nu_{\glob}, \delta_\glob, \lambda_\glob) \right] \\
= & \; (L+1)\left( \nu_{\glob} \left( \log (\lambda_{\glob}) - \log (\delta_{\glob}) \right) - \log \left( 2K_{\nu_{\glob}}(\lambda_{\glob}\delta_{\glob})\right) \right) \\
+& \;\sum_{l=1}^{L+1}(\nu_{\glob} - 1)\E \left [ \log \tau_{l}\right ] - \frac{1}{2}\left(\delta_{\glob}^2 \E \left[ \frac{1}{\tau_{l}}\right ] + \lambda_{\glob}^2 \E\left [\tau_{l}\right ] \right).
%  \E_{\loc, l}[\text{ELBO}] = & \;\E\left[\sum_{d=1}^{D_{l}}\sum_{d'=1}^{D_{l-1}} \log(\GIG(\psi_{\loc, l} \mid \nu_{\loc, l}, \delta_{\loc, l}, \lambda_{\loc, l}) \right] \\ 
% = &\;  D_{l}D_{l-1} \left( \nu_{\loc, l} \left( \log (\lambda_{\loc, l}) - \log (\delta_{\loc, l}) \right) - \log \left( 2K_{\nu_{\loc, l}}(\lambda_{\loc, l}\delta_{\loc, l})\right) \right) + \\
% + &\;\sum_{d=1}^{D_{l}}\sum_{d'=1}^{D_{l-1}}(\nu_{\loc, l} - 1)\E \left [ \log \psi_{\loc, l,d,d'}\right ] - \frac{1}{2}\left(\delta_{\loc, l}^2 \E \left[ \frac{1}{\psi_{\loc, l,d,d'}}\right ] + \lambda_{\loc, l}^2 \E\left [\psi_{\loc, l,d,d'}\right ] \right)
\end{align*} 
In the case of the IG priors, one's aim is to set optimal $\delta_{\glob}$, in the case of the Gamma and IGauss priors, the parameters of interest are $\lambda_{\glob}$. We provide specific examples of the shrinkage parameters and the corresponding optimal values in \cref{sec:emappendix}. 
%The result of combining CAVI and the EM algorithm is described in \cref{alg:CAVIwithEM}. 

\blu{The resulting algorithm (described in \cref{alg:CAVIwithEM}) scales linearly with the number of hidden layers and the number of samples but cubically with the number of hidden units; the computational complexities corresponding to individual factors of the variational family are provided in \cref{appendix:vb}. In the following subsections, we discuss two strategies to improve scalability. First, to handle large $N$, CAVI can be combined with subsampling through stochastic VI \citep{hoffman2013stochastic}, and second, a post-processing node selection algorithm is proposed to obtain a sparse variational posterior for faster predictive inference. }

% Example joint $q$: 
% \begin{align*}
%     & q\left (\begin{pmatrix}
%         \ba_{n,1} \\ \ba_{n,2}
%     \end{pmatrix} \right ) = \Norm\left (\begin{pmatrix}
%         \ba_{n,1} \\\ba_{n,2}
%     \end{pmatrix} | \begin{pmatrix}
%         \mathbf{t}_{n,1}+ \bM_{n,1} \bx_{n} \\ \mathbf{t}_{n,2}+ \bM_{n,2} \mathbf{t}_{n,1}+ \bM_{n,2} \bM_{n,1} \bx_{n}
%     \end{pmatrix}, \begin{pmatrix}
%         \mathbf{S}_{n,1}^{-1} & \mathbf{S}_{n,1}^{-1}\bM_{n,2}^T \\
%         \bM_{n,2} \mathbf{S}_{n,1}^{-1} & \mathbf{S}_{n,2}^{-1} + M_{n,2} \mathbf{S}_{n,1}^{-1}\bM_{n,2}^T        
%     \end{pmatrix}^{-1} \right )
% \end{align*}

% \begin{figure}[h]
%     \centering
% \includegraphics[scale=0.5]{pictures/a1a2.pdf}
% \end{figure}
% The Half-Thresholding Method of \citep{Tang_VarSel} selects a variable if the ratio of the posterior mean to the OLS estimate of the corresponding coefficient is greater than $0.5$, the method achieves variable selection consistency in the case of polynomial-tailed shrinkage priors but fails in the exponential-tailed case.

% 
 % One example is the signal adaptive variable selector of \citep{ray2018signal} which finds a sparse solution for the weights which is close to the posterior mean, in terms of the Euclidean distance to an approximation of the ``model fit" for each layer. 
\subsection{Stochastic Variational Inference}\label{sec:svivbnn}
{\blu{At each iteration, CAVI has to cycle through the entire data set, which can be computationally expensive and inefficient for large sample sizes. An alternative to coordinate ascent is gradient-based optimization, which extends the algorithm by employing stochastic variational inference \cite[SVI][]{hoffman2013stochastic} and subsampling. %First, recall the posterior of the parameters and latent variables for the bow tie neural networks derived in \cref{sec:modelposterior} and illustrated by \cref{fig:DAG}, and the assumed variational family is given by \cref{eq:vbnnmfamily}. In \cref{sec:vb}, we show that 
Forst. recall that the variational posterior of the latent variables $\ba, \bgamma, \bomega$ factorizes across data points, that is 
\begin{align*}
    q(\ba) q(\bgamma) q(\bomega)  = \prod_{n=1}^N q(\ba_{n}) q(\bgamma_{n}) q(\bomega_{n}).
\end{align*}
To highlight the need for stochastic VI, we observe that for each layer $l = 1, \ldots, L$ the ordinary coordinate ascent needs to iterate through $N$ local variational parameters corresponding to variables $\ba_{n,l}, \bgamma_{n,l}$ and $\bomega_{n,l}$ with $\bgamma_{n,l}$ and $\bomega_{n,l}$ both having $D_{l}$ variational parameters to update, and $\ba_{n,l}$ having of order $D_l^2$ variational parameters.  For large $N$, this can be computationally burdensome. To overcome this, stochastic VI uses the coordinate ascent variational updates obtained in \cref{sec:vb} to set the local parameters for a subset of points, and the intermediate global parameters are obtained by following the noisy natural gradient of ELBO \citep{hoffman2013stochastic}. 

At each iteration $t$, the SVI algorithm first %updates the global variational parameters $\bm{\delta}_{\glob}^{(t)},  \bm{\delta}_{\loc}^{(t)}$ as in the classical CAVI via \cref{eq:vbtau,eq:vbpsi}, where we note that, due to the simple form of the variational update, parameters $\bm{\nu}_{\glob}, \bm{\nu}_{\loc}$ are updated only during the first iteration of the algorithm. Then one sets the learning rate $\ell_{t}$ and 
uniformly samples a collection of indices $S(t)$ without replacement, where the computational gains are obtained as long as $|S(t)| \ll N$.  For each index $s \in S(t)$, the local variational parameters are optimized in a coordinate ascent manner with updates in \cref{eq:vba,eq:vbomega,eq:vbgamma}, and the convergence is monitored with noisy estimates of the local ELBO, that is
\begin{align*}
&\E \left [  \log p (\by, \ba, \bgamma,\bomega | \bW, \bb, \bSigma)\right ]   - \E \left [\log  q(\ba) \right ] - \E \left [\log q(\bgamma) \right ] - \E \left [\log q(\bomega)\right ], 
\end{align*}
the estimate of which is provided \cref{ap:svisup}.
Next, the global parameters, $\bW,\bb$ and $\bmeta$, are updated  %by taking a step of size $\ell_{t}$ in the direction of the natural gradient and updating  global parameters 
via a linear combination of previous values  and intermediate updates 
\begin{align}
    (1- \ell_{t})\text{parameter}^{(t-1)} + \ell_{t} \times \text{intermediate parameter}^{(t)}, \label{eq:sviupd}
\end{align}
where  $\ell_{t}$ is the learning rate and the intermediate parameters are obtained via \cref{eq:vbeta,eq:vbWb} but with the sufficient statistics, which involve sums over $N$, replaced with the scaled sums over $S(t)$.  %Note that  $-(\bm{\alpha}+1)$ is the natural parameter of global variable $\bmeta$, so that $\bm{\alpha}$ is only updated once and does not require an intermediate step. 
Additional details and the step-by-step algorithm are provided in \cref{ap:svisup}.
The convergence of the stochastic gradient descent depends on the choice of the step sizes $\ell_{t}$  in the Robbins-Monro sequence \citep{RobbinsMonro1951}, and we follow \citep{hoffman2013stochastic}, who set 
\begin{align}
   \ell_{t} = (1 + t)^{-k}, \; k \in (0.5, 1], \label{eq:stepsizeforSVI}
\end{align}
where $k$ is the forgetting rate. %, and the delay to down-weight the first iterations is set to $1$. 
We monitor the convergence of the algorithm by obtaining noisy estimates of the ELBO (provided in \cref{ap:svisup}) where the sums over $n =1, \ldots, N$ are replaced with the scaled sums over $n \in S$.}}
\subsection{Inferring the Network Structure} \label{sec:selectvariables}
The choice of the network architecture has significant practical implications on the generalization of the model, and so sparsity-promoting priors for the network weights have emerged as a promising approach for increased robustness to the choice of architecture. %Instead of the classical two-group discrete mixture priors, 
In this article, we focus on a class of continuous shrinkage priors, which results in more tractable computations, yet also implies non-zero posterior means and does not lead to automatic network architecture selection.  \blu{To address this, we consider a post-processing node selection algorithm to obtain a sparse variational posterior, which both aids interpretability and improves scalability of predictive inference (discussed in Section \ref{sec:prediction})}

In Bayesian linear regression with shrinkage priors, several post-processing methods have been proposed to yield a sparse solution (see e.g.\citep{Piironen2020Varsel, LI_Varsel, griffin2024expressing}). The method known as decoupling shrinkage and selection (DSS) \citep{hahn2014decoupling} obtains sparse estimates of the weights by minimizing the sum of the predictive loss function with a parsimony-inducing penalty. An alternative approach is the the penalized credible regions (PenCR) method \citep{Zhang_VarSelect}, which identifies the "sparsest" solution in posterior credible regions corresponding to different levels; it is shown to perform well in the case of global-local shrinkage priors and under certain assumptions, PenCR produces the same results as DSS. %The PenCR centres around identifying the "sparsest" solution in posterior credible regions corresponding to different levels. 
Similarly, %having the variational posterior distribution of weights and biases at hand, 
we propose to make use of credible intervals to select nodes. Following \citep{LiLin2010}, we implement an automatic \textit{credible interval criterion} which selects a node as long as its credible interval doesn't cover zero. Specifically, %let $Q$ be the post-variational cumulative distribution function of weights and 
recall that the variational posterior of the weights is  $W_{l,d,d'} \sim \Norm(m^W_{l,d,d'}, B^W_{l,d,d'})$ for $l=1,\ldots,L, d= 1, \dots, D_{l}, d' = 1, \ldots, D_{l-1}$, where $\mathbf{m}^W_{l,d}$ and $\mathbf{B}^W_{l,d}$ denote the subsets of the mean $\mathbf{m}_{l,d}$ and covariance matrix $\mathbf{B}_{l,d}$ corresponding to the weights. Then, we obtain sparse weights $\widehat{W}_{l,d,d'}$ with sparse variational distribution $\widehat{q}(b_{l,d}, \widehat{\bW}_{l,d})$  for some $ l \in  \mathcal{L} \subseteq \{1, \ldots, L+1\}, \; d \in \mathcal{D}_l \subseteq \{ 1, \ldots, D_l\}, d' \in \mathcal{D}_{l-1} \subseteq \{ 1, \ldots,  D_{l-1}\}$, defined by setting: 

\begin{align*}
     \widehat{W}_{l,d,d'} &\sim     
     \begin{cases}
       \Norm\left (m^W_{l,d,d'}, \left(B^W_{l,d}\right)_{d'd'}\right ) &\text{ if } \max \left ( Q(W_{l,d,d'} >0) , Q(W_{l,d,d'}<0)\right )  \geq \kappa, \\ 
        \delta_0   &\text{ otherwise, }
    \end{cases}
\end{align*}
where  $Q(W_{l,d,d'}<0) = 1- Q(W_{l,d,d'}>0) =\Phi(- m^W_{l,d,d'}/\sqrt{(B^W_{l,d})_{d'd'}})$. 
% \begin{align*}
%      & \widehat{W}_{l,d,d'} \sim  \left\{ 
%     \begin{aligned}
%       & \Norm\left (m^W_{l,d,d'}, \left(B^W_{l,d}\right)_{d'd'}\right ) \text{ if } \max \left ( Q(W_{l,d,d'} >0) , Q(W_{l,d,d'}<0)\right )  \geq \kappa, \\ 
%       & 0  \text{ otherwise, }
%     \end{aligned}
%   \right.  \\ 
%        &  \text{ where } \kappa \text{ is some threshold and }  Q(W_{l,d,d'}<0)  =\Phi(- \frac{m^W_{l,d,d'}}{\sqrt{\left(B^W_{l,d}\right)_{d'd'}}}).
%\end{align*} 
 The threshold $\kappa$ is chosen to control the \textit{Bayesian false discovery rate}, which is calculated as
 \begin{align*}
    \widehat{\text{FDR}}(\kappa) = \frac{\sum_{l,d,d'} (1-\mathcal{Q}_{l,d.d'})\1(\mathcal{Q}_{l,d.d'}>\kappa) }{\sum_{l,d,d'} \1(\mathcal{Q}_{l,d.d'}>\kappa)},
\end{align*}
 with $\mathcal{Q}_{l,d.d'} =  \max \left ( Q(W_{l,d,d'} >0) , Q(W_{l,d,d'}<0)\right )$. %Based on the cumulative variation distribution, for each weight, we define the threshold:
%$$\mathcal{Q}_{l,d.d'} =  \max \left ( Q(W_{l,d,d'} >0) , Q(W_{l,d,d'}<0)\right ).$$ Then, the estimated false discovery rate is calculated as
%\begin{align*}
%    \widehat{\text{FDR}}(\kappa) = \frac{\sum_{l,d,d'} (1-\mathcal{Q}_{l,d.d'})\1(\mathcal{Q}_{l,d.d'}>\kappa) }{\sum_{l,d,d'} \1(\mathcal{Q}_{l,d.d'}>\kappa)}.
%\end{align*}
Specifically, for a specified error rate $\alpha$, $\kappa$ is set to satisfy $ \widehat{\text{FDR}}(\kappa)< \alpha$. \cref{alg:vas} describes the node selection procedure which begins with ordering $\mathcal{Q}_{l,d.d'}$ in the descending order and going down through the thresholds to assign $\kappa$ to the smallest $\mathcal{Q}_{l,d.d'}$ such that its false discovery rate doesn't exceed $\alpha$.
 % In order to avoid division by 0, I find $\overline{\kappa} = \max(\text{PCI}_{l,d.d'})$ and only consider $\kappa < \overline{\kappa}.$

\begin{algorithm}[t!]
\caption{Node selection algorithm} \label{alg:vas}
\begin{algorithmic}
\REQUIRE $\mathcal{I} = \{ \mathcal{Q}_{l,d.d'} \mid  l = 1 \ldots L, \;  d = 1 \ldots D_{l+1}, \;  d' = 1 \ldots D_{l} \}$. 
\STATE{$\hat{\kappa} = \max (\mathcal{I})$}
\STATE{$\mathcal{I} = \mathcal{I}\setminus \hat{\kappa}$}
\IF{$\widehat{\text{FDR}}(\max (\mathcal{I}))< \alpha$}
\STATE{$\hat{\kappa} = \max (\mathcal{I})$}
\STATE{$\mathcal{I} = \mathcal{I}\setminus \hat{\kappa}$}
\ELSE
\STATE{\textbf{break}}
\ENDIF
\FOR{ $l = 1 \ldots L, \;  d = 1 \ldots D_{l+1}, \;  d' = 1 \ldots D_{l}$}
\IF{$\mathcal{Q}_{l,d,d'} \geq \hat{\kappa}$}
\STATE{$\widehat{W}_{l,d,d'} \sim  \Norm\left (m^W_{l,d,d'}, \left(B^W_{l,d}\right)_{d'd'}\right )$}
\ELSE
\STATE{$\widehat{W}_{l,d,d'} = 0$ a.s.}
\ENDIF
\ENDFOR
\FOR{ $l = L+1,\ldots 2,d = 1,\ldots D_l,$}
\IF{$\widehat{W}_{l,d}= 0$ a.s.}
\STATE{$\widehat{W}_{l-1,d',d} = 0 \text{ a.s.} \;\forall \; d' = 1, \ldots, D_{l-1}$ }
\ELSE
\IF{$\widehat{W}_{l-1,d',d} = 0 \text{ a.s.} \; \forall \; d' = 1, \ldots, D_{l-1}$}
\STATE{$\widehat{W}_{l,d} = 0$ a.s. }
\ENDIF
\ENDIF
\ENDFOR
\ENSURE $\widehat{q}(b_{l,d}, \widehat{\bW}_{l,d}), \;
l \in  \mathcal{L},  \; d \in \mathcal{D}_l, \; d' \in \mathcal{D}_{l-1}.$
\end{algorithmic}
\end{algorithm}
Once we sweep through $\mathcal{Q}_{l,d.d'}$, we do a backwards pass to remove the nodes with no connections. 
If the node has no outgoing connections, then all the incoming connections need to be removed, and conversely, if the node has no incoming connections, then all the outgoing connections can be removed. An example of the network resulting after applying the \cref{alg:vas}
is illustrated by the \cref{fig:toyexample}.
%\SW{I think we do a bit more than this right? Once we obtain the sparse weights from the algorithm above, we further sparsifing through a backward pass? Add more details here and point to the example in figure 7}
\subsection{Predictions}\label{sec:prediction}
For a new $\bx_*$, the predictive distribution of $\by_*$ given the data is approximated as:
\begin{align}
    p(\by_* \mid \bx_*, \mathcal{D}) &= \int p(\by_* \mid \bx_*, \btheta, \mathcal{D}) p(\btheta \mid \mathcal{D}, \bx_* ) \, d\btheta \nonumber\\
    &=\int p(\by_* \mid \ba_*, \bW, \bb, \bmeta) p( \ba_*, \bW, \bb, \bmeta \mid \mathcal{D}, \bx_* ) \, d\ba_* \, d\bW \, d\bb \, d\bmeta \nonumber \\
    &\approx \int p(\by_* \mid \ba_*, \bW, \bb, \bmeta) q( \ba_*) q(\bW, \bb) q(\bmeta) \, d\ba_* \, d\bW \, d\bb \, d\bmeta \nonumber \\
    &= \int \Norm\left(\by_* \mid  \bW_{L+1} \ba_{*,L} + \bb_{L+1}, \bSigma_{L+1}\right) q( \ba_{*,L}) q(\bW_{L+1}, \bb_{L+1}) q(\bmeta_{L+1}) \, d\ba_{*,L} \, d\bW_{L+1} \, d\bb_{L+1} \, d\bmeta_{L+1}. \label{eq:pred}
\end{align}
 \cref{eq:pred} requires first computing the approximate variational predictive distributions $q( \ba_*), q( \bgamma_*)$ and $q( \bomega_*)$, which are updated in a similar way to \cref{sec:vb}. 
 
 Specifically, the stochastic activations are conditionally Gaussian given the previous layer with variational predictive distribution:
\begin{equation*}
    q(\ba_{*}) = \prod_{l=1}^{L} \Norm \left (\ba_{*,l} \mid  \mathbf{t}_{*,l}+ \mathbf{M}_{*,l} \ba_{*,l-1}, \mathbf{S}_{*,l} \right ), 
\end{equation*}
where $\ba_{*,0}= \bx_*$. For the final layer, we have:
$$ \mathbf{S}^{-1}_{*, L} = \hat{\mathbf{\Sigma}}_{L}^{-1}; \quad \mathbf{t}_{*,L} =  \E \left[ \bgamma_{*,L}\right ] \odot \E \left [ \bb_{L}\right ]; \quad  \mathbf{M}_{*,L} =  \E \left[ \bgamma_{*,L}\right ]  \mathbf{1}^T_{D_{L-1}} \odot \E \left [  \bW_{L}\right ].$$ For all other layers $l=1, \dots, L-1$, we have:
\begin{align*}
  \mathbf{S}^{-1}_{*, l}  =&  \hat{\mathbf{\Sigma}}_l^{-1} - \mathbf{M}_{*, l+1}^T \mathbf{S}_{*, l+1}^{-1}\mathbf{M}_{*, l+1} + \sum_{d=1}^{D_{l+1}}  \left ( \E \left [\frac{1}{\eta_{l+1,d}^2} \right ] \E \left[ \gamma_{*,l,d}\right ]+ \frac{1}{T^2}   \E \left [ \omega_{*,l+1,d}\right ]  \right) \E \left [ \bW_{l+1,d}^T \bW_{l+1,d} \right ], \\
  \mathbf{t}_{*, l}  = &  \mathbf{S}_{*,l} \left ( \mathbf{M}_{*, l+1}^T \mathbf{S}_{*, l+1}^{-1} \mathbf{t}_{*,l+1} + \hat{\mathbf{\Sigma}}_l^{-1}  \E \left[ \bgamma_{*,l}\right ] \odot  \E \left[ \bb_{l}\right ]  - \sum_{d=1}^{D_{l+1}}  \E \left [\frac{1}{\eta_{l+1,d}^2} \right ] \E \left[ \gamma_{*,l,d}\right ] \E \left [ \bW_{l+1,d}^T b_{l+1,d} \right ] \right. \\
& \left.+  \frac{1}{T} \sum_{d=1}^{D_{l+1}}\E \left [ \bW_{l+1,d}^T \right ]  \left ( \E \left [ \gamma_{*,l+1,d} \right ]  - \frac{1}{2}\right )  -  \frac{1}{T^2} \sum_{d=1}^{D_{l+1}}  \E \left [ \omega_{*,l+1,d}\right ]
  \E \left [ \bW_{l+1,d} b_{l+1,d}\right ] \right ), \\
  \mathbf{M}_{*,l} = & \mathbf{S}_{*,l} \hat{\mathbf{\Sigma}}_l^{-1} \E \left[ \bgamma_{*,l}\right ]  \mathbf{1}^T_{D_{l-1}} \odot \E \left [  \bW_{l}\right ] .
\end{align*} 

The binary activations are independent across layers $l=1, \ldots, L$ and width $d = 1, \ldots, D_l$, with a Bernoulli variational predictive distribution: 
\begin{align}
  q(\bgamma_{*})  &= \prod_{l}^{L} \prod_{d}^{D_l}\Bern\left (\gamma_{*,l,d} \mid \rho_{*,l,d}  \right ), \label{eq:vbgammapred}
  \end{align}
  with 
  \begin{align*}
  \rho_{*,l,d} =    &  \sigma \left (-\frac {\E \left [ \eta_{l,d}^{-2} \right ] }{2}   \Tr \left (\E \left[\widetilde{\bW}_{l,d}^T \widetilde{\bW}_{l,d}\right ] \E \left [\widetilde{\ba}_{*, l-1} \widetilde{\ba}_{*, l-1}^T \right ] \right )  + \E \left [ \eta_{l,d}^{-2} \right ] \E \left [ \widetilde{\bW}_{l,d} \right ] \E \left [\widetilde{\ba}_{*, l-1} a_{*,l,d} \right ] + \frac{1}{T} \E \left [  \widetilde{\bW}_{l,d}  \right ] \E \left [\widetilde{\ba}_{*, l-1}\right ] \right).
\end{align*}

Finally, the Polya-Gamma augmented variables are independent across layers $l=1,\ldots, L$ and width $d=1, \ldots, D_l$, with a Polya-Gamma variational predictive distribution: 
\begin{align}
q(\bomega_*) &=  \prod_{l}^{L} \prod_{d}^{D_l} \PG(  \omega_{*,l,d} \mid 1, A_{*,l,d}), \label{eq:vbomegapred} 
\end{align}
with updated variational parameters: 
\begin{align*}
A_{*,l,d} &= \frac{1}{T} \sqrt{\left ( \Tr \left ( \E \left [ \widetilde{\bW}_{l,d}^T \widetilde{\bW}_{l,d} \right ] \E \left [ \widetilde{\ba}_{*,l-1} \widetilde{\ba}_{*,l-1}^T\right ] \right ) \right )}.
\end{align*}

Thus, before computing predictions, we first iterate to update the variational predictive distributions of $\ba_*$, $\bgamma_*$, and $\bomega_*$. The corresponding ELBO (derived in the \cref{sec:ELBOforpred}) is monitored for convergence. 
While a closed-form expression for the integral in \cref{eq:pred} is unavailable, generating samples from the variational predictive is straightforward;
  \begin{align}
       \by_*^{(j)} \sim \Norm\left(\by_* \mid  \bW_{L+1}^{(j)} \ba_{*,L}^{(j)} + \bb_{L+1}^{(j)}, \bSigma_{L+1}^{(j)}\right),
       \label{eq:pred_samp}
   \end{align} 
 for $j=1, \ldots J$,  where $(\bW_{L+1}^{(j)},\bb_{L+1}^{(j)}) \sim q(\bW_{L+1}, \bb_{L+1})$,  $\bmeta_{L+1}^{(j)} \sim q(\bmeta_{L+1}) $, and $\ba_{*,l}^{(j)} \mid \ba_{*,l-1}^{(j)} \sim q(\ba_{*,l} | \ba_{*,l-1})$ for $l=1, \ldots, L$, are iid draws from the variational posterior.  These samples can be used to obtain a    Monte Carlo approximation to investigate potential non-normality in the predictive distribution in \cref{eq:pred} and to compute credible intervals based on the highest posterior density region.
%In addition to computing the mean and variance of $\by_*$, we can use a Monte Carlo approximation to investigate potential non-normality in the predictive distribution in \cref{eq:pred}. Specifically, the predictive distribution is approximated as:  
 %  \begin{align*}
 %      p(\by_* \mid \bx_*, \mathcal{D}) \approx \frac{1}{J} \sum_{j=1}^J \Norm\left(\by_* \mid  \bW_{L+1}^{(j)} \ba_{*,L}^{(j)} + \bb_{L+1}^{(j)}, \bSigma_{L+1}^{(j)}\right),
%   \end{align*} 
   % where $(\bW_{L+1}^{(j)},\bb_{L+1}^{(j)}) \sim q(\bW_{L+1}, \bb_{L+1})$,  $\bmeta_{L+1}^{(j)} \sim q(\bmeta_{L+1}) $, and $\ba_{*,l}^{(j)} \mid \ba_{*,l-1}^{(j)} \sim q(\ba_{*,l} | \ba_{*,l-1})$ for $l=1, \ldots, L$, are iid draws from the variational posterior. We can also compute more accurate credible intervals (avoiding a normality assumption of the predictive distribution) by sampling new values of the target variables 
   % \begin{align}
   %     \by_*^{(j)} \sim \Norm\left(\by_* \mid  \bW_{L+1}^{(j)} \ba_{*,L}^{(j)} + \bb_{L+1}^{(j)}, \bSigma_{L+1}^{(j)}\right),
   %     \label{eq:pred_samp}
   % \end{align} 
   % for $j=1, \ldots J$, then computing the highest posterior density region from those samples. 

We can also  compute the expectation and variance of $\by_*$ in closed form. Specifically, the expectation of $\by_*$ under the variational predictive distribution is:
 \begin{align}
    \E[\by_* \mid \bx_*, \mathcal{D}] &\approx \E_{q_{L+1}}[\bW_{L+1}]\E_{q_{*,L}} [\ba_{*,L}] + \E_{q_{L+1}}[\bb_{L+1}], \label{eq:pred_exp} 
\end{align}
where recursively
 \begin{align*}
   \E_{q_{*,L}} [\ba_{*,L}] &=  \E_{q_{*,L-1}} \left [ \E_{q(\ba_{*,L} | \ba_{*,L-1})}[\ba_{*,L}] \right ] = \mathbf{t}_{*,L}+ \mathbf{M}_{*,L}  E_{q_{*,L-1}}[\ba_{*,L-1}].
\end{align*}
 Similarly, the variational variance of $\by_*$ is 
 \begin{align*}
     & \Var \left[y_{*, d}  \mid \bx_*, \mathcal{D}\right]  \approx  \Var_{q_{L+1}} \left[\bW_{L+1,d}\ba_{*,L} + \bb_{L+1,d}\right] + \E_{q_{L+1}} \left[\bmeta_{L+1,d}^{2}\right], 
  \end{align*}
  where the first term represents the signal variance and is computed as
\begin{align*}
 &\Var_{q_{L+1}} \left[\bW_{L+1,d} \ba_{*,L}  + \bb_{L+1,d} \right]   =\E_{q_{L+1}} \left[(\bW_{L+1,d}\ba_{*,L} + \bb_{L+1,d})^2\right] -
     \left(\E_{q_{L+1}}\left[\bW_{L+1,d} \right] \E_{q_{L}}\left[\ba_{*,L} \right] + \E_{q_{L+1}}\left[\bb_{L+1,d}\right]\right)^{2}   \\
    &= \Tr \left( \E_{q_{L+1}} \left[\bW_{L+1,d}^T\bW_{L+1,d}\right]\E_{q_{L}} \left[\ba_{*,L} \ba_{*,L}^T\right] - \E_{q_{L+1}} \left[\bW_{L+1,d}\right]^T\E_{q_{L+1}} \left[\bW_{L+1,d}\right]\E_{q_{L}} \left[\ba_{*,L}\right] \E_{q_{L}} \left[\ba_{*,L}\right]^T\right)  \\
   &  +\Var_{q_{L+1}}(\bb_{L+1,d}) + 2\text{Cov}_{q_{L+1}}(\bW_{L+1,d}, \bb_{L+1,d}) \E_{q_{L}}\left[\ba_{*,L} \right],  
  \end{align*}
which requires the recursive computation:
\begin{align*}
 &  \E_{q_{L}}  \left[\ba_{*,L} \ba_{*,L}^T\right] = \bS_{*, L} + \bt_{*, L}  \bt_{*, L}^T +  2\bM_{*,L} \E_{q_{L-1}} \left[\ba_{*,L-1} \right ] \bt_{*,L}^T + \bM_{*,L} \E_{q_{L-1}} \left[\ba_{*,L-1} \ba_{*,L-1}^T \right ] \bM_{*,L}^T. %, \\
%& \E_{q_{L+1}}  \left[\eta_{L+1,d}^{2}\right]= \frac{\beta_{L+1,d}}{\alpha_{L+1,d}-1}, 
  \end{align*}
% \SW{ I think the two middle terms equal? Can we write $2  \bM_{*,L} \E_{q_{L-1}} \left[\ba_{*,L-1} \right ] \bt_{*,L}^T$? } \AS{yes!}
%where variational parameters of $\bW_{L+1}$,  $\bb_{L+1}$  and $\bmeta_{L+1}$ are obtained from the  algorithm described in  \cref{sec:vb}.

\paragraph{Sparse prediction} \blu{Observe that the variational algorithm used for prediction scales linearly with the number of hidden layers and the number of samples but cubically with the number of hidden units, this motivates the node-selection method proposed in \cref{sec:selectvariables}.} To save on both computation and storage, the variational predictive distribution can be computed based on the sparse variational posterior (\cref{sec:selectvariables}). 
%In \cref{sec:selectvariables} we obtained sparse weights $\widehat{W}_{l,d,d'}$ and the corresponding sparse variational distribution
%$\widehat{q}(b_{l,d}, \widehat{\bW}_{l,d})$ for some
%$ \; l \in  \mathcal{L} \subseteq \{1, \ldots, L+1\}, \; d \in \mathcal{D}_l \subseteq \{ 1, \ldots, D_l\}, d' \in \mathcal{D}_{l-1} \subseteq \{ 1, \ldots,  D_{l-1}\}$. 
For a new data point $\bx_*$,  we obtain expectation and variance of $\by_*$ by first computing the sparse versions of variational predictive distributions $\widehat{q}( \ba_*), \widehat{q}( \bgamma_*)$ and $\widehat{q}( \bomega_*)$ 
as in \cref{eq:vbomegapred,eq:vba,eq:vbgammapred} by plugging $\widehat{q}(b_{l,d}, \widehat{\bW}_{l,d})$ instead of the $q(b_{l,d}, \bW_{l,d})$, which only requires updates for the subset of nodes with nonzero weights.

 \subsection{Ensembles of Variational Approximations}\label{sec:ensembles}

 % To obtain optimal contraction rates  \citep{zhang2020convergence} develop a method of choosing "the best" variational approximation (\AS{need to check}). 

While the variational algorithm described in \cref{sec:vb} increases the ELBO at each epoch, the ELBO is a non-convex function of the variational parameters and only convergence to a local optimum is guaranteed. Due to identifiability issues, the posterior distribution of a Bayesian neural network is highly multimodal, and exploring this posterior is notoriously challenging \citep{papamarkou2022challenges}. A single variational approximation tends to concentrate around one mode and can understate posterior uncertainty. 
Several approaches have been proposed to overcome such issues. Recently, \citep{ohn2024adaptive} introduced adaptive variational inference, which achieves optimal posterior contraction rate and model selection consistency by considering several variational approximations obtained in different models. Similarly, \citep{yao2022stacking} introduced an approach which uses parallel runs of inference algorithms to cover as many modes of the posterior distribution as possible and then combines these using Bayesian stacking.\blu{ While the idea of combining the outputs of several neural networks is not novel \citep{Hansen1990, Levin1990}, we note that deep non-Bayesian ensembles recently received quite a lot of attention \citep{LakshminarayananDeepEns} and several alternatives and modifications enabling uncertainty quantification were proposed  
\citep{wilson2020bayesian,angelo2021,pearce2020uncertainty,wenzel2020hyperparameter}. 
 % a scalable method for constructing a weighted average of distributions. 
Observing non-optimality of conventional deep ensembles combining point estimates \citep{wu2024posterior}, in our approach, we adopt the ideas of \citep{ohn2024adaptive, yao2022stacking} and consider ensembles of posterior approximations 
in a similar but simpler fashion. 
Specifically, we consider an ensemble of variational approximations, obtained by running in parallel the variational algorithm multiple times with different random starting points and combining those with respect to the optimization objective.} In this case, letting $k=1,\ldots, K$ index the different variational approximations, we compute the weight $w_k$ associated with each approximation in accordance with the tempered ELBO:
\begin{align*}
    w_k \propto \exp \left (\zeta \text{ELBO}_k\right), 
\end{align*}
\blu{where $\zeta$ is a tempering parameter, setting which to be in the interval $(0, 0.1]$ allows for avoiding a strong dominance of a single particular model.}  We can interpret this as a Bayesian model averaging (BMA) across the $K$ different models/approximations. While in a classical BMA setting, the weights would be proportional to the marginal likelihood for each model, the use of the ELBO is motivated as it provides a lower bound to the marginal likelihood and can be computed in closed form.  
Next, we compute predictions by taking a weighted average of the predictive distributions of each model (given in  \cref{eq:pred_exp}), that is
\begin{align*}
 \E[\by_* \mid \bx_*, \mathcal{D}] \approx  \sum_{k=1}^K  w_k \E_{q_k}[\by_* \mid \bx_*, \mathcal{D}],
\end{align*}
where each expectation is taken with respect to $q_k$ (the $k$th variational approximation). 
Similarly, we can compute the variance as
\begin{align*}
 \Var(\by_* \mid \bx_*, \mathcal{D}) \approx  \sum_{k=1}^K  w_k \Var_{q_k}(\by_* \mid \bx_*, \mathcal{D}) + \sum_{k=1}^K  w_k \left(\E_{q_k}[\by_* \mid \bx_*, \mathcal{D}]\right)^2 -  \left(\sum_{k=1}^K  w_k \E_{q_k}[\by_* \mid \bx_*. \mathcal{D}]\right)^2.
\end{align*}
The following approach has the potential to improve both predictive accuracy and uncertainty quantification.
Once again, we can investigate the variational predictive distribution (beyond the mean and variance) by first sampling a model with probability $(w_1,\ldots, w_K)$ and then given that selected model $k$, generating a sample $\by_*$ from the $k$th variational predictive distribution (as described in \cref{eq:pred_samp}). 

\section{Experiments} \label{sec:experiments}

%\SW{Start with a high-level overview and motivation for each of the experiments}
We evaluate the variational bow tie neural network (VBNN) on several datasets. First, we consider a simple nonlinear synthetic example to compare with a ground truth.  
We then validate VBNN on the \cmsspy{diabetes} dataset, first considered in \citep{efron2004least} to demonstrate the least angle regression (LARS) algorithm for variable selection, and subsequently, used in different proposals for sparsity-promoting priors algorithms (e.g. \citep{parkcasellalasso, LiLin2010}). Lastly, we consider a range of popular regression datasets from the UCI Machine Learning Repository \citep{UCIMLrep}.

The importance of suitable initialization choice in NNs is well known \citep{wenzel2020goodbayesposteriordeep, Daniely2016init, he2015delvingdeeprectifierssurpassing}, and we design two possible random initialization schemes of the VBNN, which are described in \cref{sec:initschemes} and used in all experiments. 
Convergence of the ELBO is monitored during the training and prediction stages, where if three consecutive measurements of ELBO for training differ by less than the specified threshold, the phase is stopped and the model moves to the prediction stage, where we proceed similarly. In most experiments, the thresholds during the training and prediction stages are set, respectively, to $1e-5$ and $1e-4$.  %Throughout this section, our model is referred to as VBNN standing for variational bow tie neural network. 

\blu{We compare the performance of VBNN to various frameworks (summarized in \cref{tab:models}), namely, to BNNs inferred with
 automatic differentiation VI with the mean-field variational family (mfVI) \citep{kucukelbir2017automatic} and Hamiltonian Monte Carlo with the No-U-Turn sampler (HMC) \citep{hoffman2014no}  implemented in \cmsspy{Numpyro} \citep{numpyrophan2019composable}, Bayes by Backprop (BBB) \citep{blundell2015weight} implemented with \cmsspy{Pytorch}, and BNNs with Horseshoe priors (HSBNN) of \citep{ghosh2019} which considers a structured variational family and learns the variational parameters by obtaining gradient estimators. 
 We also consider the variational bow tie neural network with Gaussian priors (GVBNN), in contrast to shrinkage priors. 
 As CAVI may be expensive for large data, in \cref{sec:UCIdatasetsap} we additionally consider VBNN inferred with SVI (SVBNN). Further comparisons between the performance of VBNN and SVBNN are provided in  \cref{ap:SVIsupexperiment}.} %To ease the navigation across the compared methods, we provide a brief description of the models in \cref{tab:models}.}
\begin{table}[t!]
\caption{List of the models considered to evaluate the performance of our method.} \label{tab:models}
\centering
\begin{tabular}{l|c} 
Model & Description                                 \\
\hline
mfVI  & BNN with Automatic Differentiation VI with mean-field family                   \\
HMC   & BNN with Hamiltonian Monte Carlo with No-U-Turn sampler \\
BBB   & Bayes by Backprob                             \\
HSBNN & BNN with Horseshoe priors inferred with Black Box VI  \\
VBNN  & Our model inferred with CAVI                  \\
GVBNN & VBNN with Gaussian priors inferred with CAVI \\
SVBNN & Our model inferred with SVI                    
\end{tabular}
\end{table}

For all the datasets, we evaluate the performance over 10 random splits, where we use $90\%$ of the data for the training and $10\%$ for testing the model. %We normalize the input but do not re-scale the output. 
We record the root mean squared error (RMSE), the predictive negative log-likelihood of the test data (NLL) and the empirical coverage (EC) (see \cref{sec:experimentsap} for additional implementation details).
\blu{
The empirical coverage reports the fraction of observations contained within the $(1-\alpha)*100\%$ credible intervals (CIs), which are computed based on the Gaussian approximation. In the ideal setting, the computed EC should equal the CI level.   %and compute empirical coverage as a 
%the fraction of observations contained within the CI; this means that in the ideal settings the computed EC should equal $1-\alpha$. Namely, 
More specifically, let $\by^*_i$ be the true target for test points $i=1,\ldots, N^*$ and %the predicted observations of the model $\by^{o}$, one can consider EC as a function of confidence interval level. If 
$q_{\alpha/2}$ and $q_{1-\alpha/2}$ be the quantiles based on the model's Gaussian approximation for some $\alpha \in [0.5, 1]$,  then
\begin{align*}
    \text{EC}(\alpha)& =\frac{1}{N^*} \sum_{i=1}^{N^*} \1(\by^{*}_i \in [q_{\alpha/2}, q_{1-\alpha/2}]).  
\end{align*} 
 If $\text{EC}(\alpha) >1-\alpha $ then the CIs are too wide; a worse scenario occurs when $\text{EC}(\alpha) < 1-\alpha $ as it means that the CIs are too narrow and the model is overconfident in its predictions. 
}
\subsection{Simulated Example}\label{sec:toy}
\begin{figure}[ht]
    \centering
        \includegraphics[width=0.8\linewidth]{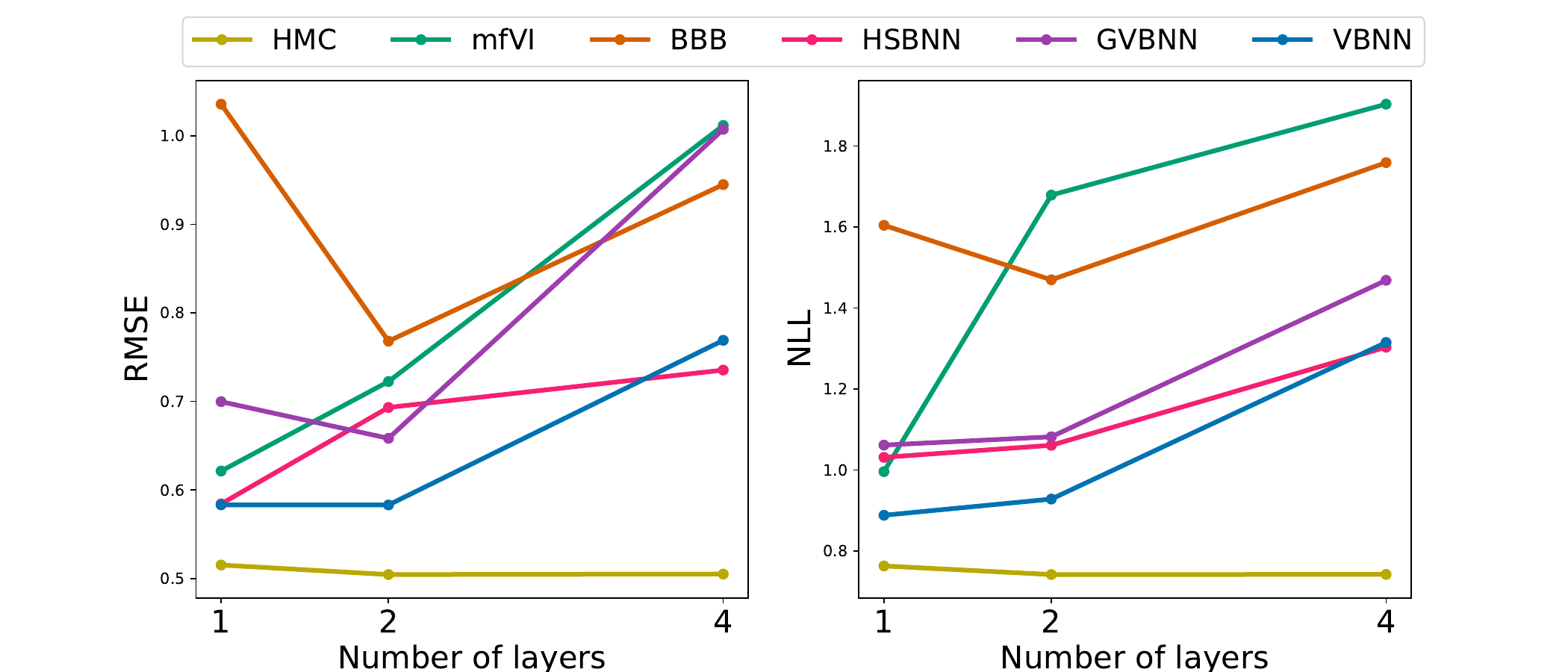}
    \caption{\blu{Simulated example. Performance in terms of the RMSE and NLL as the depth increases for different models and algorithms. HMC can be seen as a gold standard. VBNN is competitive with HSBNN and is more robust to the choice of depth and overparameterization than GVBNN, mfVI, BBB.}}
    \label{fig:rmsenlltoy}
\end{figure}

\begin{figure}[ht]
    \centering
     \includegraphics[width=0.9\linewidth]{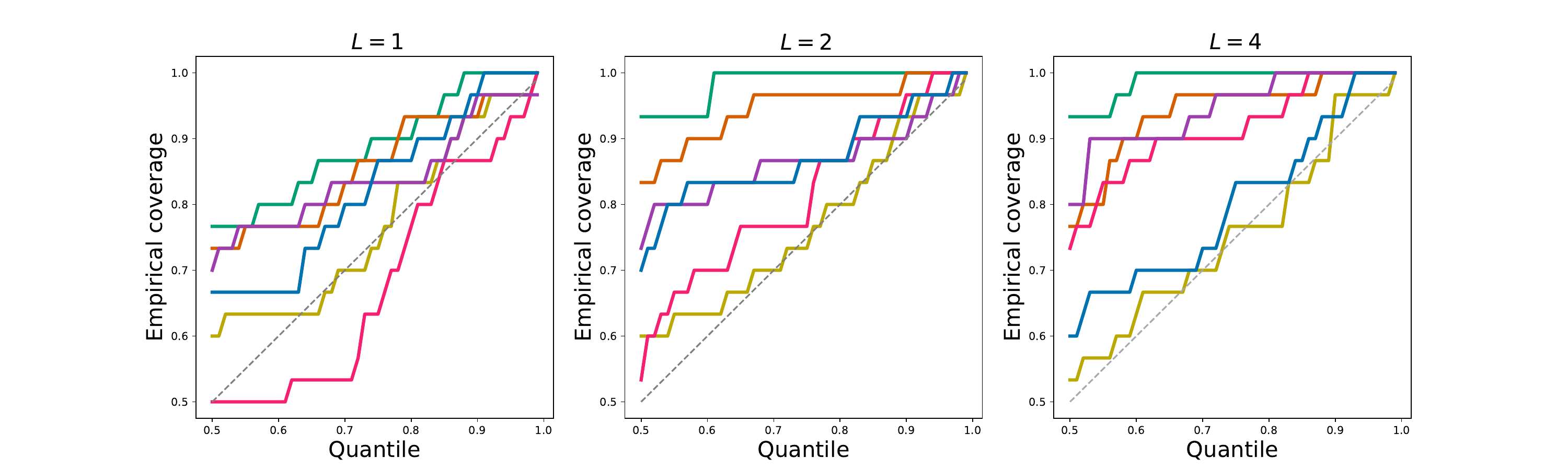}
    \caption{  \blu{Simulated example.  Empirical coverage (which is the fraction of observations contained within the CIs of level $1-\alpha$) as a function of CI level  for the simulated dataset for three different settings of the network's depth. The dashed gray line depicts the ideal scenario with empirical coverage equal to CI level, while above and below the gray line indicate coverage greater or less than CI level, i.e. CIs are too wide or too small, respectively.}
   } \label{fig:ectoy}
\end{figure}

We construct a synthetic dataset generated by first uniformly sampling a two-dimensional input vector $\bx_n = (x_{n,1}, x_{n,2})$, with $x_{n,d} \sim \Unif([-2,2])$, and assume only the first feature influences the output: $y_n  = f(x_{n,1}) + \epsilon_n = 0.1 x_{n,1}^2 + 10\sin(x_{n,1}) + \epsilon_n$, where $\epsilon_n \sim \Norm(0, 0.5)$. Then, the dataset consisting of  $N = 300$ observations is used to investigate the performance of VBNN compared to the \blu{GVBNN, mfVI, HMC}, BBB and \blu{HSBNN} baselines as we increase the number of hidden layers, setting $ L=1, 2 $ or $4$, whilst keeping the number of hidden units in each layer fixed to $D_H=20$. 
% \cref{tab:toymetrics} reports the root mean squared error (RMSE), the predictive negative log-likelihood of the test data (NLL) and the empirical coverage for three different cases: 
 In general, for this simple non-linear example, the performance tends to deteriorate with increasing architecture complexity (larger depth). \blu{While HMC performs consistently well across all depths, the cost associated with the sampling approach is high. VBNN is competitive to HSBNN and outperforms mfVI, GVBNN and BBB in terms of accuracy (see \cref{fig:rmsenlltoy}). Further, except for HMC, the empirical coverage of VBNN is the most robust to the choice of depth; for the largest choice of $L=4$, mfVI, BBB, GVBNN and even HSBNN provide overly wide CIs while VBNN more closely reaches the desired coverage (see \cref{fig:ectoy}).}

\begin{figure}[ht]
    \centering
     \subcaptionbox{The predictive mean as a function of the second coordinate for $L=1$. \label{subfigureforx2}}[0.2\textwidth]{\includegraphics[width=0.2\textwidth]{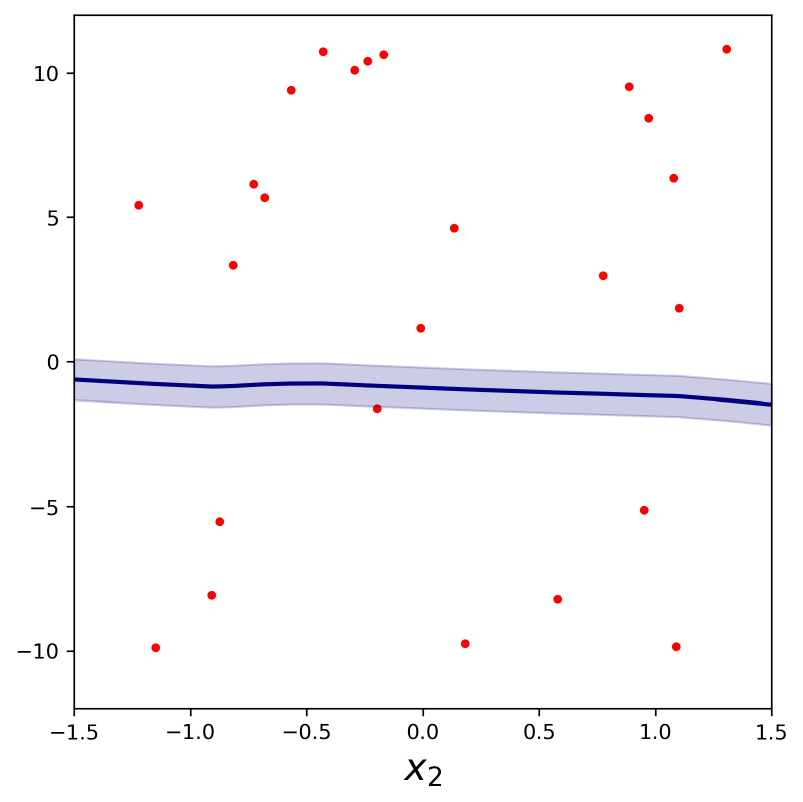}} 
   \subcaptionbox{The predictive mean and uncertainty quantification for the observations for different depth \label{subfigureforx1}}[0.79\textwidth]{\includegraphics[width=0.78\textwidth]{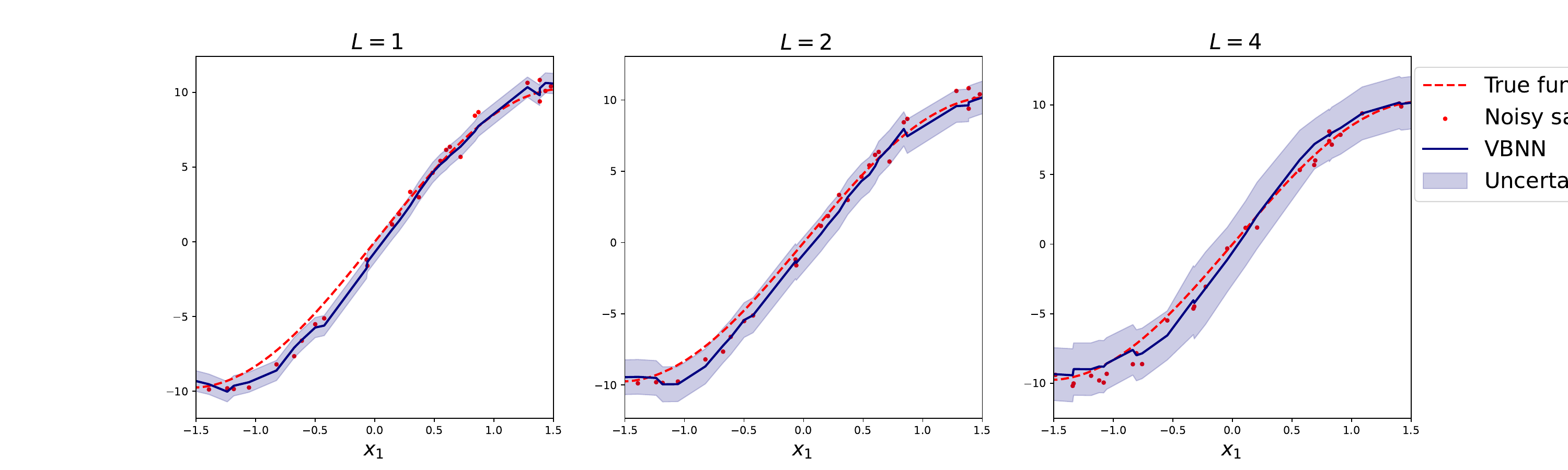}} 
   \subcaptionbox{The architecture of the network for the bound on the FDR $\alpha = 0.01$ for depths $L=1,2,4$ (left to right). \label{subfigureforstructure}}[0.9\textwidth]{\includegraphics[width=0.23\textwidth]{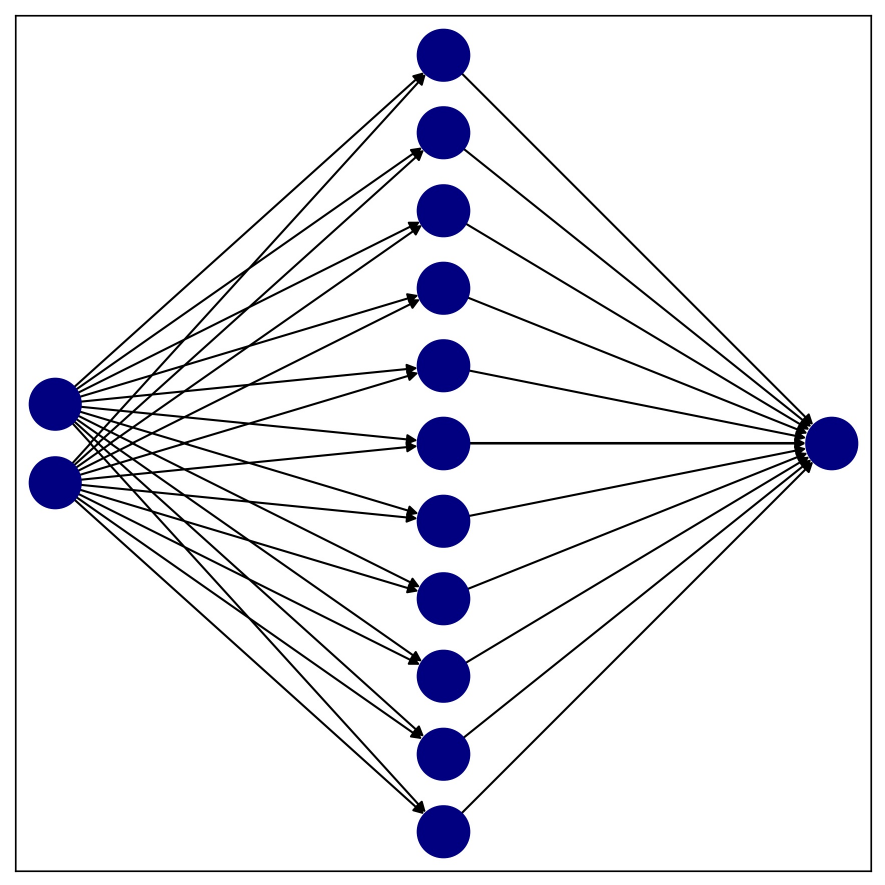} \includegraphics[width=0.23\textwidth]{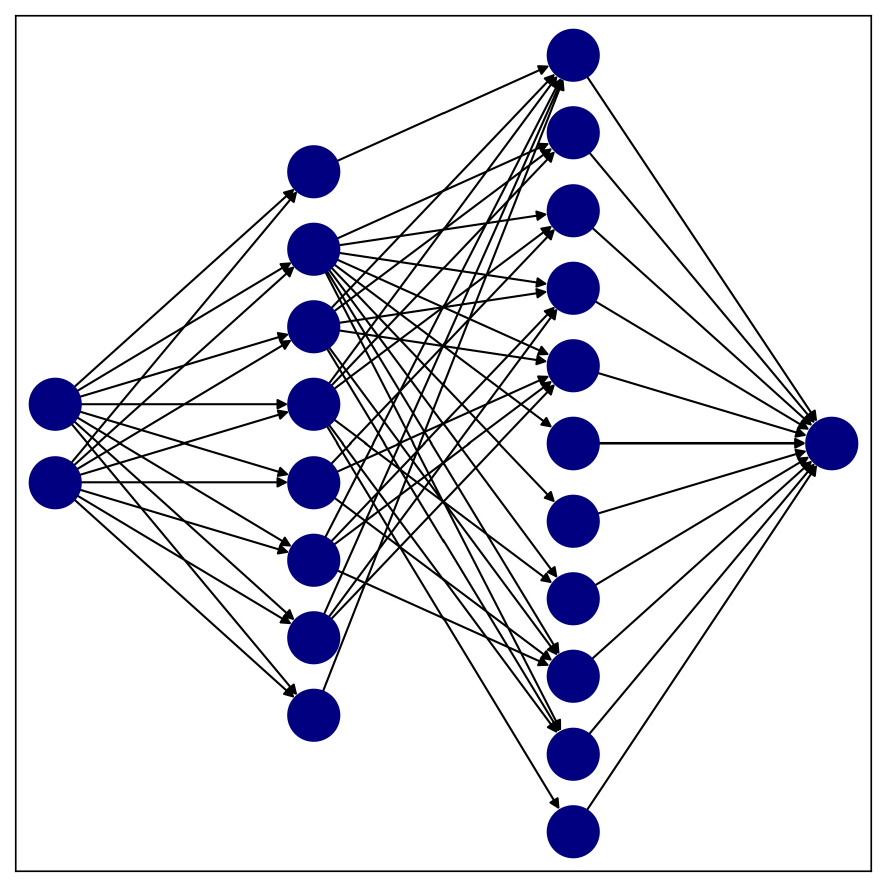} \includegraphics[width=0.3\textwidth]{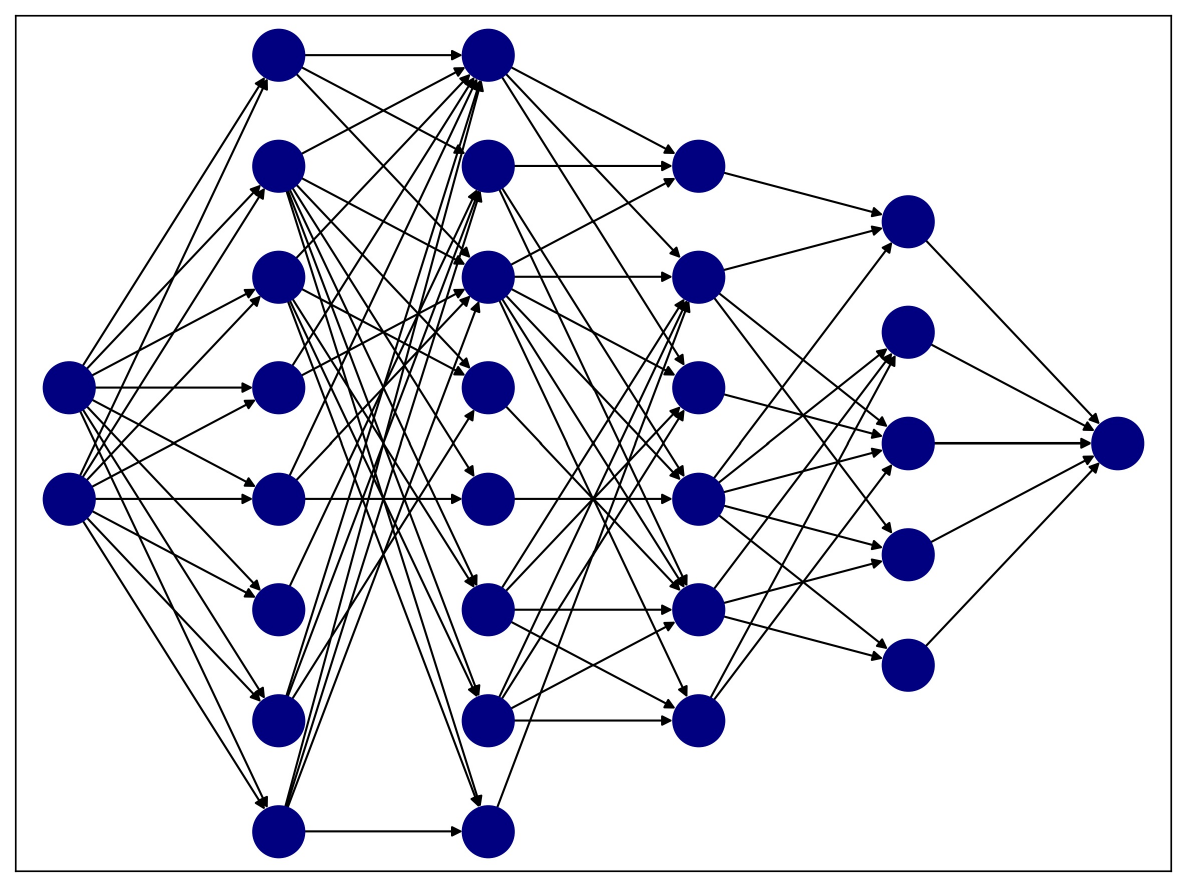}}
   \caption{Simulated example. Predictive means and pointwise CIs computed for the observations as a function of the second coordinate (a) and first coordinate for different depths (b). The architecture of the network is visualized in (c) for the bound on the FDR $\alpha = 0.01$ for different settings of the network's depth of $L=1,2,4$ (left to right).}
    \label{fig:toyexample}
\end{figure}

For each depth, \cref{fig:toyexample} illustrates the predictive means and uncertainties computed for the observations as well as DAGs of networks' structures obtained after the post-process node selection algorithm described in \cref{sec:selectvariables}, where the Bayesian false discovery rate is constrained by setting the error rate to $\alpha = 0.01$. The sparsity-promoting prior combined with the node selection algorithm can effectively prune the over-parametrized neural networks; for example, the sparse one-layer neural network contains only $11$ hidden nodes with $33$ total edges/weights from the initial $D_H= 20$ with $60$ total edges/weights. Moreover, the estimated regression function and credible intervals both from the variational predictive and the sparse variation predictive distribution, recover the true function well. In this way, VBNN provides an effective tool to reduce predictive computational complexity and storage as well as ease interpretation. Note that the predictions show no relation with the coordinate $x_2$ (\cref{subfigureforx2}), recovering the true function, but some of the connections from $x_2$ are still present in the sparse network (\cref{subfigureforstructure}), due to identifiability issues, although with overall low weight.

\subsection{Diabetes Example} \label{sec:diabetesex}
\begin{figure}[ht]
    \centering 
    \includegraphics[width = 0.9\linewidth]{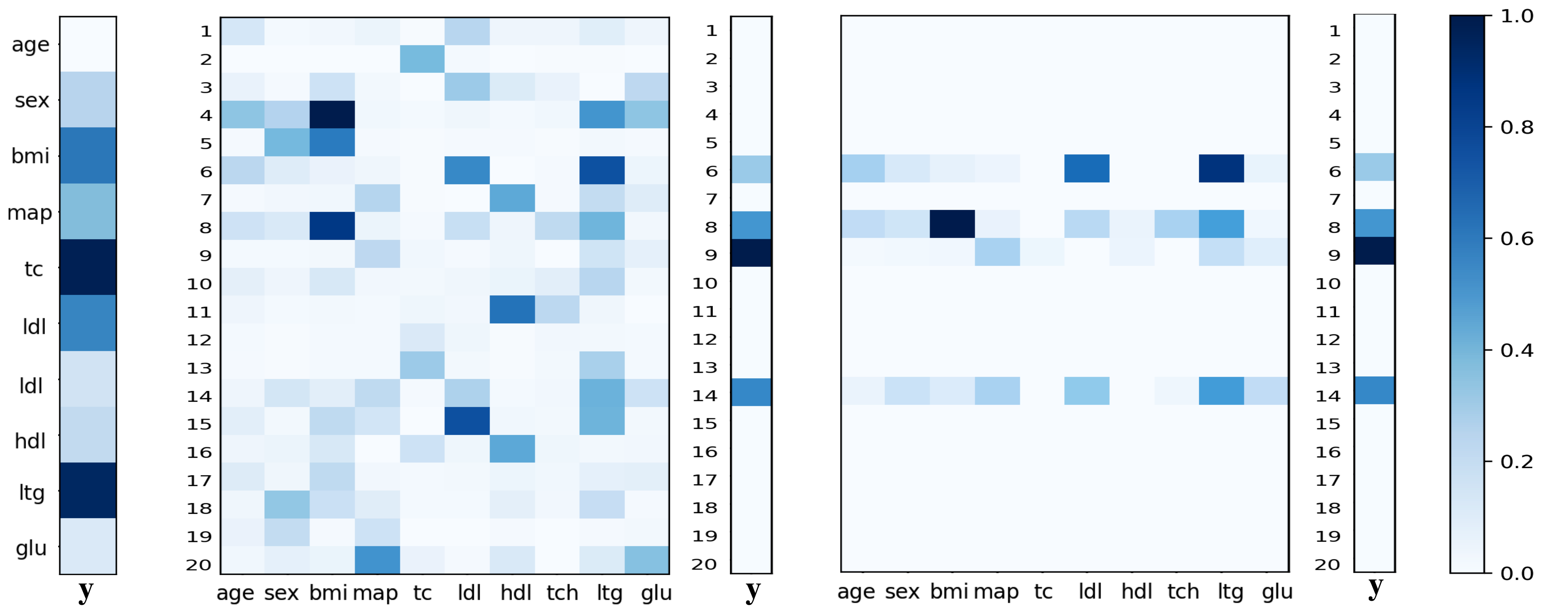}
    \caption{Diabetes example. Coefficients of LassoCV regression (on the left), 
    posterior means of the weights of the neural network (in the middle) and posterior means of the sparse weights obtained for $\alpha = 0.01$  (on the right). \blu{For illustrative purposes, absolute values of the coefficients and weights are shown with max-min scaling.}}
    \label{fig:sparsity}
\end{figure}

\begin{figure}[ht!]
    \centering
    \includegraphics[width=0.9\linewidth]{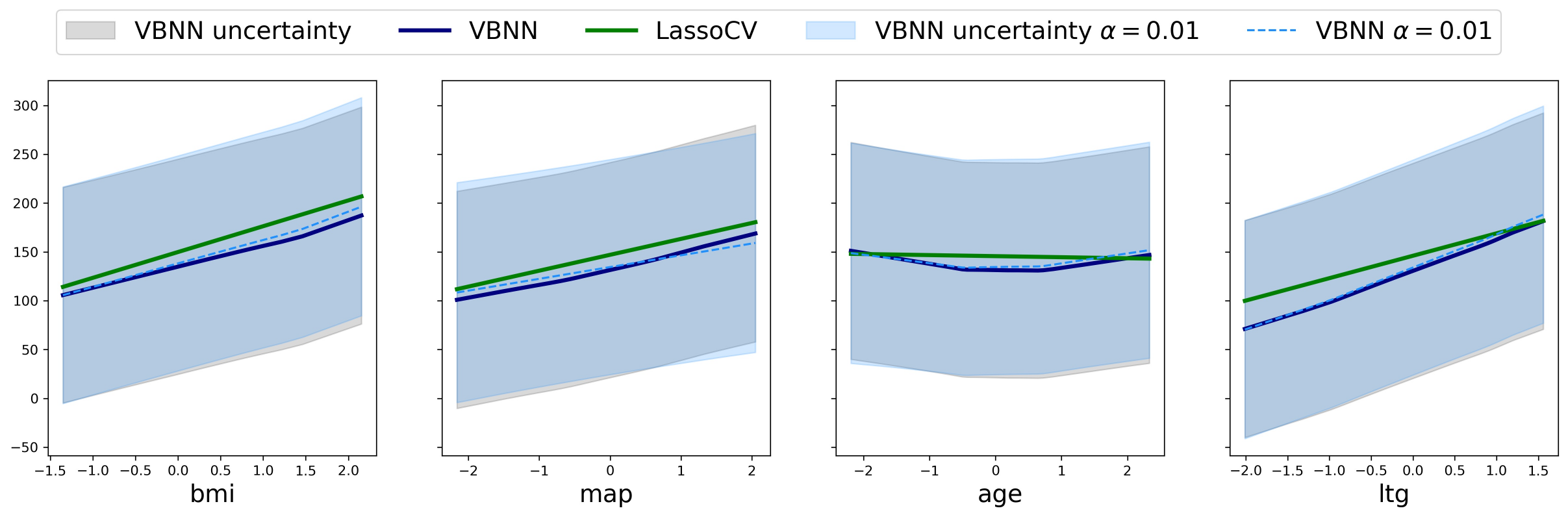}
    \caption{Diabetes example. Slices of the predictive mean and pointwise credible intervals for observations as a function of four predictors obtained by VBNN with and without node selection and by Lasso with cross-validation. 
    }
    \label{fig:vbnnvslasso}
\end{figure}

 The \cmsspy{diabetes} data consists of $n=442$ entries obtained for $p=10$ input variables and a quantitative response measuring disease progression. The predictors are age, sex, body mass index, average blood pressure and six blood serum measurements and the goal is to determine which of these are relevant for forecasting diabetes progression.
We fit a neural network with one hidden layer $L=1$ and $D_H=20$ and perform the node selection algorithm with the FDR bounded by $\alpha = 0.01$. \cref{fig:sparsity} illustrates the shrinkage and node selection algorithm and compares the coefficients of the Lasso linear model \blu{\citep{Tibshiranilasso}} to the original and the sparsified weights of our model. \blu{Lasso regression produces sparse coefficients by minimising the residual sum of squares with an added penalty term; the penalty parameter crucially determines the level of sparsity and is tuned with cross-validation (LassoCV).}  
%in this way, the sum of the absolute values of the coefficients is limited by a fixed constant.} 
Predictors with considerable effect obtained by both models coincide, whilst some of the variables the Lasso model excludes (e.g. age) are still present in the VBNN's estimates.  
Compared with Lasso, \blu{VBNN has the advantage} of learning potential nonlinear relationships between disease progression and the predictors, which is explored in  \cref{fig:vbnnvslasso}, illustrating 
the predictive means and uncertainty of the observations of VBNN for four of the predictors (with all other predictors are fixed to their mean). While the uncertainty is wide, the results suggest potential nonlinear relationships, e.g. with lamotrigine and age, the latter of which is not selected in Lasso.   
%linear model on \cref{fig:vbnnvslasso}, where we consider three of the predictors with considerable influence (body mass index, mean arterial pressure and lamotrigine) and a predictor effect of which in Lasso model is rather small (age). 
Moreover, \cref{fig:vbnnvslasso} highlights how predictions obtained from the sparse version of the variational predictive distribution almost overlap, thus providing a reasonable, cheaper approximation. 
\blu{However, we note that the predictive performance is similar to LassoCV, with the most competitive methods being VBNN, HMC and BBB (see \cref{tab:diabetespreformance} and supplementary \cref{fig:maindiabetespredictors} in \cref{sec:supplfig}).}
\begin{table}[ht]
    \centering 
    \caption{RMSE, NLL and empirical coverage for \cmsspy{diabetes} dataset. }
    \label{tab:diabetespreformance}
    \begin{tabular}{ l c c c}
    \hline
& RMSE & NLL &  Coverage  \\
\hline
LassoCV & $54.2 \pm 6.5$  & $5.4 \pm .13$ & $.96 \pm .03$ \\
\blu{mfVI} & $ 57.2 \pm 7.4 $& $ 9.3 \pm 1.7 $ & $ .47\pm .1$  \\
\blu{HMC}  & $ 54.5 \pm 7.8 $& $ 5.4 \pm .16 $ & $ .96\pm .04$ \\
BBB & $ 54.9 \pm  7.3 $ & $ 5.49 \pm .18 $ &  $ .94\pm .04 $  \\
\blu{HSBNN }& $ 56.8 \pm  7.4 $ & $ 6.8 \pm 1.6 $ &  $ .67\pm .15 $  \\
\blu{GVBNN  }& $ 55.65 \pm 7.8$ &$ 5.5 \pm .14 $ & $.96 \pm.04 $\\
VBNN & $ 54.5 \pm 7.2$ & $ 5.4 \pm .15 $ & $.96 \pm.04 $ \\
 \hline
\end{tabular}
\end{table}

\subsection{UCI Regression Datasets} 
\begin{figure}[ht!]
    \centering
\includegraphics[width=0.85\linewidth]{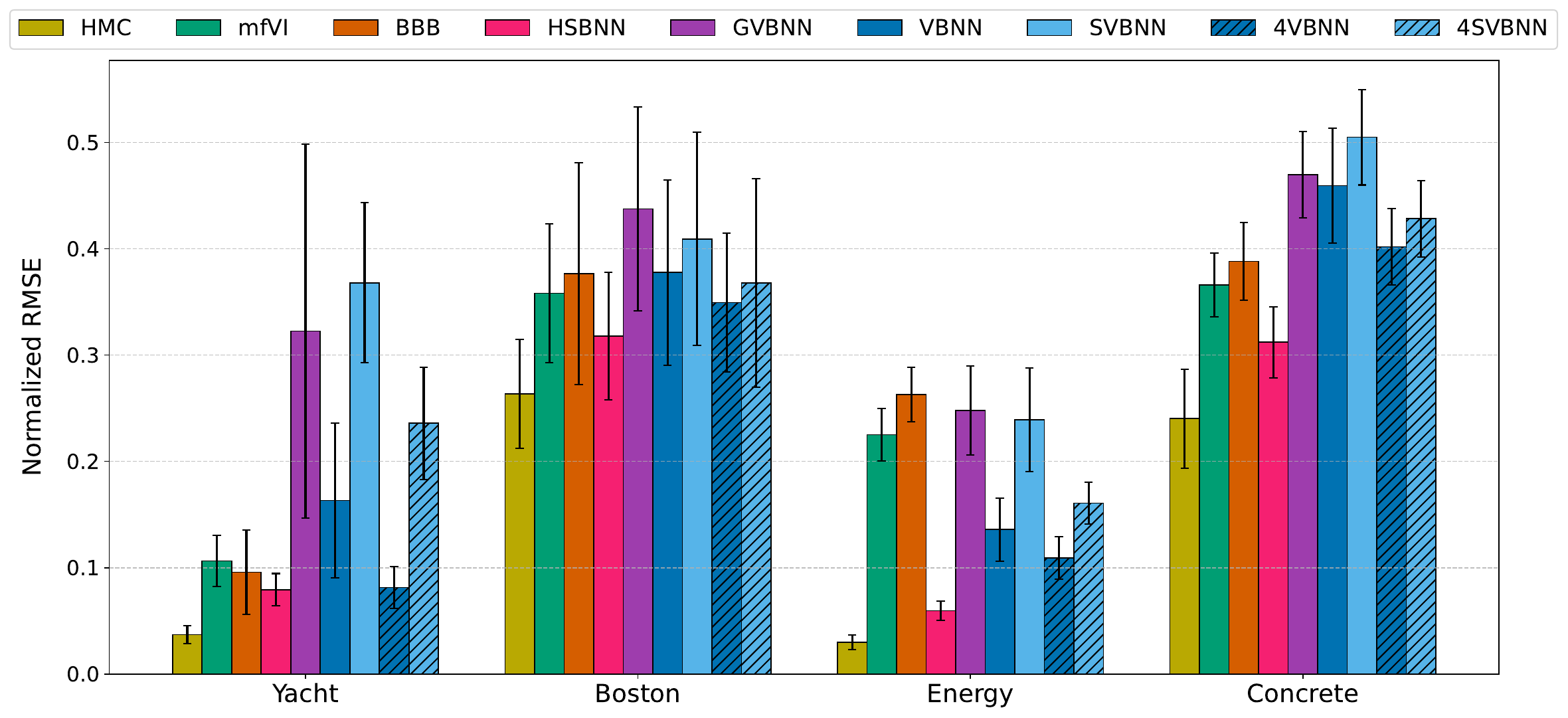}\\
\includegraphics[width=0.8\linewidth]{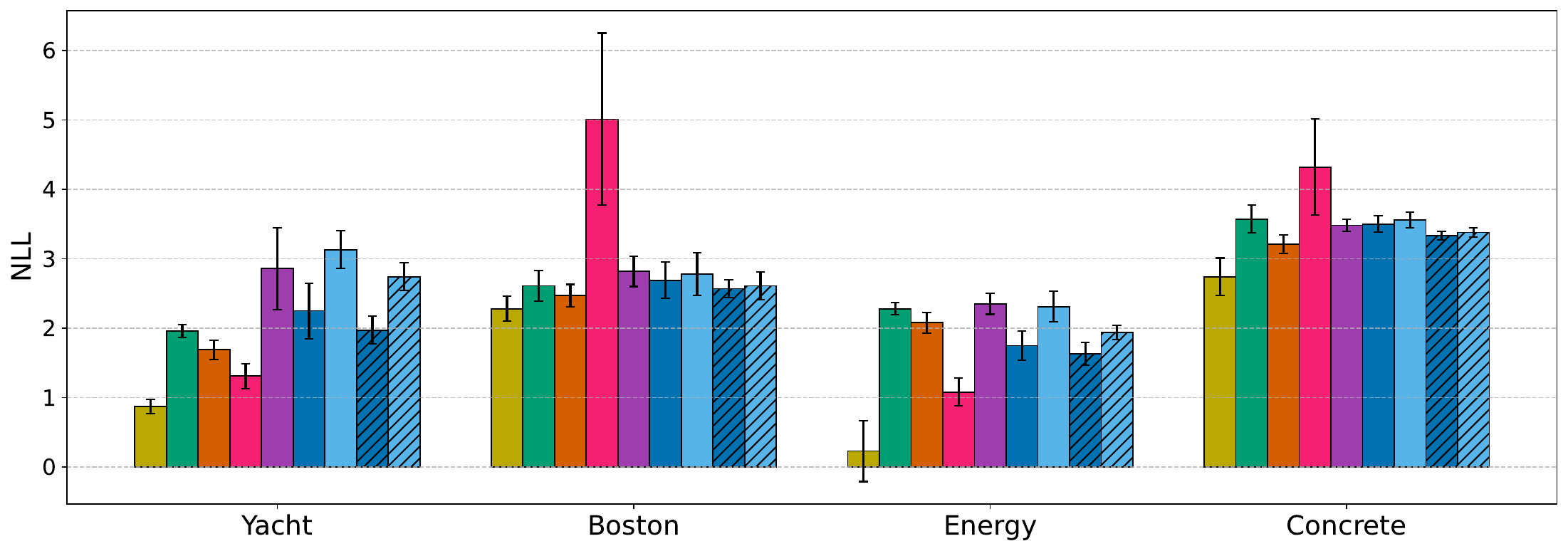}
\includegraphics[width=0.8\linewidth]{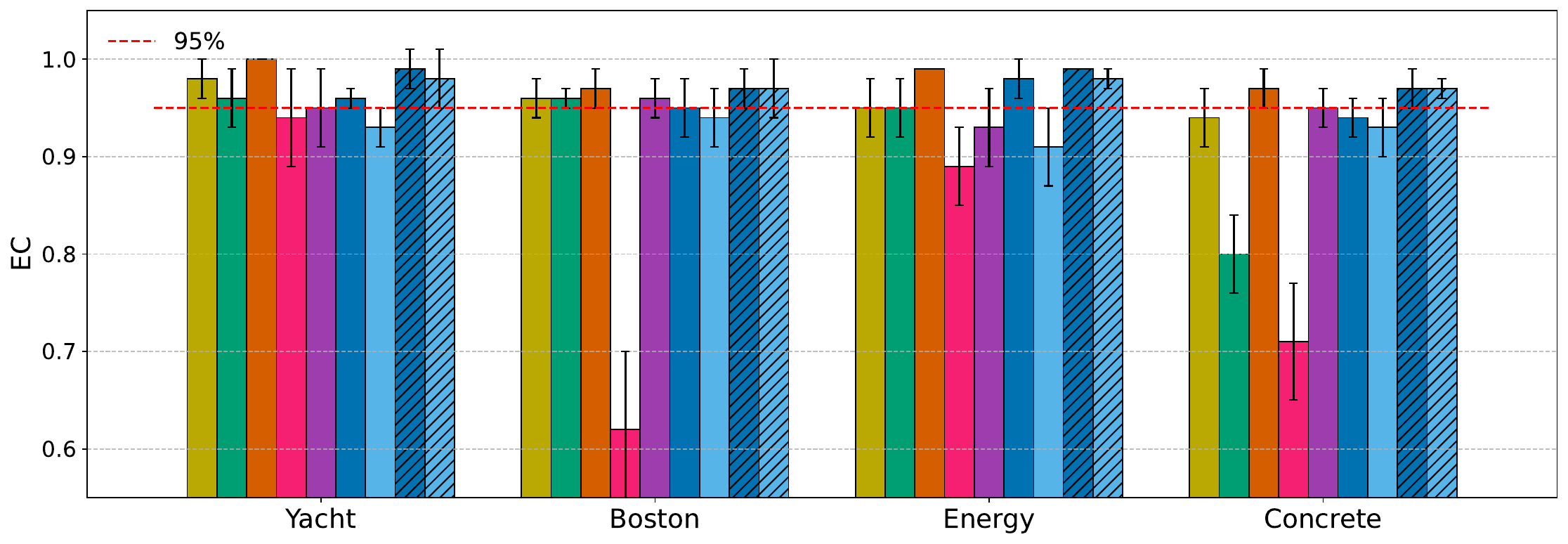}
    \caption{RMSE (normalized w.r.t. to the standard deviation of the target), NLL and empirical coverage for UCI datasets. When illustrating the coverage, the dashed red line depicts the ideal scenario with empirical coverage equal to $95\%$ CI level. }
    \label{fig:Monster}
\end{figure}
\label{sec:UCIdatasetsap}
Lastly, we consider publicly available datasets from the UCI Machine Learning Repository \citep{UCIMLrep}: \cmsspy{Boston housing} \citep{bostobhousing}, \cmsspy{Energy} \citep{misc_energy_efficiency_242}, \cmsspy{Yacht dynamics} \citep{misc_yacht_hydrodynamics_243}, \cmsspy{Concrete compressive strength} \citep{misc_concrete_compressive_strength_165} and \cmsspy{Concrete slump test} \citep{misc_concrete_slump_test_182} (see \cref{sec:experimentsdata} for the description of the datasets).
For all of the UCI regression tasks, we fit a neural network with one hidden layer and $D_H = 50$ hidden units. \cref{fig:Monster} compares RMSE, NLL and empirical coverage of the observations across the methods (see also \cref{tab:metricsUCI} in \cref{sec:monstertableforUCI}), %for \blu{mfVI, HMC, BBB,HSBNN, GVBNN, VBNN and, finally, SVBNN baselines for four of the datasets, 
\blu{where we additionally consider ensembles of four variational approximations with VBNN and SVBNN (denoted as 4VBNN and 4SVBNN, respectively).  %The corresponding to the experiments \cref{tab:metricsUCI} is provided in \cref{sec:monstertableforUCI}.
Overall, HMC outperforms all the considered methods, but at a much higher cost. VBNN provides an improvement compared to GVBNN, further motivating the choice of sparsity-inducing priors, and while SVBNN offers considerable computational savings, the quality of the approximation deteriorates compared to VBNN. 
Overall, VBNN has consistently well-calibrated uncertainty quantification and empirical coverage for the observations compared with other variational approaches, and ensembles of VBNNs are competitive with other approaches in terms of the RMSE and NLL. 
 \begin{figure}[ht!]
    \centering
\includegraphics[width=0.49\linewidth]{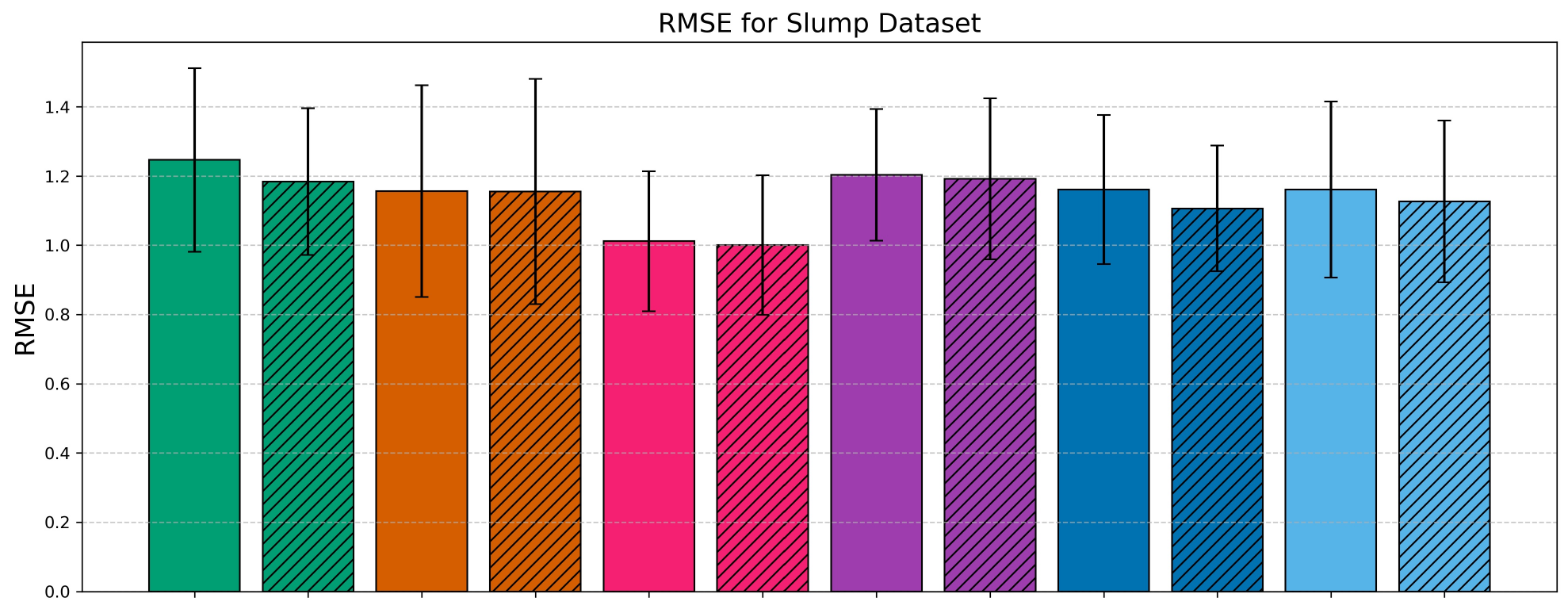}
\includegraphics[width=0.49\linewidth]{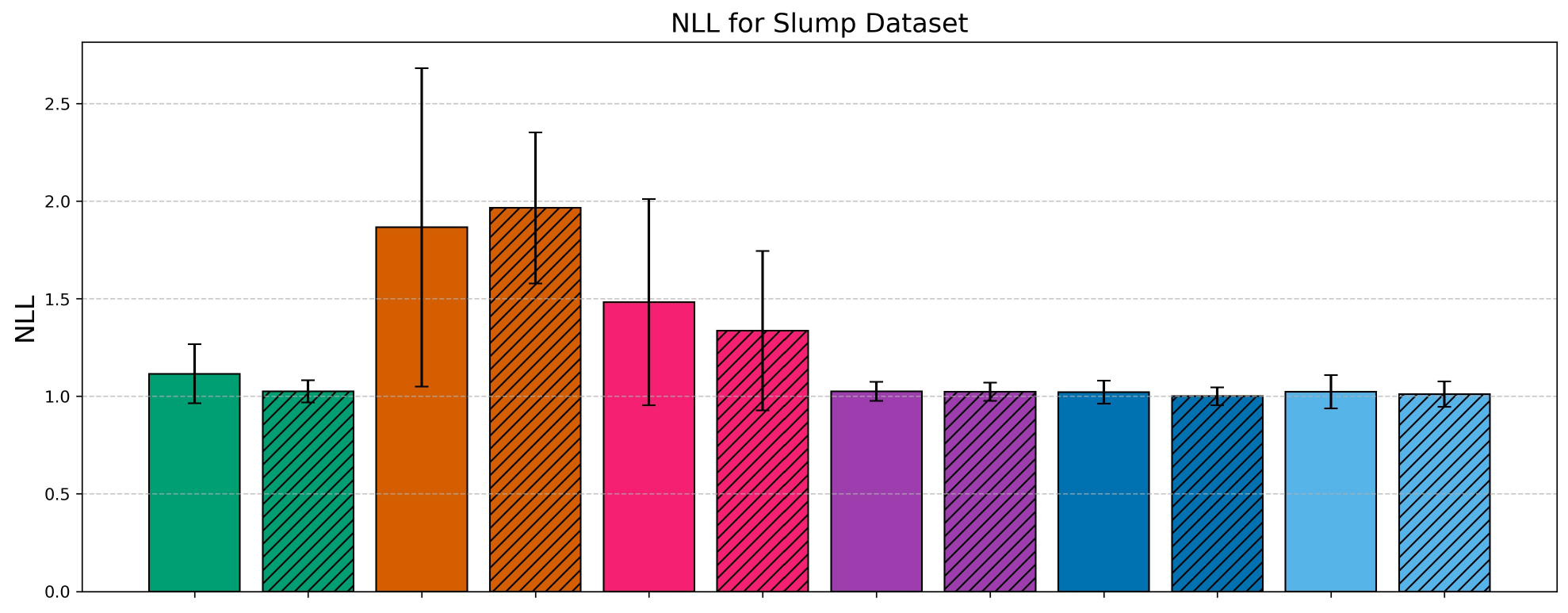}
\includegraphics[width=0.49\linewidth]{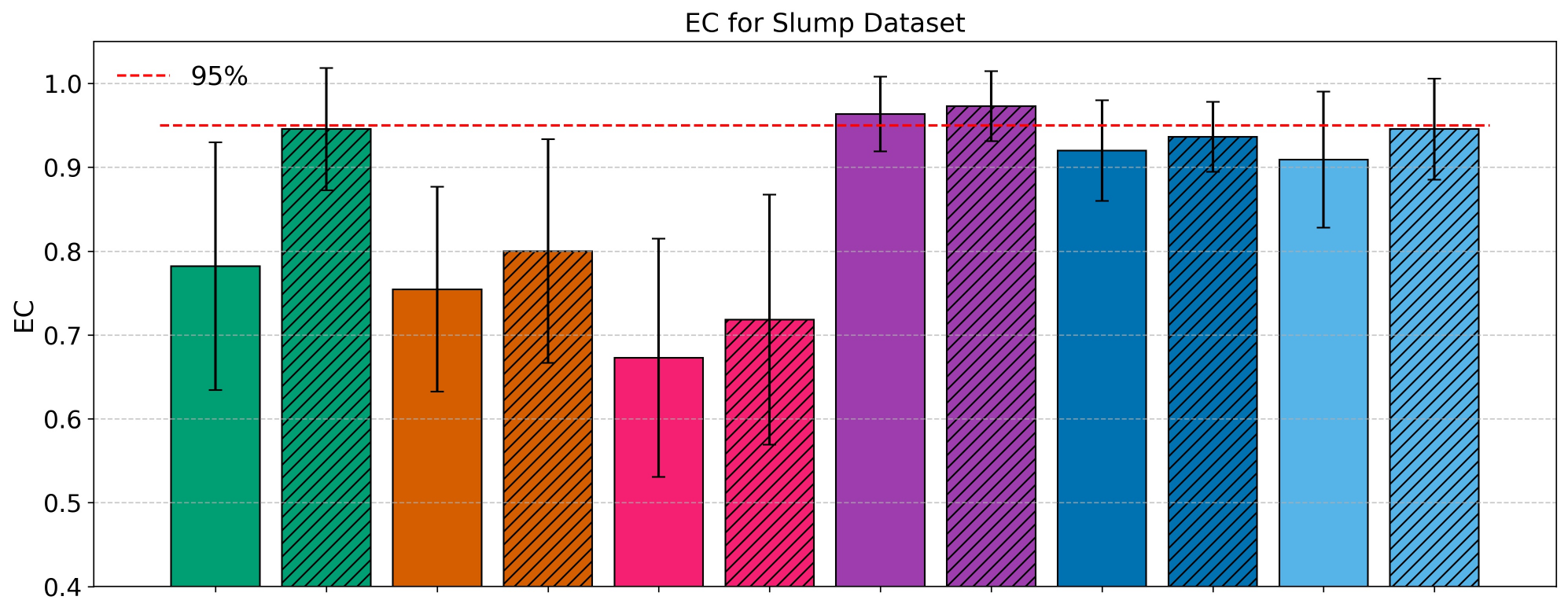}
\includegraphics[width=0.49\linewidth]{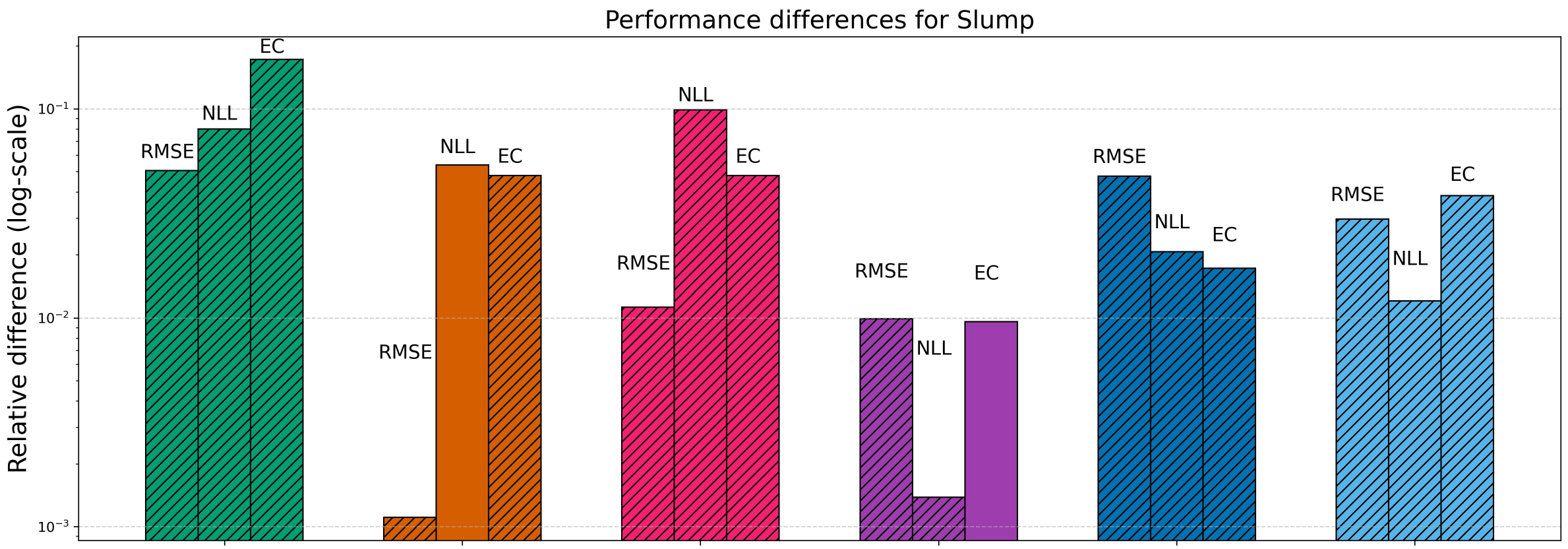}
\includegraphics[width=0.6\linewidth]{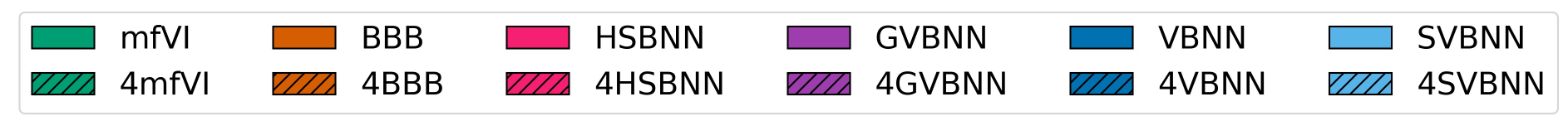}
    \caption{Slump dataset. Performance in terms of the RMSE, NLL and EC for single models (plain colored) and ensembles (color with hatches) obtained from four parallel runs. RMSE and NLL are scaled with respect to the best model (top row). The relative performance (bottom right) is illustrated on the log-scale, and color reflects if ensembles improved the metric (i.e. bar with hatches illustrates the scale of improvements obtained with ensembles, conversely, bar without the hatches illustrates the scale at which single run outperformed ensembles).}
    \label{fig:Slump}
\end{figure}
Further, we consider \cmsspy{slump} dataset and implement ensembles of 4 parallel runs of all of the considered methods (4mfVI, 4BBB, 4HSBNN, 4GVBNN, 4VBNN, 4SVBNN) and compare the results to approximations obtained in a single run (mfVI, BBB, HSBNN, GVBNN, VBNN, SVBNN). We do not consider HMC in this experiment due to its computational costs.
\cref{fig:Slump} compares RMSE, NLL, EC of the 12 methods and additionally illustrates relative differences among approaches, where for RMSE and NLL, we consider absolute relative differences between ensembles and single runs, and for empirical coverage of the observations, we illustrate the absolute deviation from the $95\%$ CI. 
 Overall, ensembles improve uncertainty quantification, and in most cases also RMSE and NLL (the only exception being NLL for BBB). While BBB and HSBNN have the lowest RMSE (although with high variability), the NLL and empirical coverage suggest overconfidence, even with ensembles. In contrast, VBNN has an improved balance between accuracy and uncertainty quantification, which is further enhaced by ensembles. % with an exception of GVBNN, which became overconfident. 
}

\section{Discussion}
In this paper, we presented a variational bow tie neural network (VBNN) that is amendable to Polya-gamma data augmentation so that the variational inference can be performed via the CAVI algorithm.  While the idea of the stochastic relaxation described in \cref{sec:bowtie} was introduced in \citep{smithbayesian}, the novelty of our model is in the employment of the variational inference techniques as well as sparsity-inducing priors. 
Namely, we implement continuous global-local shrinkage priors and propose a post-process technique for node selection.
Additionally, we consider an improvement of the classical CAVI algorithm by adding EM steps for critical hyperparameters. In this way, we enrich the class of models which are handled within the structured mean-field paradigm. We provide all the necessary computations, techniques, and illustrative experiments demonstrating the utility of the model. 
\blu{Addressing the scalability with respect to the number of data points, we extend the CAVI algorithm to SVI \citep{hoffman2013stochastic} which benefits from exploiting natural gradients and subsampling. In the future, we could improve the algorithm by employing an adaptive learning rate which is based on realisations of a noisy estimate of the natural gradient of ELBO with respect to global variational parameters and moving averages \citep{ranganath13adaptivelr, schaul2013no}. Alternatively, instead of changing the learning rate, one could adapt the mini-batch size based on the estimated gradient noise covariance and the magnitude of the gradient \citep{balles2017coupling}. To address scalability with respect to the network's width, future work will explore incorporating node selection within the CAVI algorithm when training. 

The variational bow tie neural network is also amenable to other prior choices. %has N-GIG priors on the weights and, alternatively, the model with 
For example, horseshoe priors %would be suitable for the coordinate ascent algorithm as long as one 
can be implemented through the introduction of auxiliary variables to replace each half-Cauchy random variable with the hierarchical formulation based on Inverse-Gamma variables \citep{wand2011mean,louizos2017, ghosh2018}.  Additionally,  the regularized version of horseshoe priors ("ponyshoe") could be considered, which is known to perform better than the classical horseshoe, especially when the larger coefficients are weakly identified by the data \citep{piironen2017sparsity, PiironenHorseshoe,ghosh2019}.}
Finally, an extension to other output types, such as classification tasks, can be developed through additional Polya-gamma augmentation techniques \citep{durante2019}.


\bibliographystyle{ACM-Reference-Format}
\bibliography{refs}


\appendix
\section{Derivations of the Variational Posterior}\label{appendix:vb}

%\SW{go through all these expressions and derivations, try to clean and see if some simple steps can be skipped/commented (I have gone through a bit for $\tau, \psi, \eta$).}
%\SW{remove repeated operation, e.g. $+$ at the end of  the equation that spans multiple lines. }
\paragraph{Global shrinkage parameters.}

 Using \cref{eqn::CAVIupdate}, the variational posterior for the global shrinkage parameters is: 
\begin{align*}   
q(\btau) &\propto  \exp \left (\E \left [  \log \prod_{l}^{L+1} \prod_{d}^{D_l} \prod_{d'}^{D_{l-1}} \Norm\left(W_{l,d,d'} |  0, \tau_l \psi_{l,d,d'} \right) \right ] + \log \prod_{l}^{L+1} \GIG\left( \tau_l \mid \nu_{\glob}, \delta_{\glob}, \lambda_{\glob} \right) \right )\\ 
 %&\propto  \prod_{l}^{L+1} \exp \left (\E \left [  \log \prod_{d}^{D_l} \prod_{d'}^{D_{l-1}} \Norm\left(W_{l,d,d'} |  0, \tau_l \psi_{l,d,d'} \right) \right ] + \log \GIG\left( \tau_l \mid \nu_{\glob}, \delta_{\glob}, \lambda_{\glob} \right) \right ) \\
 %&\propto  \prod_{l}^{L+1} \prod_{d}^{D_l} \prod_{d'}^{D_{l-1}} \exp \E \left [ \log \Norm\left(W_{l,d,d'} |  0, \tau_l \psi_{l,d,d'} \right) \right ]\GIG\left( \tau_l \mid \nu_{\glob}, \delta_{\glob}, \lambda_{\glob} \right)  \\
&\propto \prod_{l}^{L+1} \prod_{d}^{D_l} \prod_{d'}^{D_{l-1}} \exp \E \left [ \log \left(\frac{1}{\sqrt{\tau_l \psi_{l,d,d'}}} \exp \left(  -\frac{W_{l,d,d'}^2}{2\tau_l \psi_{l,d,d'}}\right )\right) \right ] \times \prod_{l}^{L+1} \tau_l^{\nu_{\glob}-1} \exp \left( -\frac{1}{2} \left(\frac{\delta_{\glob}^2}{\tau_l} + \lambda_{\glob}^2\tau_l \right) \right ) \\
  % & \propto \prod_{l}^{L+1} \prod_{d}^{D_l} \prod_{d'}^{D_{l-1}} \exp \left(-\frac{1}{2} \log \tau_l -\frac{1}{2\tau_l}\E \left [ \frac{1}{\psi_{l,d,d'}} \right] \E \left[ W_{l,d,d'}^2  \right] \right) \times  \prod_{l}^{L+1} \tau_l^{\nu_{\glob}-1} \exp \left( -\frac{\delta_{\glob}^2}{2\tau_l} -\frac{\lambda_{\glob}^2\tau_l}{2}   \right )  \\
     & \propto \prod_{l}^{L+1} \tau_l^{\nu_{\glob}-1} \exp \left( -\frac{\delta_{\glob}^2}{2\tau_l} -\frac{\lambda_{\glob}^2\tau_l}{2}   \right )  \prod_{d}^{D_l} \prod_{d'}^{D_{l-1}} \tau_l^{-\frac{1}{2}} \exp \left ( -\frac{\E \left [ \frac{1}{\psi_{l,d,d'}} \right] \E \left[ W_{l,d,d'}^2  \right]}{2\tau_l} \right) \\
     & \propto \prod_{l}^{L+1}\tau_l^{\nu_{\glob}-\frac{D_lD_{l-1}}{2} -1} \exp \left(-\frac{1}{2} \left (\frac{1}{\tau_l} \left ( \sum_{d}^{D_l} \sum_{d'}^{D_{l-1}} \E \left [ \frac{1}{\psi_{l,d,d'}} \right] \E \left[ W_{l,d,d'}^2  \right] + \delta_{\glob}^2 \right ) +\lambda_{\glob}^2\tau_l \right )  \right ) \\
      &\propto \prod_{l}^{L+1} \GIG \left (\tau_l \; \mid \; \hat{\nu}_{\glob,l}, \hat{\delta}_{\glob,l}, \lambda_{\glob} \right ),
\end{align*}
where for $l=1,\ldots,L+1$
\begin{align*}
 \hat{\nu}_{\glob,l} &= \nu_{\glob}-\frac{D_lD_{l-1}}{2}, \\
     \hat{\delta}_{\glob,l}  &= \sqrt{\delta_{\glob}^2 + \sum_{d}^{D_l} \sum_{d'}^{D_{l-1}} \E \left [ \frac{1}{\psi_{l,d,d'}} \right] \E \left[ w_{l,d,d'}^2  \right]}.
\end{align*}
\blu{Assuming hidden layers of the dimension $D$, the computational complexity of updating the global shrinkage variable is  $\mathcal{O}(LD\max(D_0,D, D_{L+1}))$}.
\paragraph{Local shrinkage parameters.}
Similarly, the variational posterior for the local shrinkage parameters is: 
\begin{align*}   
q(\bpsi) &\propto  \prod_l^{L+1} \exp \left ( \E \left [  \log \prod_{d}^{D_l} \prod_{d'}^{D_{l-1}} \Norm\left(W_{l,d,d'} |  0, \tau_l \psi_{l,d,d'} \right) \right ] +\log \prod_{d=1}^{D_l} \prod_{d'}^{D_{l-1}}\GIG \left( \psi_{l,d,d'} \mid \nu_{\loc,l}, \delta_{\loc,l}, \lambda_{\loc,l} \right) \right ) \\
%&\propto \prod_l^{L+1} \prod_{d}^{D_l} \prod_{d'}^{D_{l-1}}  \exp \left (\E \left [  \log  \Norm\left(W_{l,d,d'} |  0, \tau_l \psi_{l,d,d'} \right) \right ] \right )  \psi_{l,d,d'}^{\nu_{\loc, l}-1} \exp \left( -\frac{1}{2} \left(\frac{\delta_{\loc, l}^2}{\psi_{l,d,d'}} + \lambda_{\loc, l}^2\psi_{l,d,d'} \right) \right )   \\
&\propto \prod_l^{L+1} \prod_{d}^{D_l} \prod_{d'}^{D_{l-1}} \exp \left ( \frac{1}{2} \log \psi_{l,d,d'} -\frac{\E \left [ \frac{1}{\tau_l} \right ]\E \left [ \frac{1}{\psi_{l,d,d'}} \right] \E \left[ W_{l,d,d'}^2  \right]}{2} \right)  \psi_{l,d,d'}^{\nu_{\loc, l}-1} \exp \left( -\frac{1}{2} \left(\frac{\delta_{\loc, l}^2}{\psi_{l,d,d'}} + \lambda_{\loc, l}^2\psi_{l,d,d'} \right) \right ) \\
& \propto \prod_l^{L+1} \prod_{d}^{D_l} \prod_{d'}^{D_{l-1}}  \psi_{l,d,d'}^{\nu_{\loc, l}-\frac{1}{2}} \exp \left (-\frac{1}{2}  \left (\frac{1}{\psi_{l,d,d'}} \left ( \E \left [ \frac{1}{\tau_l} \right ]\E \left[ W_{l,d,d'}^2  \right]  + \delta_{\loc, l}^2 \right ) + \lambda_{\loc, l}^2\psi_{l,d,d'}  \right ) \right ) \\
&  \propto \prod_{l}^{L+1} \prod_{d}^{D_l} \prod_{d'}^{D_{l-1}} \GIG \left (\psi_{l,d,d'} \; \mid \; \hat{\nu}_{\loc,l,d,d'},\hat{\delta}_{\loc,l,d,d'}, \lambda_{\loc, l} \right ),
\end{align*}
where for $l=1, \dots, L+1, \; d=1,\ldots, D_l, D_{l-1}\; d'=1,\ldots, D_{l-1}$
\begin{align*}
    \hat{\nu}_{\loc,l,d,d'} &= \nu_{\loc, l}-\frac{1}{2}, \\
    \hat{\delta}_{\loc,l,d,d'} &=  \sqrt{ \E \left [ \frac{1}{\tau_l} \right ] \E \left[ W_{l,d,d'}^2  \right]  + \delta_{\loc, l}^2}.
\end{align*}
\blu{Similarly, given hidden layers of the dimension $D$, the computational complexity of updating the global shrinkage variable is  $\mathcal{O}(LD\max(D_0, D, D_{L+1}))$}.
\paragraph{Covariance matrix.} Under the assumption of a diagonal covariance matrix, with parameters $\bmeta_l=(\eta_{l,1}^2,\ldots  \eta_{l,D_l}^2)$, the variational posterior is: 
\begin{align*}
    q(\bmeta) & \propto \exp \left ( \E \left [ \log \prod_{n}^N \Norm\left(\by_n \mid \bz_{n,L+1}, \bSigma_{L+1}\right) +\log \prod_{n}^N \prod_{l}^L \Norm \left( \ba_{n,l} \mid \bgamma_{n,l} \odot \bz_{n,l} , \bSigma_{l} \right)  \right ) \right )  \\
    & \times \prod_{l}^{L} \prod_{d}^{D_l} \IG(\eta_{l,d}^2 \mid \alpha^h_0, \beta^h_0)  \prod_{d}^{D_{L+1}} \IG(\eta_{l,d}^2 \mid \alpha_0, \beta_0) \\
   % & \propto \exp \left ( \E \left [ \log \prod_{n}^N \frac{1}{\sqrt{ |\bSigma_{L+1}| }} \exp \left(-\frac {1}{2} \left ( \by_n - \bz_{n,L+1} \right )^{T} (\bSigma_{L+1})^{-1} \left ( \by_n - \bz_{n,L+1} \right )\right)  \right. \right. \\
   % & + \left. \left.\log \prod_{n}^N \prod_{l}^L 
   % \frac{1}{\sqrt{|\bSigma_{l}| }} \exp \left(-\frac {1}{2} \left ( \ba_{n,l} - \bgamma_{n,l} \odot \bz_{n,l}  \right )^{T} (\bSigma_{l})^{-1} \left ( \ba_{n,l} - \gamma_{n,l} \odot \bz_{n,l}  \right )\right) \right ] \right )\times \\
   % & \times \prod_{l}^{L} \prod_{d}^{D_l} \left (  (\eta_{l,d}^2)^{-\alpha^h_0 - 1}\exp \left ( -\frac{\beta^h_0 }{ \eta^2_{l,d}} \right )\right ) \times \prod_{d}^{D_{L+1}} \left (  (\eta_{L+1,d}^2)^{-\alpha_0 - 1}\exp \left ( -\frac{\beta_0 }{ \eta^2_{L+1,d}} \right )\right )  \\
    & \propto \exp \left ( -\frac {1}{2}\E \left [\sum_{n}^N \sum_{d}^{D_{L+1}}(\eta_{L+1, d})^{-2} \left(  y_{n,d} - z_{n,L+1, d} \right )^{2}   \right] \right)  \prod_{d}^{D_{L+1}} \left (  (\eta_{L+1,d}^2)^{-\alpha_0 - 1 -\frac{N}{2}}\exp \left ( -\frac{\beta_0 }{ \eta^2_{L+1,d}} \right )\right ) \\
    & \times \prod_{l}^{L} \exp \left ( -\frac {1}{2}\E \left [\sum_{n}^N \sum_{d}^{D_l}(\eta_{l, d})^{-2} \left(  \blu{\mathrm{a}_{n,l,d}} - \gamma_{n,l,d} \odot z_{n,l, d} \right )^{2}   \right] \right) \times \prod_{l}^{L}  \prod_{d}^{D_{l}} \left (  (\eta_{l,d}^2)^{-\alpha^h_0 - 1 -\frac{N}{2}}\exp \left ( -\frac{\beta^h_0 }{ \eta^2_{l,d}} \right )\right ) \\
    &  \propto \prod_{d}^{D_{L+1}} \left (  (\eta_{L+1,d}^2)^{-\alpha_0 - 1 -\frac{N}{2}} \right )  \exp \left ( - \frac{1}{\eta^2_{L+1,d}} \left ( \beta_0   + \frac {1}{2}\sum_{n}^N\E \left [ \left( y_{n,d} - z_{n,L+1, d} \right )^{2}   \right] \right ) \right)  \\
    & \times \prod_{l}^{L}\prod_{d}^{D_l} (\eta_{l,d}^2)^{-\alpha^h_0 - 1 -\frac{N}{2}}
    \exp \left ( - \frac{1}{\eta^{2}_{l,d}} \left (\beta^h_0 + \frac {1}{2}\sum_{n}^N \E \left [ \left(  \blu{\mathrm{a}_{n,l,d}} - \gamma_{n,l,d} \odot z_{n,l, d} \right )^{2} \right] \right) \right ).
\end{align*}

Thus, $ q(\bmeta) \propto \prod_l^{L+1} \prod_d^{D_l} \IG(\alpha_{l,d}, \beta_{l,d})$, where 

\begin{align*}
    \alpha_{l,d} &= \alpha^h_0+ \frac{N}{2}, \quad d = 1, \ldots, D_l, \quad l=1, \ldots, L,\\
   \alpha_{L+1,d} &= \alpha_0  + \frac{N}{2}, \quad d = 1, \ldots, D_{L+1},\\
   \beta_{l,d} &= \beta^h_0 + \frac {1}{2}\sum_{n}^N \E \left [ \left(  \blu{\mathrm{a}_{n,l,d}} - \gamma_{n,l,d} \odot z_{n,l, d} \right )^{2} \right], \quad d = 1, \ldots, D_l, \quad l=1, \ldots, L,\\
   \beta_{L+1 ,d} &= \beta_0   + \frac {1}{2}\sum_{n}^N\E \left [ \left( y_{n,d} - z_{n,L+1, d} \right )^{2}   \right], \quad d = 1, \ldots, D_{L+1}. 
\end{align*}

For the parameters $ \beta_{l,d}$, we must compute the sum of squares terms. For the last layer $l=L+1$, this term, for each data point $n$, is given by:
% \begin{align*}
%   \E \left [ \left(  y_{n,d} - z_{n,L+1, d} \right )^{2} \right ]  & =  y^2_{n,d} - 2 y_{n,d} \E \left [\bW_{L+1, d} \ba_{n,L}\right ] -2 y_{n,d} \E \left [b_{L+1, d} \right ]  + \E \left [ (\bW_{L+1,d} \ba_{n,L})^2 \right] + \E \left [ b_{L+1,d}^2 \right] + 2 \E \left [ \bW_{L+1,d} \ba_{n,L} b_{L+1, d}\right] \\
%     & =  y^2_{n,d} - 2  y_{n,d} \E \left [\bW_{L+1, d}\right ] \E \left [  \ba_{n,L} \right ] -2  y_{n,d} \E \left [b_{L+1, d} \right ] \\
%     & + \Tr \left ( \E \left [  \bW_{L+1,d}^T \bW_{L+1,d} \right ] \E \left [ \ba_{n,L} \ba_{n,L}^T\right] \right ) + \E \left [ b_{L+1,d}^2 \right] + 2 \E \left [  b_{L+1, d}  \bW_{L+1,d} \right ] \E \left [ \ba_{n,L}\right].
% \end{align*}
\begin{align*}
   \E \left [ \left(  y_{n,d} - z_{n,L+1, d} \right )^{2} \right ] & =\sum_{n}^N  \left(  y_{n,d} -    \E \left [\bW_{L+1, d}\right ] \E \left [  \ba_{n,L} \right ] - \E \left [b_{L+1, d} \right ] \right )^2   \\
    & + \sum_{n}^N  \E \left [b_{L+1, d}^2 \right ] - \E \left [b_{L+1, d} \right ]^2 + 2 \E \left [  b_{L+1, d}  \bW_{L+1,d} \right ] \E \left [ \ba_{n,L}\right] - 2 \E \left [  b_{L+1, d} \right] \E \left [\bW_{L+1,d} \right ] \E \left [ \ba_{n,L}\right]  \\
    & + 2\sum_{n}^N  \Tr \left ( \E \left [  \bW_{L+1,d}^T \bW_{L+1,d} \right ] \E \left [ \ba_{n,L} \ba_{n,L}^T\right] \right ) - \Tr \left ( \E \left [  \bW_{L+1,d}^T \right ] \E \left [ \bW_{L+1,d} \right ] \E \left [ \ba_{n,L} \right ] \E \left [  \ba_{n,L}^T\right] \right ).
\end{align*}

% Alternatively, it can be computationally more efficient and accurate to use \SW{is this what we use? Just keep the expression for the one that we use...}

% \begin{align*}
%  \E \left [ \left(  y_{n,d} - z_{n,L+1, d} \right )^{2} \right ] &=
%    \E \left [ \left(  y_{n,d} - \E \left [z_{n,L+1, d}\right ] + \E \left [z_{n,L+1, d}\right ]- z_{n,L+1, d} \right )^{2} \right ]\\
%    &=   \left(  y_{n,d} - \E \left [z_{n,L+1, d}\right ] \right )^{2} + \Var(z_{n,L+1, d}),
% \end{align*}
% where
% \begin{align*}
%     \E \left [z_{n,L+1, d}\right ] &= \E \left [b_{L+1, d} \right ]+\E \left [\bW_{L+1, d}\right ] \E \left [  \ba_{n,L} \right ], \\
%     \Var \left (z_{n,L+1, d}\right ) &= \Var(\widetilde{\bW}_{d} \widetilde{\ba}_{n}) = \Tr \left ( \E \left [  \widetilde{\bW}_{d}^T \widetilde{\bW}_{d} \right ] \E \left [ \widetilde{\ba}_{n} \widetilde{\ba}_{n}^T\right] - \E [  \widetilde{\bW}_{d}]^T \E[\widetilde{\bW}_{d} ] \E [ \widetilde{\ba}_{n}] \E[\widetilde{\ba}_{n}^T]\right ).
% \end{align*}

% \begin{align*}
%     \beta_{L+1,d}&  = \beta_0 +  \frac{1}{2}\sum_{n}^N  \left( y_{n,d} - \E \left [  \widetilde{\bW}_{L+1,d} \right ]  \E [ \widetilde{\ba}_{n,L}] \right)^2+ \\
%     & +\frac{1}{2} \sum_n^N  \Tr \left ( \E \left [  \widetilde{\bW}_{L+1,d}^T \widetilde{\bW}_{L+1,d} \right ] \E \left [ \widetilde{\ba}_{n,L} \widetilde{\ba}_{n,L}^T\right] \right) - \Tr \left (\E [  \widetilde{\bW}_{L+1,d}]^T \E[\widetilde{\bW}_{L+1,d} ] \E [ \widetilde{\ba}_{n,L}] \E[\widetilde{\ba}_{n,L}^T]\right ).
% \end{align*}

Instead, for an intermediate layer $l=1,\ldots, L$, the sum of squares term, for each data point $n$, is given by:
% \begin{align*}
%     \E \left [ \left(  \blu{\mathrm{a}_{n,l,d}} - \gamma_{n,l,d}\cdot z_{n,l, d} \right )^{2} \right ] &= \E \left [  \blu{\mathrm{a}_{n,l,d}}^2 + \left( \gamma_{n,l,d}\bW_{l,d} \ba_{n,l-1} \right )^{2} + \left(\gamma_{n,l,d} b_{l, d} \right ) ^{2}  \right ]  \\
%    & + \E \left [ 2 \gamma_{n,l,d}^2 \bW_{l,d} \ba_{n,l-1}b_{l, d} - 2a_{n,l,d}\left(\gamma_{n,l,d}\bW_{l,d} \ba_{n,l-1} + \gamma_{n,l,d} b_{l, d} \right )  \right ] \\
%     & =  \E \left [a^2_{n,l,d} \right ] + \E \left [\gamma_{n,l,d}^2 \right ] \Tr \left ( \E \left [ \bW_{l,d}^T \bW_{l,d} \right ] \E \left [ \ba_{n,l-1} \ba_{n,l-1}^T\right ] \right ) + \E \left [\gamma_{n,l,d} \right ] \E \left [ b_{l,d}^2 \right ]  \\
%    & - 2  \E \left [\gamma_{n,l,d} \right ] \E \left [ \bW_{l, d} \right ] \E \left [ \ba_{n,l-1} \blu{\mathrm{a}_{n,l,d}}\right ] - 2  \E \left [\gamma_{n,l,d} \right ]\E \left [\blu{\mathrm{a}_{n,l,d}} \right ] \E \left [ b_{l, d} \right ]  + 2 \E \left [\gamma_{n,l,d} \right ]\E \left [   b_{l,d} \bW_{l,d} \right ] \E \left [ \ba_{n,l-1}\right].
% \end{align*}
% Then 
\begin{align*}
 \E \left [ \left(  \blu{\mathrm{a}_{n,l,d}} - \gamma_{n,l,d}\cdot z_{n,l, d} \right )^{2} \right ] &
%  = \sum_{n}^N \left ( \E \left [a^2_{n,l,d} \right ] + \E \left [\gamma_{n,l,d} \right ] \Tr \left ( \E \left [ \bW_{l,d}^T \bW_{l,d} \right ] \E \left [ \ba_{n,l-1} \ba_{n,l-1}^T\right ] \right ) +  \E \left [\gamma_{n,l,d} \right ] \E \left [ b_{l,d}^2 \right ]\right )  \\
% & + 2 \sum_{n}^N \left ( \E \left [\gamma_{n,l,d} \right ]\E \left [   b_{l,d} \bW_{l,d} \right ] \E \left [ \ba_{n,l-1}\right]  - \E \left [\gamma_{n,l,d} \right ] \left ( \E \left [ \bW_{l, d} \right ] \E \left [ \ba_{n,l-1} \blu{\mathrm{a}_{n,l,d}}\right ] + \E \left [\blu{\mathrm{a}_{n,l,d}} \right ] \E \left [ b_{l, d} \right ]  
%      \right ) \right ) \\
%  & 
 = \sum_{n}^N   \left(  \E \left [\blu{\mathrm{a}_{n,l,d}} \right ] -  \E \left [\gamma_{n,l,d} \right ] \E \left [ b_{l,d} \right ] - \E \left [\gamma_{n,l,d} \right ]   \E \left [\bW_{l,d} \right ]   \E \left [\ba_{n,l-1} \right ]\right )^2  \\
 & +\sum_{n}^N\E \left [a^2_{n,l,d} \right ] -  \E \left [\blu{\mathrm{a}_{n,l,d}} \right ]^2 + \E \left [\gamma_{n,l,d} \right ] \E \left [ b_{l,d}^2 \right ] - \E \left [\gamma_{n,l,d} \right ]^2 \E \left [ b_{l,d} \right ]^2  \\
 & + \sum_{n}^N\E \left [\gamma_{n,l,d} \right ] \Tr \left ( \E \left [ \bW_{l,d}^T \bW_{l,d} \right ]  \E \left [\ba_{n,l-1} \ba_{n,l-1}^T \right ] \right ) -    \E \left [\gamma_{n,l,d} \right ]^2  \Tr \left ( \E \left [ \bW_{l,d}^T \right ] \E \left [\bW_{l,d} \right ] \left [\ba_{n,l-1} \ba_{n,l-1}^T \right ]  \right ) \\
& + 2 \sum_{n}^N  \E \left [\gamma_{n,l,d} \right ]\E \left [   b_{l,d} \bW_{l,d} \right ] \left [\ba_{n,l-1}  \right ]   - 
   \E \left [\gamma_{n,l,d} \right ]^2\E \left [   b_{l,d} \right ] \E \left [ \bW_{l,d} \right ] \left [\ba_{n,l-1} \right ].
\end{align*}

 \blu{The complexity of obtaining variational update for $\bmeta$ is then $\mathcal{O}(N L\max(D, D_0)^2\max(D_{L+1},D))$. } 

\paragraph{Weights and biases.} The variational posterior for the weights and biases is: 
\begin{align*}
&q(\bb, \bW)\propto  \exp \left( \E \left [ \log  \prod_{n}^N \Norm \left ( y_n \mid \bW_{L+1} \ba_{n, L} + \bb_{L+1}, \bSigma_{L+1} \right)    \prod_{n}^{N} \prod_{l}^{L} \Norm \left ( \ba_{n, l} \mid \bgamma_{n,l} \odot (\bW_{l} \ba_{n,l-1} + \bb_{l}), \bSigma_l\right )  \right] \right) \\
& \times \exp \left( \E \left [ \log \prod_{n}^N \prod_{l}^L \prod_{d}^{D_l}  \exp\left(\frac{(\gamma_{n,l,d} - \frac{1}{2}) z_{n,l,d}}{T }\right) \exp\left(-\frac{\omega_{n,l,d}z_{n,l,d}^2}{2T^2}\right)    \right ]\right )\\
&\times 
\exp \left( \E \left [   \log \prod_{l}^{L+1} \prod_{d}^{D_l} \prod_{d'}^{D_{l-1}}\Norm\left(W_{l,d,d'} |  0, \tau_l \psi_{l,d,d'} \right) \right ]\right ) \prod_{l}^{L+1} \prod_{d}^{D_l}\Norm (b_{l,d} \mid 0, s^2_0) \\
&\propto \prod_{n}^N \exp \left ( \E \left [ \log  \frac{1}{\sqrt{ |\bSigma_{L+1}| }} \right] \right)  \exp \left( \E \left [ -\frac {1}{2} \left ( \by_n - \bW_{L+1} \ba_{n, L} - \bb_{L+1} \right )^{T} (\bSigma_{L+1})^{-1} \left ( \by_n - \bW_{L+1} \ba_{n, L} - \bb_{L+1} \right )  \right] \right ) \\
&\times \prod_{l}^L\prod_{n}^N \exp \left ( \E  \left [\log 
\frac{1}{\sqrt{|\bSigma_{l}| }} \right ] \right )   \exp \left ( \E  \left [\left(-\frac {1}{2} \left ( \ba_{n,l} - \bgamma_{n,l}\bW_{l} \ba_{n,l-1} - \bgamma_{n,l} \bb_{l}   \right )^{T} (\bSigma_{l})^{-1} \left ( \ba_{n,l} - \bgamma_{n,l}\bW_{l} \ba_{n,l-1} - \bgamma_{n,l} \bb_{l}   \right )\right) \right ] \right )  \\
& \times \prod_{l}^L\prod_{n}^N \prod_{d}^{D_l} \exp \left (\E \left [ \frac{(\gamma_{n,l,d} - \frac{1}{2}) \left (\bW_{l,d}\ba_{n,l-1} + b_{l,d}\right )}{T } \right ] \right )  \exp \left (\E \left [-\frac{\omega_{n,l,d}\left ( \bW_{l,d}\ba_{n,l-1} + b_{l,d}\right )^2}{2T^2}\right ] \right )\\
& \times   \prod_{l}^{L+1} \prod_{d}^{D_l} \prod_{d'}^{D_{l-1}}   \exp \left(  -\frac{W_{l,d,d'}^2}{2} \E \left [\frac{1}{\tau_l} \right ] \E \left [ \frac{1}{\psi_{l,d,d'}} \right ] \right )   \prod_{l}^{L+1} \prod_{d}^{D_l}  \exp \left(  -\frac{b_{l,d}^2}{2s_0^2}\right ). 
\end{align*}
Therefore, using also the fact that $\bSigma_l$ is diagonal,  we have that the variational posterior factorizes as $q(\bb, \bW)= \prod_{l}^{L+1} \prod_{d=1}^{D_l} q(b_{l,d}, \bW_{l,d})$. We consider the terms  $q(b_{l,d}, \bW_{l,d})$ for the intermediate layers $l=1,\dots, L$ and $q(b_{L+1,d}, \bW_{L+1,d})$ for the last layer separately.

% \begin{align*}
% &q(\bb_{L+1}, \bW_{L+1}) \propto 
%  \prod_{d}^{D_{L+1}}\frac{1}{s_0} \exp \left(  -\sum_{d'}^{D_{L}} \frac{W_{L+1,d,d'}^2}{2} \E \left [\frac{1}{\tau_L+1} \right ] \E \left [ \frac{1}{\psi_{L+1,d,d'}} \right ] -\frac{b_{L+1,d}^2}{2s_0^2}\right ) \times \\
% &\times \prod_{d}^{D_{L+1}}\exp \left ( \E \left [ \log  \left ( (\eta_{L+1,d})^{-N} \right )\exp \left(-\frac {1}{2} \sum_{n}^N (\eta_{L+1,d})^{-2}\left ( y_{n,d} - \bW_{L+1,d} \ba_{n, L} - b_{L+1,d} \right )^{2}\right)  \right] \right ) \\
% & \propto  \prod_{d}^{D_{L+1}} \exp \left(  -\sum_{d'}^{D_{L}} \frac{W_{L+1,d,d'}^2}{2}\E \left [\frac{1}{\tau_L+1} \right ] \E \left [ \frac{1}{\psi_{L+1,d,d'}} \right ] -\frac{b_{L+1,d}^2}{2s_0^2} \right ) \times \\
% &\times \prod_{d}^{D_{L+1}}\exp \left ( -\frac {1}{2} \sum_{n}^N \E \left [(\eta_{L+1,d})^{-2}\left ( y_{n,d} - \bW_{L+1,d} \ba_{n, L} - b_{L+1,d} \right )^{2} \right] \right ).
% \end{align*}

% Let $\widetilde{\bW}_{L+1, d} = (b_{L+1,d}, \bW_{L+1,d}) $,  $\widetilde{\ba}_{n, L+1} = (1, \ba_{n, L}^T)^T$ and
Starting with the last layer $L+1$, we 
first introduce the matrix $$\bD_{L+1,d}^{-1}= \diag \left( s_0^{-2}, \E \left [\tau_{L+1}^{-1}\right ] \E \left [ \psi_{L+1,d,1}^{-1} \right ], \ldots, \E \left [\tau_{L+1}^{-1} \right ] \E \left [\psi_{L+1,d,D_{L}}^{-1} \right ]\right).$$   
Then, for the variational posterior of the weights and biases for the $d$th dimension of the final layer, we only need to consider the relevant terms:  
\begin{align*}
& q(b_{L+1,d}, \bW_{L+1,d}) \propto   \exp \left(  - \frac{1}{2} \widetilde{\bW}_{L+1,d} \bD^{-1}_{L+1,d} \widetilde{\bW}_{L+1,d}^T -\frac {1}{2}\E \left [(\eta_{L+1,d})^{-2} \right ] \sum_{n}^N \E \left [\left ( y_{n,d} - \widetilde{\bW}_{L+1, d} \widetilde{\ba}_{n, L } \right )^{2} \right] \right ) \\
& \propto   \exp \left(  - \frac{1}{2} \widetilde{\bW}_{L+1, d} \bD^{-1}_{L+1,d} \widetilde{\bW}_{L+1, d}^T -\frac {1}{2}\E \left [(\eta_{L+1,d})^{-2} \right ] \left ( \widetilde{\bW}_{L+1, d} \left ( \sum_{n}^N \E \left [\widetilde{\ba}_{n, L} \widetilde{\ba}_{n, L}^T \right]  \right )  \widetilde{\bW}_{L+1, d}^T - 2 \widetilde{\bW}_{L+1, d} \left ( \sum_{n}^N y_n \E \left [\widetilde{\ba}_{n, L} \right]  \right ) \right )\right ) \\
& \propto   \exp \left(  - \frac{1}{2}  \left (\widetilde{\bW}_{L+1, d} \left (\bD^{-1}_{L+1,d} +\E \left [(\eta_{L+1,d})^{-2} \right ] \sum_{n}^N \E \left [\widetilde{\ba}_{n, L} \widetilde{\ba}_{n, L}^T \right]  \right ) \widetilde{\bW}_{L+1, d}^T - 2 \E \left [(\eta_{L+1,d})^{-2} \right ] \widetilde{\bW}_{L+1, d} \left ( \sum_{n}^N y_n \E \left [\widetilde{\ba}_{n, L} \right]  \right ) \right )\right )\\
& \propto  \exp \left ( -\frac{1}{2} \left ( \widetilde{\bW}_{L+1, d} \mathbf{B}^{-1}_{L+1, d} \widetilde{\bW}_{L+1, d}^T - 2 \widetilde{\bW}_{L+1, d} \mathbf{B}^{-1}_{L+1, d} \mathbf{m}^T_{L+1, d} \right )\right ),
\end{align*}
where
\begin{align*}   \mathbf{B}^{-1}_{L+1,d}=&\bD^{-1}_{L+1,d} +\E \left [(\eta_{L+1,d})^{-2} \right ] \sum_{n}^N \E \left [\widetilde{\ba}_{n, L} \widetilde{\ba}_{n, L}^T \right], \\
    \mathbf{m}^T_{L+1,d} =& \mathbf{B}_{L+1,d}\E \left [(\eta_{L+1,d})^{-2} \right ] \left ( \sum_{n}^N y_n \E \left [\widetilde{\ba}_{n, L} \right]  \right ).
\end{align*}
% Note that $\mathbf{B}_{L+1,d} \in \R^{D_{L}+1 \times D_{L}+1}$ and $\mathbf{m}_{L+1,d} \in \R^{D_L+1}$. 
Thus, completing the square, we have that 
\begin{align*}
  q ( b_{L+1,d}, \bW_{L+1,d}) %\propto \prod_{d}^{D_{L+1}}  q(b_{L+1,d}, \bW_{L+1,d}) \\
%& \propto  \prod_{d}^{D_{L+1}} \exp \left ( \frac{1}{2} \left ( \widetilde{\bW}_{L+1, d} \mathbf{B}^{-1}_{L+1, d} \widetilde{\bW}_{L+1, d}^T - 2 \widetilde{\bW}_{L+1, d} \mathbf{B}^{-1}_{L+1, d} \mathbf{m}^T_{L+1, d} \right )\right ) \\
% &\propto \prod_{d}^{D_{L+1}}  \exp \left ( \frac{1}{2} \left ( \widetilde{\bW}_{L+1, d} - \mathbf{m}_{L+1, d} \right ) \mathbf{B}^{-1}_{L+1, d}\left ( \widetilde{\bW}_{L+1, d} - \mathbf{m}_{L+1, d} \right )^T  \right ) \\
  &=  \Norm \left  (\widetilde{\bW}_{L+1, d} \mid \mathbf{m}_{L+1, d}, \mathbf{B}_{L+1, d} \right ). 
\end{align*}

Next, for the intermediate layers $l=1,\dots, L,$ we can similarly obtain the variational posterior of the weights and biases $q( b_{l,d}, \bW_{l,d})$ for dimensions $\; d = 1, \ldots, D_{l}$. We introduce the matrices $$\bD_{l,d}^{-1}= \diag \left( s_0^{-2}, \E \left [\tau_l^{-1}  \right ] \E \left [\psi_{l,d,1}^{-1}\right ], \ldots, \E \left [\tau_l^{-1}  \right ] \E \left [ \psi_{l,d,D_{l-1}^{-1}} \right ]\right),$$ and consider the terms relevant to derive each  $q( b_{l,d}, \bW_{l,d})$ separately:

\begin{align*}
q( b_{l,d}, \bW_{l,d})&\propto \exp \left(  - \frac{1}{2} \widetilde{\bW}_{l,d} \bD^{-1}_{l,d} \widetilde{\bW}_{l,d}^T 
- \frac{1}{2T^2} \widetilde{\bW}_{l,d} \left (  \sum_{n}^N \E \left [\omega_{n,l,d}\right ] \E \left [\widetilde{\ba}_{n, l-1}\widetilde{\ba}_{n, l-1}^T \right ] \right ) \widetilde{\bW}_{l,d}^T \right . \\
& \left. -\frac {1}{2} \E \left [   (\eta_{l,d})^{-2} \right ]\widetilde{\bW}_{l,d} \left (  \sum_{n}^N \E \left [\gamma_{n,l,d}^2\right ] \E \left [\widetilde{\ba}_{n, l-1}\widetilde{\ba}_{n, l-1}^T \right ] \right ) \widetilde{\bW}_{l,d}^T  +\E \left [  (\eta_{l,d})^{-2} \right ]\widetilde{\bW}_{l,d} \left (\sum_{n}^N \E \left [   \gamma_{n,l,d}\right ] \E \left [ \blu{\mathrm{a}_{n,l,d}} \ba_{n,l-1}\right ] \right ) \right .   \\
& + \left. \frac{1}{T} \widetilde{\bW}_{l,d}  \left (\sum_{n}^N \E \left [\gamma_{n,l,d} \right ] \E \left [\widetilde{\ba}_{n, l-1} \right ] \right )- \frac{ 1}{2T} \widetilde{\bW}_{l,d} \left (\sum_{n}^N \E \left [ \widetilde{\ba}_{n, l-1}\right ]\right )  \right ) \\
% &  \propto \exp \left(  - \frac{1}{2} \left (\widetilde{\bW}_{l,d} \left (\bD^{-1}_{l,d}+\sum_{n}^N   \left (\left ( \frac{1}{T^2} \E \left [\omega_{n,l,d}\right ] + \E \left [   (\eta_{l,d})^{-2}\right ]\E \left [\gamma_{n,l,d}^2\right ]  \right) \E \left [\widetilde{\ba}_{n, l-1}\widetilde{\ba}_{n, l-1}^T \right ] \right )\right ) \widetilde{\bW}_{l,d}^T \right. \right. \\
% & \left. \left. -2\widetilde{\bW}_{l,d} \left (\sum_{n}^N \left (\E \left [  (\eta_{l,d})^{-2} \right ] \E \left [   \gamma_{n,l,d}\right ] \E \left [ \blu{\mathrm{a}_{n,l,d}} \widetilde{\ba}_{n,l-1}\right ] + \frac{1}{T} \E \left [\widetilde{\ba}_{n, l-1} \right ] \left ( \E \left [\gamma_{n,l,d} \right ]  - \frac{ 1}{2}\right) \right ) 
% \right ) \right )\right) \\
% &\propto \textbf{}\exp \left(  - \frac{1}{2}\left (\widetilde{\bW}_{l,d} \mathbf{B}_{l,d}^{-1}  \widetilde{\bW}_{l,d}^T -2  \widetilde{\bW}_{l,d}\mathbf{B}^{-1}_{l,d}\mathbf{m}^{T}_{l,d}  \right )\right ) \\
 &\propto \exp \left(  - \frac{1}{2}\left (\widetilde{\bW}_{l,d} \mathbf{B}_{l,d}^{-1}  \widetilde{\bW}_{l,d}^T -2  \widetilde{\bW}_{l,d}\mathbf{B}^{-1}_{l,d}\mathbf{m}^{T}_{l,d}  \right )\right ), \\
%&\propto \exp \left(  - \frac{1}{2}\left (\widetilde{\bW}_{l,d} - \mathbf{m}_{l,d} \right)  \mathbf{B}_{l,d}^{-1} \left (\widetilde{\bW}_{l,d} - \mathbf{m}_{l,d} \right)^{T} \right )  \\
%   & \propto \Norm \left ((b_{l,d}, \bW_{l,d}) \mid \mathbf{m}_{l,d}, \mathbf{B}_{l,d}\right ),
\end{align*}
where 
\begin{align*}
\mathbf{B}_{l,d}^{-1}&= \bD^{-1}_{l,d}+\sum_{n}^N   \left (\left ( \frac{1}{T^2} \E \left [\omega_{n,l,d}\right ] + \E \left [   (\eta_{l,d})^{-2}\right ]\E \left [\gamma_{n,l,d}\right ]  \right) \E \left [\widetilde{\ba}_{n, l-1}\widetilde{\ba}_{n, l-1}^T \right ] \right ), \\
    \mathbf{m}_{l,d}^T &= \mathbf{B}_{l,d}\left (\sum_{n}^N \left (\E \left [  (\eta_{l,d})^{-2} \right ] \E \left [   \gamma_{n,l,d}\right ] \E \left [ \blu{\mathrm{a}_{n,l,d}} \widetilde{\ba}_{n,l-1}\right ] + \frac{1}{T} \E \left [\widetilde{\ba}_{n, l-1} \right ] \left ( \E \left [\gamma_{n,l,d} \right ]  - \frac{ 1}{2}\right) \right ) \right). 
\end{align*}
Again, completing the square, we obtain the Gaussian variational posterior
% Then 
 \begin{align*}
     q( b_{l,d}, \bW_{l,d})  %&\propto \textbf{}\exp \left(  - \frac{1}{2}\left (\widetilde{\bW}_{l,d} \mathbf{B}_{l,d}^{-1}  \widetilde{\bW}_{l,d}^T -2  \widetilde{\bW}_{l,d}\mathbf{B}^{-1}_{l,d}\mathbf{m}^{T}_{l,d}  \right )\right ) \\
%     &\propto \textbf{}\exp \left(  - \frac{1}{2}\left (\widetilde{\bW}_{l,d} - \mathbf{m}_{l,d} \right)  \mathbf{B}_{l,d}^{-1} \left (\widetilde{\bW}_{l,d} - \mathbf{m}_{l,d} \right)^{T} \right )  \\
%    
&= \Norm \left ((b_{l,d}, \bW_{l,d}) \mid \mathbf{m}_{l,d}, \mathbf{B}_{l,d}\right ).
 \end{align*}

% Finally, we combine the factors of the variation posterior of the weights and biases we have obtained:
% \begin{equation*}
%      q(\bb, \bW) \propto \prod_{l}^{L+1} \prod_d^{D_l} \Norm \left ((b_{l,d}, \bW_{l,d}) \mid \mathbf{m}_{l,d}, \mathbf{B}_{l,d}\right ). 
% \end{equation*}

 \blu{The complexity of obtaining variational update for $\bW, \bb$ is then $\mathcal{O}((L\max(D_{L+1},D) (N\max(D, D_0)^2 + \max(D, D_0)^3))$. Assuming $\max(D, D_0)<N$, one gets   the same complexity as when updating $\bmeta$, i.e. $\mathcal{O}(LN\max(D, D_0)^2\max(D_{L+1},D) $.}

\paragraph{Augmented variables.} The variational posterior of the augmented variables is 
\begin{align*}
q(\bomega)&\propto  \exp \left ( \E \left [ \log \prod_{n}^N \prod_{l}^L \prod_{d}^{D_l} \exp \left (-\frac{\omega_{n,l,d}z_{n,l,d}^2}{2T^2}\right ) p(\omega_{n,l,d}) \right] \right ) \\
 & \propto \prod_{n}^N \prod_{l}^L \prod_{d}^{D_l} \exp \left ( \E \left [-\frac{\omega_{n,l,d} z_{n,l,d}^2}{2T^2}  \right] \right ) p(\omega_{n,l,d}). 
  \end{align*}
  Thus, they are independent across width, depth, and observations, with 
\begin{align*}
q(\omega_{n,l,d})&=\PG(\omega_{n,l,d}\mid 1, \frac{1}{T}\sqrt{ \E \left [z_{n,l,d}^2 \right ]} ) \\
&=  \PG(  \omega_{n,l,d} \mid 1, A_{n,l,d}),
\end{align*}
where
$$ A_{n,l,d} = \frac{1}{T} \sqrt{\left ( \Tr \left ( \E \left [ \widetilde{\bW}_{l,d}^T \widetilde{\bW}_{l,d} \right ] \E \left [ \widetilde{\ba}_{n,l-1} \widetilde{\ba}_{n,l-1}^T\right ] \right ) \right )}.$$

 \blu{The complexity of obtaining variational update for $\bomega$ is then $\mathcal{O}((NLD\max(D, D_0)^2))$. } 
\paragraph{Binary activation.} The variational posterior of the binary activations is:
\begin{align*}
    &q(\bgamma)  \propto \exp \left ( \E \left [\log \prod_{n}^N \prod_{l}^L \Norm \left( \ba_{n,l} \mid \bgamma_{n,l} \odot \bz_{n,l} , \bSigma_{l} \right) + \log \left (\prod_{n}^N \prod_{l}^L\prod_{d}^{D_l}  \exp(\frac{\gamma_{n,l,d} z_{n,l,d}}{T })  \right )  \right ] \right )  \\
    & \propto \prod_{n}^N\prod_{l}^L \prod_{l}^{D_l}\exp \left (  -\frac {1}{2\eta_{l,d}^{2} } \E \left [\left ( \blu{\mathrm{a}_{n,l,d}} - \gamma_{n,l,d} \left (\bW_{l,d} \ba_{n,l-1} + b_{l,d}   \right) \right )^2  \right ] %(\eta_{l,d})^{-2} \left ( \blu{\mathrm{a}_{n,l,d}} - \gamma_{n,l,d} \left (\bW_{l,d} \ba_{n,l-1} +  b_{l,d}  \right )  \right )  \right ]\right.  \\
   % & \left. 
   + \E \left[\frac{\gamma_{n,l,d} (\bW_{l,d} \ba_{n,l-1} + b_{l,d})}{T }\right ] \right).
\end{align*}
Therefore, the variational posterior $q(\bgamma)$ factories across observations $n = 1, \ldots, N$, layers $l = 1, \ldots, L,$ and dimensions of the layer $ d = 1, \ldots, D_l$, % as so that $q(\bgamma) =  \prod_{n}^N\prod_{l}^L \prod_{l}^{D_l} q(\gamma_{n,l,d}).$ We derive 
with each factor $q(\gamma_{n,l,d})$ given by:
\begin{align*}
  q(\gamma_{n,l,d}) \propto & \exp \left (  -\frac {1}{2} \E \left [ \eta_{l,d}^{-2} \right ] \left( \gamma_{n,l,d}^2 \E \left[ \left (\widetilde{\bW}_{l,d} \widetilde{\ba}_{n, l-1} \right )^2 \right ]  - 2 \gamma_{n,l,d}\E \left [\blu{\mathrm{a}_{n,l,d}} \widetilde{\bW}_{l,d} \widetilde{\ba}_{n, l-1} \right ] \right )  + \frac{1}{T} \gamma_{n,l,d}  \E \left [ \widetilde{\bW}_{l,d} \widetilde{\ba}_{n, l-1}\right ]  \right) \\
 \propto  & \exp \left (  \gamma_{n,l,d} \left ( -\frac {1}{2}  \E \left [ \eta_{l,d}^{-2} \right ]  \Tr \left (\E \left[\widetilde{\bW}_{l,d}^T \widetilde{\bW}_{l,d} \right ] \E \left [ \widetilde{\ba}_{n, l-1}\widetilde{\ba}_{n, l-1}^T  \right ] \right )   \right. \right.  \\
 & \left. \left.  + \E \left [ \eta_{l,d}^{-2} \right ] \E \left [ \widetilde{\bW}_{l,d} \right ] \E \left [ \widetilde{\ba}_{n, l-1} \blu{\mathrm{a}_{n,l,d}} \right ] + \frac{1}{T} \E \left [ \widetilde{\bW}_{l,d} \right ] \E \left [ \widetilde{\ba}_{n, l-1}\right ] \right ) \right )\\
& \propto  \exp \left (\gamma_{n,l,d} \sigma^{-1}\left(\rho_{n,l,d} \right) \right), 
\end{align*}
where $\sigma$ is the logistic function and
\begin{align*}
  \rho_{n,l,d} = &  \sigma \left(  \E \left [ \eta_{l,d}^{-2} \right ] \left(-\frac {1}{2}   \Tr \left (\E \left[\widetilde{\bW}_{l,d}^T \widetilde{\bW}_{l,d} \right ] \E \left [ \widetilde{\ba}_{n, l-1}\widetilde{\ba}_{n, l-1}^T  \right ] \right )  +  
 \E \left [ \widetilde{\bW}_{l,d} \right ] \E \left [ \widetilde{\ba}_{n, l-1} \blu{\mathrm{a}_{n,l,d}} \right ] \right) + \frac{1}{T} \E \left [ \widetilde{\bW}_{l,d} \right ] \E \left [ \widetilde{\ba}_{n, l-1}\right ] \right).
\end{align*}
Then noticing that $\sigma^{-1}(\rho) = \log(\rho(1-\rho)^{-1})$ and combining separate factors of the variational posterior of the binary activations, we obtain: 
\begin{align*}
     q(\bgamma) & \propto \prod_{n}^N\prod_{l}^L \prod_{d}^{D_l}\rho_{n,l,d}^{\gamma_{n,l,d}}\left(1-\rho_{n,l,d}\right)^{1-\gamma_{n,l,d}}\\
     & \propto \prod_{n}^N\prod_{l}^L \prod_{d}^{D_l} \Bern \left (\gamma_{n,l,d} \mid\rho_{n,l,d} \right ). 
\end{align*}

 \blu{The complexity of obtaining variational update for $\bgamma$ is then $\mathcal{O}(LND\max(D, D_0)^2)$. }

\paragraph{Stochastic activation.} The variational posterior of the stochastic activation is 
\begin{align*}
   & q(\ba) \propto \exp \left ( \E \left [ \log \prod_{n}^N \Norm\left(\by_n \mid \bz_{n,L+1}, \bmeta_{L+1}\right)  + \log \prod_{n}^N \prod_{l}^L \Norm \left( \ba_{n,l} \mid \bgamma_{n,l} \odot \bz_{n,l} , \bmeta_{l} \right) \right. \right.\\
    & + \left. \left.\log \prod_{n}^N \prod_{l}^L \prod_{d}^{D_l}  \exp(\frac{(\gamma_{n,l,d} - \frac{1}{2}) z_{n,l,d}}{T }) \exp(-\frac{\omega_{n,l,d}z_{n,l,d}^2}{2T^2}) \right ] \right ) \\
    % &\propto \prod_{n}^N\exp \left ( \E \left [ -\frac {1}{2} \left ( \by_n - \bW_{L+1} \ba_{n, L} - \bb_{L+1} \right )^{T} (\bSigma_{L+1})^{-1} \left ( \by_n - \bW_{L+1} \ba_{n, L} - \bb_{L+1} \right ) \right] \right) \times  \\
    % & \times \prod_{n}^N\prod_{l}^L \exp \left (\E \left [-\frac {1}{2} \left ( \ba_{n,l} - \bgamma_{n,l} \odot \left (\bW_{l} \ba_{n,l-1} + \bb_{l} \right )  \right )^{T} (\bSigma_{l})^{-1} \left ( \ba_{n,l} - \bgamma_{n,l} \odot  \left (\bW_{l}  \ba_{n,l-1} + \bb_{l}  \right ) \right ) \right]  \right) \times \\
    % & \times \prod_{n}^N\prod_{l}^L\exp \left(\sum_{d}^{D_l} \E \left [   \left (\frac{\left (\gamma_{n,l,d} - \frac{1}{2}\right ) z_{n,l,d}}{T }-\frac{\omega_{n,l,d}z_{n,l,d}^2}{2T^2}\right ) \right ] \right ) \\
    %  &\propto \prod_{n}^N \prod_{d}^{D_{L+1}} \exp \left (-\frac {1}{2} \E \left [  \frac{1}{\eta_{L+1,d}^2} \right ] \E \left [ \left ( y_{n,d} - \bW_{L+1,d} \ba_{n, L} - b_{L+1,d} \right )^{2} \right] \right) \times  \\
    % & \times \prod_{n}^N\prod_{l}^L \prod_{d}^{D_l} \exp \left (-\frac {1}{2} \E \left [\frac{1}{\eta_{l,d}^2} \right ] \E \left[\left ( \blu{\mathrm{a}_{n,l,d}} - \gamma_{n,l,d}\left (\bW_{l,d} \ba_{n,l-1} + b_{l,d} \right )  \right )^{2}  \right]  \right) \times \\
    % & \times \prod_{n}^N\prod_{l}^L \prod_{d}^{D_l} \exp \left( \E \left [  \frac{\left (\gamma_{n,l,d} - \frac{1}{2}\right ) z_{n,l,d}}{T}-\frac{\omega_{n,l,d}z_{n,l,d}^2}{2T^2} \right ] \right ) \\
     &\propto \prod_{n}^N \exp \left (-\frac {1}{2} \sum_{d}^{D_{L+1}} \E \left [  \frac{1}{\eta_{L+1,d}^2} \right ] \E \left [ \left ( y_{n,d} - \bW_{L+1,d} \ba_{n, L} - b_{L+1,d} \right )^{2} \right] \right)  \\
    & \times \prod_{n}^N\exp \left (-\frac {1}{2} \sum_{l}^L \sum_{d}^{D_l}   \E \left [\frac{1}{\eta_{l,d}^2} \right ] \E \left[\left ( \blu{\mathrm{a}_{n,l,d}} - \gamma_{n,l,d}\left (\bW_{l,d} \ba_{n,l-1} + b_{l,d} \right )  \right )^{2}  \right]  \right) \\
    & \times \prod_{n}^N \exp \left(  \sum_{l}^L\sum_{d}^{D_l}  \E \left [  \frac{\left (\gamma_{n,l,d} - \frac{1}{2}\right ) z_{n,l,d}}{T}-\frac{\omega_{n,l,d}z_{n,l,d}^2}{2T^2} \right ] \right ).
\end{align*}
Therefore, the variational posterior of the stochastic activations factories across observations $n = 1, \ldots, N$ and we derive $q(\ba_n)$ separately. 
For each layer $l =1, \ldots, L$, we  introduce the following diagonal matrix $\hat{\mathbf{\Sigma}}_{l}^{-1}=\diag \left ( \E \left[\eta_{l,1}^{-2} \right ], \dots, \E \left[\eta_{l,D_l}^{-2}  \right ]\right )$ and consider the relevant terms of the variational posterior:
\begin{align*}
 &q(\ba_n) \propto  \exp \left (-\frac {1}{2} \ba_{n,L}^T \left (  \sum_{d}^{D_{L+1}}\E \left [  \frac{1}{\eta_{L+1,d}^2}\right ]  \E \left [ \bW_{L+1,d}^T \bW_{L+1,d} \right] \ba_{n,L} \right )  \right )   \\
 &\times \exp \left ( -\ba_{n,L}^T \left ( \sum_{d}^{D_{L+1}} \E \left [  \frac{1}{\eta_{L+1,d}^2}\right ] \left (  \E \left [ \bW_{L+1,d}^T b_{L+1,d} \right]  -  \E \left [ \bW_{L+1,d}^T \right] y_{n,d} \right ) \right )  \right )  \\
 & \times \exp \left (-\frac {1}{2} \left (  \ba_{n,L}^T \hat{\mathbf{\Sigma}}_{L}^{-1}  \ba_{n,L} - 2  \ba_{n,L}^T \hat{\mathbf{\Sigma}}_{L}^{-1} \left ( \left ( \E \left[ \bgamma_{n,L}\right ]  \mathbf{1}^T_{D_{L-1}} \odot \E \left [  \bW_{L}\right ] \right ) \ba_{n, L-1} + \E \left[ \bgamma_{n,L}\right ] \odot \E \left [ \bb_{L}\right ]\right )  \right ) \right ) \\
 & \times \prod_{l=1}^{L-1} \exp \left ( -\frac {1}{2} \left (  \ba_{n,l}^T \hat{\mathbf{\Sigma}}_{l}^{-1}  \ba_{n,l} - 2  \ba_{n,l}^T \hat{\mathbf{\Sigma}}_{l}^{-1} \left ( \left ( \E \left[ \bgamma_{n,l}\right ]  \mathbf{1}^T_{D_{l-1}} \odot \E \left [  \bW_{l}\right ] \right ) \ba_{n, l-1} + \E \left[ \bgamma_{n,l}\right ] \odot \E \left [ \bb_{l}\right ]\right )  \right ) \right )  \\
 & \times \prod_{l=1}^{L}  \exp  \left ( -\frac {1}{2} \left (  \ba_{n,l-1}^T \left ( 
\sum_{d=1}^{D_{l}} \E \left [\frac{1}{\eta_{l,d}^2} \right ] \E \left[ \gamma_{n,l,d}\right ] \E \left [ \bW_{l,d}^T \bW_{l,d} \right ] \ba_{n,l-1} \right) \right )  \right ) \times \\
& \times \prod_{l=1}^{L}  \exp  \left (- \ba_{n,l-1}^T \left ( \sum_{d=1}^{D_{l}} \E \left [\frac{1}{\eta_{l,d}^2} \right ] \E \left[ \gamma_{n,l,d}\right ] \E \left [ \bW_{l,d}^T \bb_{l,d} \right ] \right ) \right )    \\
& \times \prod_{l=1}^{L}  \exp  \left ( -\frac {1}{2} \left ( \ba_{n,l-1}^T \left ( \frac{1}{T^2} \sum_{d=1}^{D_{l}}  \E \left [ \omega_{n,l,d}\right ] \E \left [ \bW_{l,d}^T \bW_{l,d} \right ] \right ) \ba_{n,l-1} 
\right ) \right )  \\
& \times \prod_{l=1}^{L}  \exp  \left (  \ba_{n,l-1}^T \left ( \frac{1}{T} \sum_{d=1}^{D_{l}}\E \left [ \bW_{l,d}^T \right ]  \left ( \E \left [ \gamma_{n,l,d} \right ]  - \frac{1}{2}\right ) - \frac{1}{T^2} \sum_{d=1}^{D_{l}}  \E \left [ \omega_{n,l,d}\right ]
\E \left [ \bW_{l,d}^T b_{l,d}\right ] 
\right ) \right ).
\end{align*}
The variational posterior of the stochastic activations does not factories into independent blocks, however it does have a structured sequential factorization $q(\ba_n)  = \prod_{l=1}^{L}  q(\ba_{n,l} \mid \ba_{n, l-1})$.And, we can derive the variational factor $q(\ba_{n, L} \mid \ba_{n, L-1})$ by only considering the terms with $\ba_{n, L}$. First, introduce the matrices $\mathbf{S}_{n, L}$ and $ \mathbf{M}_{n,L}$ and a vectors $\mathbf{t}_{n,L}$:
\begin{align*}
    & \mathbf{S}^{-1}_{n, L} = \hat{\mathbf{\Sigma}}_{L}^{-1} + \sum_{d=1}^{D_{L+1}}\E \left [  \frac{1}{\eta_{L+1,d}^2}\right ]  \E \left [ \bW_{L+1,d}^T \bW_{L+1,d} \right],  \\
    & \mathbf{t}_{n,L} = \mathbf{S}_{n,L} \left ( \left (\sum_{d=1}^{D_{l+1}} \E \left [\frac{1}{\eta_{L+1,d}^2} \right ] 
   \left ( - \E \left [ \bW_{L+1,d}^T b_{L+1,d} \right]  +  \E \left [ \bW_{L+1,d}^T \right] y_{n,d} \right ) \right )  + \hat{\mathbf{\Sigma}}_{l}^{-1}  \E \left[ \bgamma_{n,L}\right ] \odot \E \left [ \bb_{L}\right ] \right ), \\
   & \mathbf{M}_{n,L} = \mathbf{S}_{n,L} \hat{\mathbf{\Sigma}}_{L}^{-1} \E \left[ \bgamma_{n,L}\right ]  \mathbf{1}^T_{D_{L-1}} \odot \E \left [  \bW_{L}\right ].
\end{align*}
Then we consider relevant terms of the variational posterior:
\begin{align*}
    & q(\ba_{n, L} \mid \ba_{n, L-1}) \propto \exp \left (- \frac{1}{2} \left ( \ba_{n,L}^T \mathbf{S}_{n,L}^{-1} \ba_{n,L} - 2 \ba_{n,L}^T \mathbf{S}_{n,L}^{-1} \left ( \mathbf{t}_{n,L}+ \mathbf{M}_{n,L} \ba_{n,L-1} \right )  \right ) \right ) \\
    & \propto \exp \left (- \frac{1}{2} \left ( \ba_{n,L} - \left ( \mathbf{t}_{n,L}+ \mathbf{M}_{n,L} \ba_{n,L-1} \right )\right )^T \mathbf{S}_{n,L}^{-1 } \left ( \ba_{n,L} - \left ( \mathbf{t}_{n,L}+ \mathbf{M}_{n,L} \ba_{n,L-1} \right ) \right ) \right ) \times \\
    & \times \exp \left (\frac{1}{2} \left ( \mathbf{t}_{n,L}+ \mathbf{M}_{n,L} \ba_{n,L-1}\right )^T \mathbf{S}_{n,L}^{-1 } \left ( \mathbf{t}_{n,L}+ \mathbf{M}_{n,L} \ba_{n,L-1} \right ) \right ) \\
    & \propto \Norm \left (\ba_{n,L} \mid  \mathbf{t}_{n,L}+ \mathbf{M}_{n,L} \ba_{n,L-1}, \mathbf{S}_{n,L} \right ) \times \exp \left (\frac{1}{2} \left ( \mathbf{t}_{n,L}+ \mathbf{M}_{n,L} \ba_{n,L-1}\right )^T \mathbf{S}_{n,L}^{-1 } \left ( \mathbf{t}_{n,L}+ \mathbf{M}_{n,L} \ba_{n,L-1} \right ) \right ), 
\end{align*}
where the first term in the equation above provides $q(\ba_{n, L} \mid \ba_{n, L-1})$ and the second terms is relevant for computing the subsequent $q(\ba_{n, L-1} \mid \ba_{n, L-2})$. Recursively repeating a similar procedure for $l=L-1, \ldots, 1$, we are then able to obtain each of the variational posteriors $q(\ba_{n, l} \mid \ba_{n, l-1})$. Each time we define 
$\mathbf{S}_{n, l},  \mathbf{M}_{n,l}$ and $\mathbf{t}_{n, l}$ as follows:
\begin{align*}
\mathbf{S}^{-1}_{n, l} &  = \hat{\mathbf{\Sigma}}_l^{-1} - \mathbf{M}_{n, l+1}^T \mathbf{S}_{n, l+1}^{-1}\hat{\mathbf{M}}_{n, l+1} + \sum_{d=1}^{D_{l+1}} \left (  \E \left [\frac{1}{\eta_{l+1,d}^2} \right ] \E \left[ \gamma_{n,l+1,d}\right ] + \frac{1}{T^2} \sum_{d=1}^{D_{l+1}}  \E \left [ \omega_{n,l+1,d}\right ]\right) \E \left [ \bW_{l+1,d}^T \bW_{l+1,d} \right ]  \\
 \mathbf{t}_{n, l}  & = \mathbf{S}_{n,l} \left ( \mathbf{M}_{n, l+1}^T \mathbf{S}_{n, l+1}^{-1} \mathbf{t}_{n,l+1} + \hat{\mathbf{\Sigma}}_l^{-1}  \E \left[ \bgamma_{n,l}\right ] \odot  \E \left[ \bb_{l}\right ]  + \frac{1}{T} \sum_{d=1}^{D_{l+1}}\E \left [ \bW_{l+1,d}^T \right ]  \left ( \E \left [ \gamma_{n,l+1,d} \right ]  - \frac{1}{2}\right ) \right.  \\
& \left. - \sum_{d=1}^{D_{l+1}} \left(  \E \left [\frac{1}{\eta_{l+1,d}^2} \right ] \E \left[ \gamma_{n,l+1,d}\right ] + \frac{1}{T^2}  \E \left [ \omega_{n,l+1,d}\right ]\right ) \E \left [ \bW_{l+1,d}^T b_{l+1,d}\right ] \right ), \\
 \mathbf{M}_{n,l} &= \mathbf{S}_{n,l} \hat{\mathbf{\Sigma}}_l^{-1} \E \left[ \bgamma_{n,l}\right ]  \mathbf{1}^T_{D_{l-1}} \odot \E \left [  \bW_{l}\right ].
\end{align*}
Then substituting the above into the terms of the variational posterior containing $\ba_{n, l}$:
\begin{align*}
      q(\ba_{n, l} \mid \ba_{n, l-1}) & \propto \exp \left (- \frac{1}{2} \left ( \ba_{n,l} - \left ( \mathbf{t}_{n,l}+ \mathbf{M}_{n,l} \ba_{n,l-1} \right )\right )^T \mathbf{S}_{n,l}^{-1} \left ( \ba_{n,l} - \left ( \mathbf{t}_{n,l}+ \mathbf{M}_{n,l} \ba_{n,l-1} \right ) \right ) \right )  \\
    & \times \exp \left (\frac{1}{2} \left ( \mathbf{t}_{n,l}+ \mathbf{M}_{n,l} \ba_{n,l-1}\right )^T \mathbf{S}_{n,l}^{-1} \left ( \mathbf{t}_{n,l}+ \mathbf{M}_{n,l} \ba_{n,l-1} \right ) \right ) \\
   & \propto  \Norm \left (\ba_{n,l} \mid  \mathbf{t}_{n,l}+ \mathbf{M}_{n,l} \ba_{n,l-1}, \mathbf{S}_{n,l} \right ) \times \exp \left (\frac{1}{2} \left ( \mathbf{t}_{n,l}+ \mathbf{M}_{n,l} \ba_{n,l-1}\right )^T \mathbf{S}_{n,l}^{-1} \left ( \mathbf{t}_{n,l}+ \mathbf{M}_{n,l} \ba_{n,l-1} \right ) \right ).
\end{align*}

Finally, we combine the terms $q(\ba_{n, l} \mid \ba_{n, l-1})$ for $l=1, \ldots, L+1$ and get the variational posterior of the stochastic activation
\begin{align*}
   q(\ba)&  \propto  \prod_{n=1}^{N}  \prod_{l=1}^{L} \Norm \left (\ba_{n,l} \mid  \mathbf{t}_{n,l}+ \mathbf{M}_{n,l} \ba_{n,l-1}, \mathbf{S}_{n,l} \right ).
\end{align*}

\section{ELBO computation} \label{appendix:ELBO}

\subsection{ELBO for training}

Recall that optimal variational parameters maximize the ELBO function of \cref{eqn:ELBO}, which for our model is: 
\begin{align*}
& \text{ELBO} 
% =
% \E_{q(\btheta) } \left [ \log \left( \frac{p(\btheta,  \mathcal{D})  }{q(\btheta)}  \right)\right ] = \\
 = \E \left [  \log p (\by, \ba, \bgamma,\bomega | \bW, \bb, \bSigma)\right ] +  \E \left [  \log p (\bW | \bpsi, \btau)\right ] + \E \left [  \log p (\bpsi)\right ] +  \E \left [  \log p (\btau)\right ] +  \E \left [  \log p (\bb)\right ] +  \E \left [  \log p (\bSigma)\right ]  \\
 &  - \E \left [\log  q(\ba) \right ] - \E \left [\log q(\bgamma) \right ] - \E \left [\log q(\bomega)\right ]- \E \left [\log q(\bW,\bb)\right ] - \E \left [\log q(\bmeta)\right ]- \E \left [\log q(\bpsi)\right ]- \E \left [\log q(\btau)\right ].
\end{align*}

Similar to the variational update, we compute the terms of the ELBO corresponding to different blocks of parameters separately. 
 
% Consider a NN with $L$ layers, $D_0$-dimensional input and $D_{L+1}$-dimensional output. 
% Suppose additionally that shrinkage priors are inverse-gamma. 
% As before we consider parameters separately. 
%  the dimension of the hidden layer $l$ is $D_l$ and the number of observations is $N$.

%  Note on the notations: $q(\bb, \bW) \propto \prod_{l}^{L+1} \prod_d^{D_l} \Norm \left ((b_{l,d}, \bW_{l,d})^T \mid \mathbf{m}_{l,d}, \mathbf{B}_{l,d}\right )$, where $\mathbf{m}_{l,d} \in \R^{D_{l-1}+1}$ are row vectors and $\mathbf{B}_{l,d} \in \R^{D_{l-1}+1} \times \R^{D_{l-1}+1}$ are matrices. Then we assume that 
% \begin{align*}
%     & \mathbf{m}_{l,d} = \left (m^b_{l,d}, \left(\mathbf{m}^W_{l,d}\right) ^T  \right )^T, \\
%     & \mathbf{B}_{l,d} = \begin{pmatrix} B_{l,d}^{b} &  \mathbf{B}_{l,d}^{bW} \\ \mathbf{B}_{l,d}^{Wb} & \mathbf{B}_{l,d}^{W} \end{pmatrix}.
% \end{align*}

\paragraph{ELBO of $\btau$.} First, consider the terms of the ELBO containing the global shrinkage parameters:
\begin{align*}
  & \E \left [ \log p(\btau) - \log q (\btau)\right ]  =  \sum_{l=1}^{L+1} \E \left [ \log   \GIG\left( \tau_l \mid \nu_{\glob}, \delta_{\glob}, \lambda_{\glob} \right) - \log  \GIG \left (\tau_l | \hat{\nu}_{\glob,l}, \hat{\delta}_{\glob,l}, \lambda_{\glob} \right ) \right ] \\
  & =C_{\tau} + \sum_{l=1}^{L+1}  \E \left [ \log \tau_{l}^{\nu_{\glob}-1} \exp\left( -\frac{1}{2} \left(\frac{\delta_{\glob}^2}{\tau_{l}} +\lambda_{\glob}^2\tau_{l}\right) \right) \right ]  - \sum_{l=1}^{L+1}  \E \left [ \log  \left ( \tau_{l}^{\hat{\nu}_{\glob,l}-1}\right) \exp \left( -\frac{1}{2} \left(\frac{\hat{\delta}_{\glob,l}^2}{\tau_{l} } +\lambda_{\glob}^2\tau_{l} \right) \right ) \right ]  \\
 & = C_{\tau} +\frac{1}{2}\sum_{l=1}^{L+1} D_l D_{l-1}\E \left [ \log \tau_{l} \right ] + \E \left [ \frac{1}{\tau_{l}} (\hat{\delta}_{\glob,l}^2 - \delta_{\glob}^2)  \right ],   \end{align*}
where the normalizing constant is 
\begin{align*}
    C_{\tau} & = \sum_{l=1}^{L+1} (\nu_{\glob}-\hat{\nu}_{\glob,l})\log(\lambda_{\glob})+\hat{\nu}_{\glob,l}\log(\hat{\delta}_{\glob,l}) -\nu_{\glob}\log(\delta_{\glob})  \\
     & + \sum_{l=1}^{L+1}\log(K_{\hat{\nu}_{\glob,l}}(\lambda_{\glob}\hat{\delta}_{\glob,l})) - \log(K_{\nu_{\glob}}(\lambda\delta_{\glob})).  
\end{align*}

\paragraph{ELBO of $\bpsi$.} Similarly, the terms of the ELBO containing the local shrinkage parameters are
\begin{align*}
&\E \left [ \log p (\bpsi) - \log q(\bpsi)\right ] = C_{\bpsi}+  \sum_{l=1}^{L+1} \sum_{d=1}^{D_l} \sum_{d'=1}^{D_{l-1}} \E \left [ \log \GIG\left( \psi_{l,d,d'} \mid \nu_{\loc, l}, \delta_{\loc, l}, \lambda_{\loc, l} \right) - \log \GIG \left (\psi_{l,d,d'} | \hat{\nu}_{\loc,l,d,d'},\hat{\delta}_{\loc,l,d,d'}, \lambda_{\loc, l} \right ) \right ] \\
&=  C_{\bpsi} +  \sum_{l=1}^{L+1} \sum_{d=1}^{D_l} \sum_{d'=1}^{D_{l-1}} \E \left [ \log \psi_{l,d,d'}^{\nu_{\loc, l}-1} \exp \left( -\frac{1}{2} \left(\frac{\delta_{\loc, l}^2}{\psi_{l,d,d'}} + \lambda_{\loc, l}^2\psi_{l,d,d'}\right) \right) - \log
\left(\psi_{l,d,d'}^{\hat{\nu}_{\loc,l,d,d'}-1}\right) \exp \left( -\frac{1}{2} \left(
\frac{\hat{\delta}_{\loc,l,d,d'}^2}{\psi_{l,d,d'}} +  \lambda_{\loc, l} ^2 \psi_{l,d,d'}\right) \right )\right ] \\
&= C_{\bpsi} + \frac{1}{2}\sum_{l=1}^{L+1} \sum_{d=1}^{D_l} \sum_{d'=1}^{D_{l-1}} \E \left [\log \psi_{l,d,d'} \right ] + \E \left [  
\frac{1}{\psi_{l,d,d'}}  \right ]\left(\hat{\delta}_{\loc,l,d,d'}^2 - \delta_{\loc, l}^2\right), \end{align*}
where the normalizing constant is 
\begin{align*}
     C_{\bpsi}= &\sum_{l=1}^{L+1} \sum_{d=1}^{D_l} \sum_{d'=1}^{D_{l-1}}  \left( \nu_{\loc, l} - \hat{\nu}_{\loc,l,d,d'}\right)\log(\lambda_{\loc, l})+\hat{\nu}_{\loc,l,d,d'}\log(\hat{\delta}_{\loc,l,d,d'}) -\nu_{\loc, l}\log(\delta_{\loc, l})  \\
     & + \sum_{l=1}^{L+1}  \sum_{d=1}^{D_l} \sum_{d'=1}^{D_{l-1}} \log(K_{\hat{\nu}_{\loc,l,d,d'}}(\lambda_{\glob}\hat{\delta}_{\loc,l,d,d'})) - \log(K_{\nu_{\loc, l}}(\lambda_{\loc, l}\delta_{\loc, l})).
\end{align*}

\paragraph{ELBO of $\bmeta$.} As before, the covariance matrix is assumed to be diagonal so that the relevant ELBO is:
\begin{align*}
      &\E \left [ \log p (\bSigma) - \log q(\bmeta)\right ] \\
      &=C_{\bmeta} + \sum_{l=1}^{L} \sum_{d=1}^{D_l} \E \left [ \log \IG(\eta^2_{l,d} | \alpha^h_0,\beta^h_0) \right ]  +  \sum_{d=1}^{D_{L+1}}  \E \left [ \log \IG(\eta^2_{l,d} | \alpha_0,\beta_0) \right ] - \sum_l^{L+1} \sum_d^{D_l} \E \left [ \log \IG(\eta_{l,d}^2 \mid \alpha_{l,d}, \beta_{l,d}) \right ]  \\
    % &=  C_{\bmeta} + \sum_{l=1}^{L} \sum_{d=1}^{D_l} \E \left [ \log 
    % \eta_{l,d}^{-2\alpha^h_0-2} \exp \left( -\frac{\beta^h_0}{\eta_{l,d}^2}  \right) \right ]  + \\ 
    % &+ \sum_{d=1}^{D_{L+1}} \E \left [ \log \eta_{L+1,d}^{-2\alpha_0-2} \exp \left( -\frac{\beta_0}{\eta_{L+1,d}^2}  \right) \right ]  - \sum_l^{L+1} \sum_d^{D_l} \E \left [ \log \eta_{l,d}^{-2\alpha_{l,d}-2} \exp \left( -\frac{\beta_{l,d}}{\eta_{l,d}^{2}}  \right)  \right ] = \\
    &=  C_{\bmeta} + \sum_l^{L} \sum_d^{D_l}\left( \alpha_{l,d} - \alpha^h_0 \right) \E \left [ \log \eta_{l,d}^2 \right ]  +
    \sum_{d=1}^{D_{L+1}} \left( \alpha_{L+1,d} - \alpha_0 \right)\E \left [ \log \eta_{L+1,d}^2 \right ]  \\ 
    &+   \sum_{l=1}^{L} \sum_{d=1}^{D_l} \left( \beta_{l,d}- \beta^h_0 \right) \E \left [  \frac{1}{\eta_{l,d}^2}  \right ]  
    +  \sum_{d=1}^{D_{L+1}}  \left( \beta_{L+1,d}- \beta_0\right)  \E \left [ \frac{1}{\eta_{L+1,d}^2}   \right ]   \\
     & =  C_{\bmeta} + \frac{N}{2} \sum_l^{L+1} \sum_d^{D_l} \E \left [ \log \eta_{l,d}^2 \right ] +   \sum_{l=1}^{L} \sum_{d=1}^{D_l} \left( \beta_{l,d}- \beta^h_0 \right) \E \left [  \frac{1}{\eta_{l,d}^2}  \right ]  +  \sum_{d=1}^{D_{L+1}}  \left( \beta_{L+1,d}- \beta_0\right)  \E \left [ \frac{1}{\eta_{L+1,d}^2}   \right ], 
\end{align*}
where the normalizing constant is 
\begin{align*}
    C_{\bmeta}& = \sum_{l=1}^{L} \sum_{d=1}^{D_l}  \alpha^h_0  \log \beta^h_0 -\alpha_{l,d} \log \beta_{l,d} + \log \Gamma (\alpha_{l,d})  - \log \Gamma (\alpha^h_0)     \\
    & + \sum_{d=1}^{D_{L+1}}  \alpha_0 \log \beta_0  -\alpha_{L+1,d} \log \beta_{L+1,d}+ \log \Gamma (\alpha_{L+1,d}) - \log \Gamma (\alpha_0).
\end{align*}

\paragraph{ELBO of  $(\bW, \bb)$.} 
Recall, previously introduced matrices $\bD_{l,d}= \diag \left ( s_0^{-2},  \E \left [ \tau_l^{-1}\right]  \E \left[\psi_{l,d,1}^{-1}\right ] ,\dots,  \E \left[\tau_l^{-1} \right]\E \left[\psi_{l, d,D_{l-1}}^{-1}\right ]\right )$ and denote further $\bD_{l,d}^0= \diag \left ( s_0^{2},  \tau_l \psi_{l,d,1},\dots, \tau_l \psi_{l, d,D_{l-1}}\right)$. Then the ELBO of weights and biases is:
\begin{align*}
    & \E \left [  \log p (\bW | \bpsi, \btau)\right ] + \E \left [  \log p (\bb)\right ] - \E \left [\log q(\bW,\bb)\right ]  \\
    &= \sum_{l=1}^{L+1} \sum_{d=1}^{D_l} \sum_{d'=1}^{D_{l-1}} \E \left [ \log \Norm\left(\widetilde{\bW}_{l,d} |  0, \bD_{l,d}^0 \right)  \right ] - \sum_{l}^{L+1} \sum_d^{D_l} \E \left [ \log \Norm \left (\widetilde{\bW}_{l,d} \mid \mathbf{m}_{l,d}, \mathbf{B}_{l,d}\right ) \right ]   \\
    & = \sum_{l=1}^{L+1} \sum_{d=1}^{D_l} \E \left [ \log \left(|\bD_{l,d}|\right )^{-\frac{1}{2}}
\exp \left( -\frac{1}{2} \widetilde{\bW}_{l,d} \left( \bD_{l,d}^0\right )^{-1} \widetilde{\bW}_{l,d}^T\right) \right ]  \\
    &  - \sum_{l}^{L+1} \sum_d^{D_l} \E \left [ \log |\mathbf{B}_{l,d}|^{-\frac{1}{2}} \exp \left (- \frac{1}{2} 
    \left (\widetilde{\bW}_{l,d}- \mathbf{m}_{l,d} \right ) \mathbf{B}_{l,d}^{-1}\left (\widetilde{\bW}_{l,d} - \mathbf{m}_{l,d} \right )^T\right )  \right ]   \\
      & = \frac{1}{2} \sum_{l}^{L+1} \sum_d^{D_l} \E \left [ \log |\mathbf{B}_{l,d}| \right ] - \E \left [ \log \left(|\bD_{l,d}^0|\right )\right ] - \E \left [
 \widetilde{\bW}_{l,d}  (\bD_{l,d}^0)^{-1} \widetilde{\bW}_{l,d}^T \right ]  +  \E \left [ 
    \left (\widetilde{\bW}_{l,d}- \mathbf{m}_{l,d} \right ) \mathbf{B}_{l,d}^{-1}\left (\widetilde{\bW}_{l,d} - \mathbf{m}_{l,d} \right )^T \right ] \\
    & = \frac{1}{2} \sum_{l}^{L+1} \sum_d^{D_l} \log |\mathbf{B}_{l,d}| - \E \left[ \log \left(|\bD_{l,d}^0|\right ) \right ] - \Tr \left( \E \left [\widetilde{\bW}_{l,d}^T
 \widetilde{\bW}_{l,d} \right ]  \E \left [ (\bD_{l,d}^0)^{-1}  \right ] \right)  +   \sum_{l}^{L+1} \frac{D_l}{2}      \\
& = \frac{1}{2} \sum_{l}^{L+1} \sum_d^{D_l} \left(  \log |\mathbf{B}_{l,d}| -  \Tr \left( \E \left [\widetilde{\bW}_{l,d}^T \widetilde{\bW}_{l,d} \right ] \bD_{l,d} \right) - \sum_{d'=1}^{D_{l-1}}  \E\left [\log \psi_{l,d,d'} \right ]  \right) 
     -   \frac{1}{2}  \sum_{l}^{L+1} D_l \left( \log s_0^{2} + D_{l-1} \E\left [\log \tau_l\right ] - 1\right). 
\end{align*}
% where
%  \begin{align*}
% & \E \left [\widetilde{\bW}_{l,d}^T\widetilde{\bW}_{l,d} \right ] = \mathbf{m}_{l,d}^T\mathbf{m}_{l,d}+  \mathbf{B}_{l,d}.
% \end{align*}

\paragraph{ELBO of  $\ba$, $\bgamma$ and $\bomega$.} 
The remaining terms of the ELBO are the ones with stochastic and binary activations and additional augmented variables:
\begin{align*}
&\E \left [  \log p (\by, \ba, \bgamma,\bomega | \bW, \bb, \bSigma)\right ]   - \E \left [\log  q(\ba) \right ] - \E \left [\log q(\bgamma) \right ] - \E \left [\log q(\bomega)\right ]  \\
&= \sum_{n=1}^N \sum_{d=1}^{D_{L+1}} \E \left [ \log  \Norm\left( y_{n,d} \mid \bz_{n,L+1,d}, \bSigma_{L+1,d}\right) \right ] +  \sum_{n=1}^N  \sum_{l=1}^L \sum_{d=1}^{D_l} \E \left [ \log \Norm \left( \blu{\mathrm{a}_{n,l,d}} \mid \bgamma_{n,d} \odot \bz_{n,l,d} , \bSigma_{l,d} \right) \right ]   \\
 &+ \sum_{n=1}^N  \sum_{l=1}^L \sum_{d=1}^{D_l} \E \left [ \log \left ( \exp \left (\frac{\kappa_{n,l,d} z_{n,l,d}}{T}\right ) \exp \left (-\frac{\omega_{n,l, d}z_{n,l,d}^2}{2T^2} \right ) \PG(\omega_{n,l,d} \mid 1,0) \right )  \right ]  \\
 &-  \sum_{n=1}^N \sum_{l=1}^L \E \left [ \log \Norm \left (\ba_{n,l} \mid  \bt_{n,l} + \bM_{n,l} \ba_{n,l-1}, \bS_{n,l} \right ) \right ] - \sum_{n=1}^N \sum_{l=1}^L \sum_{d=1}^{D_l} \E \left [ \log \Bern\left (\gamma_{n,l,d} \mid \rho_{n,l,d} \right ) \right ] + \E \left [ \log \PG(  \omega_{n,l,d} \mid 1, A_{n,l,d}) \right ]  \\
&= \sum_{n=1}^N \sum_{d=1}^{D_{L+1}} \E \left [ \log  (\eta_{L+1,d}^2)^{-1/2} \exp \left(-\frac {1}{2\eta_{L+1,d}^{2} } \left ( y_{n,d} - \bW_{L+1,d} \ba_{n, L} - b_{L+1,d} \right )^{2}  \right)  \right] - \frac{N D_{L+1}}{2} \log(2\pi) \\
&+  \sum_{n=1}^N  \sum_{l=1}^L \sum_{d=1}^{D_l} \E \left [ \log  (\eta_{l,d}^2)^{-1/2} \exp \left(-\frac {1}{2\eta_{l,d}^{2} } \left ( \blu{\mathrm{a}_{n,l,d}} - \gamma_{n,l,d} \odot \left ( \bW_{l,d} \ba_{n,l-1} + b_{l,d} \right )  \right )^{2} \right)  \right ]  - N \sum_{l=1}^{L} D_l \log(2)  \\
&  +  \frac{1}{T} \sum_{n=1}^N  \sum_{l=1}^{L}  \sum_{d=1}^{D_l}  \E \left [ \left(\gamma_{n,d} - \frac{1}{2} \right)  \left ( \bW_{l,d} \ba_{n,l-1} + b_{l,d} \right )\right ]  -\frac{1}{2T^2} \sum_{n=1}^N  \sum_{l=1}^{L}\sum_{d=1}^{D_l} \E \left [ \omega_{n,l,d} \left ( \bW_{l,d} \ba_{n,l-1} + b_{l,d} \right )^2 \right ] \\
&- \sum_{n=1}^N  \sum_{l=1}^{L} \E \left [ \log  |\bS_{n,l}|^{-\frac{1}{2}}  \exp \left (-\frac{1}{2} \left ( \ba_{n,l} -  \bt_{n,l}- \bM_{n,l} \ba_{n,l-1}  \right)^T \bS_{n,l}^{-1}\left ( \ba_{n,l} -  \bt_{n,l}- \bM_{n,l} \ba_{n,l-1}  \right)  \right) \right ] \\
& - \sum_{n=1}^N  \sum_{l=1}^{L} \sum_{d=1}^{D_l} \left ( \rho_{n,l,d}  \log  \rho_{n,l,d} + (1- \rho_{n,l,d} )\log (1- \rho_{n,l,d}) \right )  + \sum_{n=1}^N \sum_{l=1}^{L} \sum_{d=1}^{D_l} \E \left [ \log \frac{\PG(\omega_{n,l,d} \mid 1,0)}{\PG(  \omega_{n,l,d}  \mid 1, A_{n,d})}  \right ]  \\
=& - \frac{1}{2} \sum_{n=1}^N \sum_{d=1}^{D_{L+1}} \E \left [\frac {1}{\eta_{L+1,d}^{2}} \right] \left ( y_{n,d}^2 - 2y_{n,d}  \E \left [ \widetilde{\bW}_{L+1,d} \right]\E \left [ \widetilde{\ba}_{n, L} \right] +  \Tr \left (\E \left [  \widetilde{\bW}_{L+1,d}^T\widetilde{\bW}_{L+1,d}\right] \E \left [ \widetilde{\ba}_{n, L}\widetilde{\ba}_{n, L}^T  \right] \right )  \right )  \\
& -  \frac{1}{2} \sum_{n=1}^N  \sum_{l=1}^L \sum_{d=1}^{D_l} \E \left [ \frac {1}{\eta_{l,d}^{2} } \right ] \left( \E \left [  \blu{\mathrm{a}_{n,l,d}}^2 \right ]  - 2\E \left [ \gamma_{n,l,d} \right ] \E \left [  \widetilde{\bW}_{l,d}  \right ] \E \left [ \widetilde{\ba}_{n,l-1} \blu{\mathrm{a}_{n,l,d}}  \right ]  \right)   \\
& -  \frac{1}{2} \sum_{n=1}^N  \sum_{l=1}^L \sum_{d=1}^{D_l} \E \left [ \frac {1}{\eta_{l,d}^{2} } \right ]  \E \left [ \gamma_{n,l,d}^2 \right ] \Tr \left(  \E \left [  \widetilde{\bW}_{l,d}^T  \widetilde{\bW}_{l,d} \right ]   \E \left [  \widetilde{\ba}_{n,l-1}\widetilde{\ba}_{n,l-1}^T  \right ] \right)   \\
&-\frac{N}{2} \sum_{d=1}^{D_{L+1}} \E \left [ \log  \eta_{L+1,d}^{2}   \right]   - \frac{N}{2} \sum_{l=1}^{L} \sum_{d=1}^{D_l} \E \left [ \log  \eta_{l,d}^{2}   \right ] + \frac{1}{2} \sum_{n=1}^{N}\sum_{l=1}^{L} \log(|  \bS_{n,l} |)  \\
&+  \sum_{n=1}^N \sum_{l=1}^{L} \sum_{d=1}^{D_l}\left( \frac{1}{T} \left(\rho_{n,l,d} - \frac{1}{2} \right) \E \left [  \widetilde{\bW}_{l,d} \right ] \E \left [ \widetilde{\ba}_{n,l-1} \right]  -\frac{1}{2T^2}  \E \left [ \omega_{n,l,d} \right ] \left ( \Tr \left( \E \left [\widetilde{\bW}_{l,d}^T\widetilde{\bW}_{l,d} \right ] \E \left [ \widetilde{\ba}_{n,l-1}\widetilde{\ba}_{n,l-1}^T\right]\right) \right ) \right) \\
&- \sum_{n=1}^N \sum_{l=1}^{L} \sum_{d=1}^{D_l} \left ( \rho_{n,l,d}  \log  \rho_{n,l,d} + (1- \rho_{n,l,d} )\log (1- \rho_{n,l,d}) - \frac{A_{n,l,d}^2}{2} \E \left [  \omega_{n,l,d}  \right ] + \log(\cosh(\frac{A_{n,l,d}}{2})) \right )  + C_a, 
\end{align*}
where the normalizing constant is 
\begin{align*}
    C_a =  - \frac{N D_{L+1}}{2} \log(2\pi)   - N \sum_{l=1}^{L} D_l \log(2).
\end{align*}
% \begin{align*}
%     &  \E \left [  \omega_{n,l,d}  \right ] =  \frac{\exp\left( A_{n,l,d} \right) -1}{  2A_{n,l,d}\left(\exp\left( A_{n,l,d}\right) + 1 \right)},  \\
%     &   \E \left [ \bW_{l,d}^T \bW_{l,d} \right ] = \left( \mathbf{m}_{l,d}^W\right) ^T\mathbf{m}_{l,d}^W+  \mathbf{B}_{l,d}^W, \quad
%      \E \left [  b_{L+1, d} \bW_{L+1,d} \right ] = \mathbf{m}_{l,d}^b \mathbf{m}_{l,d}^W + 
%  \mathbf{B}_{l,d}^{bW}, \\
% &  \E \left [ b_{l,d}^2 \right ] =  \left( \mathbf{m}_{l,d}^b\right)^2  + \mathbf{B}_{l,d}^b, \quad
%     \E \left [ \bW_{l, d} \right ] = \mathbf{m}_{l,d}^W, \quad
%       \E \left [ b_{l,d} \right ] = \mathbf{m}_{l,d}^b. \\
% \end{align*}

\paragraph{Total ELBO} Then, we can sum the derived parts above to get the total ELBO of our model:
\begin{align*}
&\text{ELBO} =\text{const.} + \sum_{l=1}^{L+1} \frac{1}{2}\E \left [   \frac{1}{\tau_{l}}  \right ] \left(\hat{\delta}_{\glob,l}^2 -  \delta_{\glob}^2\right) + (\hat{\nu}_{\glob,l}\log(\hat{\delta}_{\glob,l})  + \log(K_{\hat{\nu}_{\glob,l}}(\lambda_{\glob}\hat{\delta}_{\glob,l})) \\
&+ \sum_{l=1}^{L+1} \sum_{d=1}^{D_l} \sum_{d'=1}^{D_{l-1}} \frac{1}{2}\E \left [  \frac{1}{\psi_{l,d,d'}}  \right ]\left(\hat{\delta}_{\loc,l,d,d'}^2 - \delta_{\loc, l}^2\right) + \hat{\nu}_{\loc,l,d,d'}\log(\hat{\delta}_{\loc,l,d,d'}) + \log(K_{\hat{\nu}_{\loc,l,d,d'}}(\lambda_{\loc, l}\hat{\delta}_{\loc,l,d,d'})) \\ 
&+ \sum_{l=1}^{L+1}\sum_{d=1}^{D_{l}}\E \left [ \frac{1}{\eta_{l,d}^2}   \right ] \left(\beta_{l,d} - \beta^l_0 \right) - \alpha_{l,d} \log \beta_{l,d} + \frac{1}{2}  \log |\mathbf{B}_{l,d}| -\frac{1}{2} \left( \frac{1}{s_0^2}\E[b_{l,d}^2] + \sum_{d'=1}^{D_{l-1}} \E \left [   \frac{1}{\tau_{l}}  \right ] \E \left [  
\frac{1}{\psi_{l,d,d'}}  \right ] \E[w_{l,d,d'}^2]  \right)\\
&+ \frac{1}{2} \sum_{n=1}^N \sum_{l=1}^L\log(\bS_{n,l}) - \frac{1}{2}   \sum_{d=1}^{D_y}  \E \left [   \frac{1}{\eta_{L+1,d}^2}  \right ] \left( \sum_{n=1}^{N}  \E \left [\left(y_{n,d} - \E \left [\tilde{\bW}_{L+1,d} \right] \E \left [\tilde{\ba}_{n,L} \right]\right)^2 \right]  \right) \\
& -\frac{1}{2}   \sum_{d=1}^{D_y}  \E \left [   \frac{1}{\eta_{L+1,d}^2}  \right ] \left( \sum_{n=1}^{N} \Tr \left( \left(\mathbf{B}_{L+1,d} + \mathbf{m}_{L+1,d}\mathbf{m}_{L+1,d}^T \right) \E \left [ \tilde{\ba}_{n,L}\tilde{\ba}_{n,L}^T \right ]\right) - \Tr \left( \mathbf{m}_{L+1,d}\mathbf{m}_{L+1,d}^T \E \left [ \tilde{\ba}_{n,L} \right ]\E \left[ \tilde{\ba}_{n,L}^T \right ]\right)  \right) \\
 &- \frac{1}{2}  \sum_{l=1}^L \sum_{d=1}^{D_l}  \E \left [   \frac{1}{\eta_{1,d}^2}  \right ] \left( \sum_{n=1}^{N}   \left(\rho_{n,l, d} \E \left[\tilde{\bW}_{l,d} \right]\E \left[\tilde{\ba}_{n,l-1}\right] - \E\left [\ba_{n,l,d} \right]\right)^2  +  \E\left [\ba_{n,l,d}^2 \right] - \E \left[\ba_{n,l,d} \right]^2 \right)\\
  &- \frac{1}{2}  \sum_{l=1}^L \sum_{d=1}^{D_l}  \E \left [   \frac{1}{\eta_{1,d}^2}  \right ] \left( \sum_{n=1}^{N}  \rho_{n,l, d}  \Tr \left ( \left(\mathbf{B}_{l,d} + \mathbf{m}_{l,d}\mathbf{m}_{l,d}^T \right) \E \left[\tilde{\ba}_{n,l-1} \tilde{\ba}_{n,l-1}^T \right]  \right)- \rho_{n,l, d}^2\Tr \left( \mathbf{m}_{l,d}\mathbf{m}_{l,d}^T \E \left[ \tilde{\ba}_{n,l-1} \right]  \E \left[ \tilde{\ba}_{n,l-1}^T \right]\right) \right) \\
& -  \sum_{l=1}^L \sum_{d=1}^{D_l}  \E \left [   \frac{1}{\eta_{1,d}^2}  \right ] \left( \sum_{n=1}^{N}   \rho_{n,l, d}\E \left[\tilde{\bW}_{l,d} \right]\left( \E \left[ \ba_{n,l,d} \right] \E \left[\tilde{\ba}_{n,l-1}\right] -  \E \left[ \ba_{n,l,d}   \tilde{\ba}_{n,l-1} \right ]\right) \right) \\
&+ \sum_{n=1}^N \sum_{l=1}^L \sum_{d=1}^{D_l}  \frac{1}{T}\left(\rho_{n,l,d} - \frac{1}{2} \right)  \left (\E \left [  \tilde{\bW}_{l,d} \right ] \E \left [\tilde{\ba}_{n,l-1} \right]\right ) -\frac{1}{2T^2} \E \left [ \omega_{n,l,d} \right ] \left ( \Tr \left(\left(\mathbf{B}_{l,d} + \mathbf{m}_{l,d}\mathbf{m}_{l,d}^T \right) \E \left [\tilde{\ba}_{n,l-1}\tilde{\ba}_{n,l-1}^T \right] \right)  \right )\\
&- \sum_{n=1}^N \sum_{l=1}^L \sum_{d=1}^{D_l} \left ( \rho_{n,l,d}  \log  \rho_{n,l,d} + (1- \rho_{n,l,d} )\log (1- \rho_{n,l,d}) \right )- \frac{A_{n,l,d}^2}{2} \E \left [  \omega_{n,l,d}  \right ]  + \log(\cosh(A_{n,l,d}/2)).
\end{align*}

Note that when implementing VI with EM scheme, we adjust the formula above by adding the term which arises in the normalizing constant  $C_{\tau}$ defined when computing the ELBO of global shrinkage parameters, specifically, we add
\begin{align*}
 \text{ELBO}_{EM}  & = (L+1)\left( \nu_{\glob} \left( \log (\lambda_{\glob}) - \log (\delta_{\glob}) \right) - \log \left( K_{\nu_{\glob}}(\lambda_{\glob}\delta_{\glob})\right) \right)   \\
     &+\sum_{l=1}^{L+1}(\nu_{\glob} - 1)\E \left [ \log \tau_{l}\right ] - \frac{1}{2}\lambda_{\glob}^2 \E\left [\tau_{l}\right ]  -  \nu_{ l}  \log (\lambda_{\glob}).
% &+ D_{l}D_{l-1} \left( \nu_{\loc, l} \left( \log (\lambda_{\loc, l}) - \log (\delta_{\loc, l}) \right) - \log \left( K_{\nu_{\loc, l}}(\lambda_{\loc, l}\delta_{\loc, l})\right) \right) + \\
%      &+\sum_{d=1}^{D_{l}}\sum_{d'=1}^{D_{l-1}}(\nu_{\loc, l} - 1)\E \left [ \log \psi_{\loc, l,d,d'}\right ] - \frac{1}{2}\ \lambda_{\loc, l}^2 \E\left [\psi_{\loc, l,d,d'}\right ] -  \nu_{\loc, l,d,d'}  \log (\lambda_{\loc, l}) 
\end{align*}

 \subsection{ELBO for prediction} \label{sec:ELBOforpred}
To obtain the posterior predictive distribution, we compute the approximate variational predictive distributions of  $\ba_{*}$, $\bgamma_{*}$ and $\bomega_{*}$ with the objective function being the ELBO of \cref{eqn:ELBO}. Thus, in the predictive step of our algorithm, we monitor the convergence of the ELBO of $\ba_{*}$, $\bgamma_{*}$ and $\bomega_{*}$, which we derive as follows: 
\begin{align*}
&\E \left [  \log p (\ba_*, \bgamma_*,\bomega_* | \bW, \bb, \bSigma)\right ]   - \E \left [\log  q(\ba_*) \right ] - \E \left [\log q(\bgamma_*) \right ] - \E \left [\log q(\bomega_*)\right ]  \\
&=   \sum_{l=1}^L \sum_{d=1}^{D_l} \E \left [ \log \Norm \left( a_{*,l,d} \mid \bgamma_{*,d} \odot \bz_{*,l,d} , \bSigma_{l,d} \right) \right ]  +  \E \left [ \log \left ( \exp \left (\frac{\kappa_{*,l,d} z_{*,l,d}}{T}\right ) \exp \left (-\frac{\omega_{*,l, d}z_{*,l,d}^2}{2T^2} \right ) \PG(\omega_{n,l,d} \mid 1,0) \right )  \right ] - \\
 &-  \sum_{l=1}^L \E \left [ \log \Norm \left (\ba_{*,l} \mid  \bt_{*,l} + \bM_{*,l} \ba_{*,l-1}, \bS_{*,l} \right ) \right ] - \sum_{l=1}^L \sum_{d=1}^{D_l} \left( \E \left [ \log \Bern\left (\gamma_{*,l,d} \mid \rho_{*,l,d} \right ) \right ] +  \E \left [ \log \PG(  \omega_{*,l,d} \mid 1, A_{*,l,d}) \right ] \right) \\
% &=  \sum_{l=1}^L \sum_{d=1}^{D_l} \E \left [ \log  (\eta_{l,d}^2)^{-1/2} \exp \left(-\frac {1}{2\eta_{l,d}^{2} } \left ( a_{*,l,d} - \gamma_{*,l,d} \odot \left ( \bW_{l,d} \ba_{*,l-1} + b_{l,d} \right )  \right )^{2} \right)  \right ]  -  \sum_{l=1}^{L} D_l \log(2)  \\
% &  +   \sum_{l=1}^{L}  \sum_{d=1}^{D_l}  \frac{1}{T}  \E \left [ \left(\gamma_{*,d} - \frac{1}{2} \right)  \left ( \bW_{l,d} \ba_{*, l-1} + b_{l,d} \right )\right ]  -\frac{1}{2T^2}   \E \left [ \omega_{*,l,d} \left ( \bW_{l,d} \ba_{*,l-1} + b_{l,d} \right )^2 \right ] \\
% &-  \sum_{l=1}^{L} \E \left [ \log  |\bS_{*,l}|^{-\frac{1}{2}}  \exp \left (-\frac{1}{2} \left ( \ba_{*,l} -  \bt_{*,l}- \bM_{*,l} \ba_{*,l-1}  \right)^T \bS_{*,l}^{-1}\left ( \ba_{*,l} -  \bt_{*,l}- \bM_{*,l} \ba_{*,l-1}  \right)  \right) \right ] - \\
% & -   \sum_{l=1}^{L} \sum_{d=1}^{D_l} \left ( \rho_{*,l,d}  \log  \rho_{*,l,d} + (1- \rho_{*,l,d} )\log (1- \rho_{*,l,d}) \right )  +  \sum_{l=1}^{L} \sum_{d=1}^{D_l} \E \left [ \log \frac{\PG(\omega_{*,l,d} \mid 1,0)}{\PG(  \omega_{*,l,d}  \mid 1, A_{*,d})}  \right ]  \\
=&  -  \frac{1}{2}  \sum_{l=1}^L \sum_{d=1}^{D_l} \E \left [ \frac {1}{\eta_{l,d}^{2} } \right ] \left( \E \left [  a_{*,l,d}^2 \right ]  - 2\E \left [ \gamma_{*,l,d} \right ] \E \left [  \tilde{\bW}_{l,d}  \right ] \E \left [ \tilde{\ba}_{*,l-1} a_{*,l,d}  \right ] \right)  \\
&   - \frac{1}{2} \sum_{l=1}^{L} \sum_{d=1}^{D_l} \E \left [ \log  \eta_{l,d}^{2}   \right ]   +  \frac{1}{2}\sum_{l=1}^{L} \log(|  \bS_{*,l} |)  + \frac{1}{T} \sum_{l=1}^{L} \sum_{d=1}^{D_l}  \left(\rho_{*,l,d} - \frac{1}{2} \right) \E \left [  \widetilde{\bW}_{l,d} \right ] \E \left [ \widetilde{\ba}_{*,l-1} \right]   \\
& -  \frac{1}{2}  \sum_{l=1}^L \sum_{d=1}^{D_l} \left( \E \left [ \frac {1}{\eta_{l,d}^{2} } \right ] \E \left [ \gamma_{*,l,d}^2 \right ] +  \frac{1}{T^2}\E \left [ \omega_{*,l,d} \right ] \right)   \Tr \left(  \E \left [  \tilde{\bW}_{l,d}^T  \tilde{\bW}_{l,d} \right ]   \E \left [  \tilde{\ba}_{*,l-1}\tilde{\ba}_{*,l-1}^T  \right ] \right)    \\
&-  \sum_{l=1}^{L} \sum_{d=1}^{D_l} \left ( \rho_{*,l,d}  \log  \rho_{*,l,d} + (1- \rho_{*,l,d} )\log (1- \rho_{*,l,d}) \right )  \\
&+  \sum_{l=1}^{L}  \sum_{d=1}^{D_l} \frac{A_{*,l,d}^2}{2} \E \left [  \omega_{*,l,d}  \right ]  - \sum_{l=1}^{L}  \sum_{d=l}^{D_l} \log(\cosh(\frac{A_{*,l,d}}{2})) + \text{const}\\
=&- \frac{1}{2}  \sum_{l=1}^L \sum_{d=1}^{D_l}  \E \left [   \frac{1}{\eta_{1,d}^2}  \right ] \left(    \left(\rho_{*,l, d} \E \left[\tilde{\bW}_{l,d} \right]\E \left[\tilde{\ba}_{*,l-1}\right] - \E\left [a_{*,l,d} \right]\right)^2  +  \E\left [a_{*,l,d}^2 \right] - \E \left[a_{*,l,d} \right]^2 \right)\\
&- \frac{1}{2}  \sum_{l=1}^L \sum_{d=1}^{D_l}  \E \left [   \frac{1}{\eta_{l,d}^2}  \right ] \left(  \rho_{*,l, d}  \Tr \left ( \left(\mathbf{B}_{l,d} + \mathbf{m}_{l,d}\mathbf{m}_{l,d}^T \right) \E \left[\tilde{\ba}_{*,l-1} \tilde{\ba}_{*,l-1}^T \right]  \right)- \rho_{*,l, d}^2\Tr \left( \mathbf{m}_{l,d}\mathbf{m}_{l,d}^T \E \left[ \tilde{\ba}_{*,l-1} \right]  \E \left[ \tilde{\ba}_{*,l-1}^T \right]\right) \right) \\
& -  \sum_{l=1}^L \sum_{d=1}^{D_l}  \E \left [   \frac{1}{\eta_{l,d}^2}  \right ] \left(    \rho_{*,l, d}\E \left[\tilde{\bW}_{l,d} \right]\left( \E \left[ \ba_{*,l,d} \right] \E \left[\tilde{\ba}_{*,l-1}\right] -  \E \left[ a_{*,l,d}   \tilde{\ba}_{*,l-1} \right ]\right) \right) + \frac{1}{2}  \E \left [ \log  \eta_{l,d}^{2}   \right ]  \\
&   +  \frac{1}{2}\sum_{l=1}^{L} \log(|  \bS_{*,l} |)    + \sum_{l=1}^{L} \sum_{d=1}^{D_l} \frac{1}{T}  \left(\rho_{*,l,d} - \frac{1}{2} \right) \E \left [  \widetilde{\bW}_{l,d} \right ] \E \left [ \widetilde{\ba}_{*,l-1} \right]  + \frac{1}{2T^2}  \left(  \E \left [ \omega_{*,l,d} \right ] \right)   \Tr \left(  \E \left [  \tilde{\bW}_{l,d}^T  \tilde{\bW}_{l,d} \right ]   \E \left [  \tilde{\ba}_{*,l-1}\tilde{\ba}_{*,l-1}^T  \right ] \right)   \\
&-  \sum_{l=1}^{L} \sum_{d=1}^{D_l} \left ( \rho_{*,l,d}  \log  \rho_{*,l,d} + (1- \rho_{*,l,d} )\log (1- \rho_{*,l,d})  - \frac{A_{*,l,d}^2}{2} \E \left [  \omega_{*,l,d}  \right ]  + \log(\cosh(\frac{A_{*,l,d}}{2}))\right ) + \text{const}.
\end{align*}
% where 
% \begin{align*}
% % \text{const} & =  \left(\frac{1}{2}  -  \log(2) \right) \sum_{l=1}^{L} D_l, \\
%    &  \E \left [  \omega_{*,l,d}  \right ]  =  \frac{\exp\left( A_{*,l,d} \right) -1}{  2A_{*,l,d}\left(\exp\left( A_{*,l,d}\right) + 1 \right)}, \\
%     &  \E \left [ \bW_{l,d}^T \bW_{l,d} \right ] = \left( \mathbf{m}_{l,d}^W\right) ^T\mathbf{m}_{l,d}^W+  \mathbf{B}_{l,d}^W, \quad \E \left [ b_{l,d}^2 \right ] =  \left( \mathbf{m}_{l,d}^b\right)^2  + \mathbf{B}_{l,d}^b, \\
%    & \E \left [ \bW_{l, d} \right ] = \mathbf{m}_{l,d}^W, \quad
%       \E \left [ b_{l,d} \right ] = \mathbf{m}_{l,d}^b.
% \end{align*}

 \blu{The complexity of obtaining variational update for $\ba$ is then $\mathcal{O}(NLD^3)$. }
\blu{\section{Supplementary details for the stochastic variational inference algorithm} \label{ap:svisup}
Here we provide additional details on the SVI developed for the VBNN in \cref{sec:svivbnn}. During one iteration $t$ of the algorithm, one proceeds as follows: 
\begin{enumerate}
\item Sample indices $S_t$ uniformly, without replacement. 
\item For $t=1$ initialize as in \cref{sec:initschemes} (similarly to CAVI) but where the input of \cref{alg:laplace} is taken to be $\bx_{n}$ for $n \in S_t$.
For $t>1$ only initialize local parameters of $\ba$ and $\gamma$ by setting $\bz_{n, 0} = \bx_n$ and iterating for $l=1, \ldots, L$ and $n \in S_t$ through 
\begin{align*}
   & \rho_{n,l,d} = \sigma \left ( \frac{(m_{l,d}^b)^{(t)} + (\bm{m}_{l,d}^W)^{(t)} \bz_{n,l-1}}{T}\right) \quad d = 1, \ldots, D_{l},\\
  &  \bM_{n,l} = (\bm{m}_{l}^W)^{(t)}  \odot  \bm{\rho}_{n,l}\bm{1}^T_{D_{l}}, \text{ where by } \bm{1} \text{ we denote a vector of ones,} \\
  & \bt_{n,l} =( m_{l}^b)^{(t)} \odot \bm{\rho}_{n,l}, \\
  &\bz_{n,l} = \bM_{n,l} \bz_{n,l-1} + \bt_{n,l}.
\end{align*}
\item Set $\ell_t = (t + 1)^{-k}, \; k \in (0.5, 1].$
 \item Update global shrinkage parameters of $\btau$ as in CAVI, for $l=1,\ldots,L+1$,
\begin{align*}
 \nu_{\glob,l}^{(t)}&= \nu_{\glob}-\frac{D_lD_{l-1}}{2}, \\
\delta_{\glob,l}^{(t)}  &= \sqrt{\delta_{\glob}^2 + \sum_{d}^{D_l} \sum_{d'}^{D_{l-1}} \E \left [ \frac{1}{\psi_{l,d,d'}} \right] \E \left[ W_{l,d,d'}^2  \right]}, 
\end{align*}
where $ \nu_{\glob,l}$ is only updated in the first iteration of the algorithm. 
\item   Update local shrinkage parameters of $\bpsi$ as in CAVI, for $l=1, \dots, L+1, \; d=1,\ldots, D_l, \; d'=1,\ldots, D_{l-1}$,
\begin{align*}
  \nu_{\loc,l,d,d'}^{(t)} &= \nu_{\loc,l}-\frac{1}{2}, \\  
    \delta_{\loc,l,d,d'}^{(t)} &=  \sqrt{ \E \left [ \frac{1}{\tau_l} \right ] \E \left[ W_{l,d,d'}^2  \right]  + \delta_{\loc,l}^2}, 
\end{align*}    
where $ \nu_{\loc,l,d,d'}$ is only updated once. 
% \AS{Same, the natural parameter is $-\delta^{-2}/2$, : 
% \begin{align*}
% &  \hat{\delta}_{\loc,l,d,d'}^2  \E \left [ \frac{1}{\tau_l} \right ] \E \left[ W_{l,d,d'}^2  \right]  + \delta_{\loc,l}^2\\
%    & \delta_{\loc,l,d,d'}^{(t)} = \left (\frac{(1-\ell_t)}{ (\delta_{\loc,l,d,d'}^{(t)})^2} + \frac{\ell_t}{\hat{\delta}_{\loc,l,d,d'}^2}  \right)^2
% \end{align*}
% }
\item Find optimal variational parameters of local variables $\bomega$, $\bgamma$, $\ba$, namely, update $A^{(t)},\bm{S}^{(t)}, \bm{t}^{(t)}, \bm{M}^{(t)},  \bm{\rho}^{(t)}$ in a coordinate ascent algorithm and monitor the local ELBO for convergence: 
\begin{itemize}
    \item For $n\in S,\; l=1,\ldots, L, d=1, \ldots, D_l$, update 
    \begin{align*}
    \mathrm{a}_{n,l,d} &= \frac{1}{T} \sqrt{\left ( \Tr \left ( \E \left [ \widetilde{\bW}_{l,d}^T \widetilde{\bW}_{l,d} \right ] \E \left [ \widetilde{\ba}_{n,l-1} \widetilde{\ba}_{n,l-1}^T\right ] \right ) \right )}.
\end{align*}
\item Starting from the final layer $l = L$, update for  $n\in S$
\begin{align*}
    & \bm{S}^{-1}_{n,L} = \hat{\bm{\Sigma}}_{L}^{-1} + \sum_{d=1}^{D_{L+1}}\E \left [  \frac{1}{\eta_{L+1,d}^2}\right ]  \E \left [ \bW_{L+1,d}^T \bW_{L+1,d} \right]  \quad \text{(same for all n),} \\
    & \bm{t}_{n,L} = \bm{S}_{L} \left ( \hat{\bm{\Sigma}}_{L}^{-1}  \E \left[ \bgamma_{n,L}\right ] \odot \E \left [ \bb_{L}\right ] \right.\\
    &\left. + \sum_{d=1}^{D_{l+1}} \E \left [\frac{1}{\eta_{L+1,d}^2} \right ] 
   \left ( -  \E \left [ \bW_{L+1,d}^T b_{L+1,d} \right] +  \E \left [ \bW_{L+1,d}^T \right] y_{n,d} \right )  \right ), \\
   & \bm{M}_{n,L} = \bm{S}_{L} \hat{\bm{\Sigma}}_{L}^{-1} \E \left[ \bgamma_{n,L}\right ]  \bm{1}^T_{D_{L-1}} \odot \E \left [  \bW_{L}\right ], \\
   &\hat{\bm{\Sigma}}_{L}^{-1}=\diag \left ( \E \left[\eta_{L,1}^{-2} \right ], \dots, \E \left[\eta_{L,D_L}^{-2}\right ]\right ). 
\end{align*}
Then, in reverse order, for $l = L-1, \ldots, 1$ and for  $n\in S$ update 
\begin{align*}
   \bm{S}^{-1}_{n, l} &  = \hat{\bm{\Sigma}}_l^{-1} - \bm{M}_{n, l+1}^T \bm{S}_{n,l+1}^{-1}\bm{M}_{n, l+1} \\
   &+ \sum_{d=1}^{D_{l+1}} \left ( \E \left [\frac{1}{\eta_{l+1,d}^2} \right ] \E \left[ \gamma_{n,l+1,d}\right ] + \frac{1}{T^2}  \E \left [ \omega_{n,l+1,d}\right ] \right )\E \left [ \bW_{l+1,d}^T \bW_{l+1,d} \right ], \\
   \bm{t}_{n, l}  & = \bm{S}_{n,l} \left ( \bm{M}_{n, l+1}^T \bm{S}_{n, l+1}^{-1} \bm{t}_{n,l+1} + \hat{\bm{\Sigma}}_l^{-1}  \E \left[ \bgamma_{n,l}\right ] \odot  \E \left[ \bb_{l}\right ] \right. \\
   & \left. + \frac{1}{T} \sum_{d=1}^{D_{l+1}}\E \left [ \bW_{l+1,d}^T \right ]  \left ( \E \left [ \gamma_{n,l+1,d} \right ]  - \frac{1}{2}\right )  \right.  \\
    & \left. - \sum_{d=1}^{D_{l+1}} \left ( \E \left [\frac{1}{\eta_{l+1,d}^2} \right ] \E \left[ \gamma_{n,l+1,d}\right ] +  \frac{1}{T^2} \E \left [ \omega_{n,l+1,d}\right ]\right)
\E \left [ \bW_{l+1,d} b_{l+1,d}\right ] \right ),  \\
 \bm{M}_{n,l} &= \bm{S}_{n,l} \hat{\bm{\Sigma}}_l^{-1} \E \left[ \bgamma_{n,l}\right ]  \bm{1}^T_{D_{l-1}} \odot \E \left [  \bW_{l}\right ], \\
 \hat{\bm{\Sigma}}_{l}^{-1} &=\diag \left ( \E \left[ \eta_{l,1}^{-2} \right ], \dots, \E \left[\eta_{l,D_l}^{-2}\right ]\right ). 
\end{align*}

\item   For $n\in S, \; l=1,\ldots, L, \; d=1, \ldots, D_l$, update   \begin{align*}
  \rho_{n,l,d} =    \sigma & \left ( -\frac {\E \left [ \eta_{l,d}^{-2} \right ] }{2}   \Tr \left (\E \left[\widetilde{\bW}_{l,d}^{T} \widetilde{\bW}_{l,d}\right ] \E \left [\widetilde{\ba}_{n,l-1} \widetilde{\ba}_{n, l-1}^T \right ] \right ) \right. \\
  & + \left. \E \left [ \eta_{l,d}^{-2} \right ] \E \left [ \widetilde{\bW}_{l,d} \right ] \E \left [\widetilde{\ba}_{n,l-1} \mathrm{a}_{n,l,d} \right ] + \frac{1}{T} \E \left [  \widetilde{\bW}_{l,d}  \right ] \E \left [\widetilde{\ba}_{n,l-1}\right ] \right).
\end{align*}
The local ELBO is given by 

\footnotesize
\begin{align*}
&\E \left [  \log p (\by, \ba, \bgamma,\bomega | \bW, \bb, \bSigma)\right ]   - \E \left [\log  q(\ba) \right ] - \E \left [\log q(\bgamma) \right ] - \E \left [\log q(\bomega)\right ]  \\
&= \frac{N}{|S|} \sum_{n\in S} \sum_{d=1}^{D_{L+1}} \E \left [ \log  \Norm\left( y_{n,d} | \bz_{n,L+1,d}, \bSigma_{L+1,d}\right) \right ] \\
&+\frac{N}{|S|} \sum_{n\in S}  \sum_{l=1}^L \sum_{d=1}^{D_l} \E \left [ \log \Norm \left( \mathrm{a}_{n,l,d} | \bgamma_{n,d} \odot \bz_{n,l,d} , \bSigma_{l,d} \right) \right ]   \\
 &+ \frac{N}{|S|} \sum_{n\in S}  \sum_{l=1}^L \sum_{d=1}^{D_l} \E \left [ \log \left ( \exp \left (\frac{\kappa_{n,l,d} z_{n,l,d}}{T}\right ) \exp \left (-\frac{\omega_{n,l, d}z_{n,l,d}^2}{2T^2} \right ) \PG(\omega_{n,l,d} | 1,0) \right )  \right ]  \\
 &-  \frac{N}{|S|} \sum_{n\in S} \sum_{l=1}^L \E \left [ \log \Norm \left (\ba_{n,l} |  \bt_{n,l} + \bM_{n,l} \ba_{n,l-1}, \bS_{n,l} \right ) \right ]   \\
  &- \frac{N}{|S|} \sum_{n\in S} \sum_{l=1}^L \sum_{d=1}^{D_l} \E \left [ \log \Bern\left (\gamma_{n,l,d} | \rho_{n,l,d} \right ) \right ] + \E \left [ \log \PG(  \omega_{n,l,d} | 1, \mathrm{a}_{n,l,d}) \right ]  \\
&= \frac{N}{|S|} \sum_{n\in S} \sum_{d=1}^{D_{L+1}} \E \left [ \log  (\eta_{L+1,d}^2)^{-1/2} \exp \left(-\frac {1}{2\eta_{L+1,d}^{2} } \left ( y_{n,d} - \bW_{L+1,d} \ba_{n, L} - b_{L+1,d} \right )^{2}  \right)  \right] \\
&+  \frac{N}{|S|} \sum_{n\in S}  \sum_{l=1}^L \sum_{d=1}^{D_l} \E \left [ \log  (\eta_{l,d}^2)^{-1/2} \exp \left(-\frac {1}{2\eta_{l,d}^{2} } \left ( \mathrm{a}_{n,l,d} - \gamma_{n,l,d} \odot \left ( \bW_{l,d} \ba_{n,l-1} + b_{l,d} \right )  \right )^{2} \right)  \right ] \\
& - N \sum_{l=1}^{L} D_l \log(2)  - \frac{N D_{L+1}}{2} \log(2\pi) \\
&  +  \frac{1}{T} \frac{N}{|S|} \sum_{n\in S}  \sum_{l=1}^{L}  \sum_{d=1}^{D_l}  \E \left [ \left(\gamma_{n,d} - \frac{1}{2} \right)  \left ( \bW_{l,d} \ba_{n,l-1} + b_{l,d} \right )\right ] \\
& -\frac{1}{2T^2} \frac{N}{|S|} \sum_{n\in S}  \sum_{l=1}^{L}\sum_{d=1}^{D_l} \E \left [ \omega_{n,l,d} \left ( \bW_{l,d} \ba_{n,l-1} + b_{l,d} \right )^2 \right ] \\
&- \frac{N}{|S|} \sum_{n\in S}  \sum_{l=1}^{L} \E \left [ \log  |\bS_{n,l}|^{-\frac{1}{2}}  \right ] \\
&- \frac{N}{|S|} \sum_{n\in S}  \sum_{l=1}^{L} \E \left [ -\frac{1}{2} \left ( \ba_{n,l} -  \bt_{n,l}- \bM_{n,l} \ba_{n,l-1}  \right)^T \bS_{n,l}^{-1}\left ( \ba_{n,l} -  \bt_{n,l}- \bM_{n,l} \ba_{n,l-1}  \right)   \right ] \\
& - \frac{N}{|S|} \sum_{n\in S}  \sum_{l=1}^{L} \sum_{d=1}^{D_l} \left ( \rho_{n,l,d}  \log  \rho_{n,l,d} + (1- \rho_{n,l,d} )\log (1- \rho_{n,l,d}) \right )  \\
& + \frac{N}{|S|} \sum_{n\in S} \sum_{l=1}^{L} \sum_{d=1}^{D_l} \E \left [ \log \frac{\PG(\omega_{n,l,d} | 1,0)}{\PG(  \omega_{n,l,d}  | 1, A_{n,d})}  \right ]  \\
=& - \frac{1}{2} \frac{N}{|S|} \sum_{n\in S} \sum_{d=1}^{D_{L+1}} \E \left [\frac {1}{\eta_{L+1,d}^{2}} \right] \left ( y_{n,d}^2 - 2y_{n,d}  \E \left [ \widetilde{\bW}_{L+1,d} \right]\E \left [ \widetilde{\ba}_{n, L} \right]  \right )  \\
& - \frac{1}{2} \frac{N}{|S|} \sum_{n\in S} \sum_{d=1}^{D_{L+1}} \E \left [\frac {1}{\eta_{L+1,d}^{2}} \right] \left (  \Tr \left (\E \left [  \widetilde{\bW}_{L+1,d}^T\widetilde{\bW}_{L+1,d}\right] \E \left [ \widetilde{\ba}_{n, L}\widetilde{\ba}_{n, L}^T  \right] \right )  \right )  \\
& -  \frac{1}{2} \frac{N}{|S|} \sum_{n\in S}  \sum_{l=1}^L \sum_{d=1}^{D_l} \E \left [ \frac {1}{\eta_{l,d}^{2} } \right ] \left( \E \left [  \mathrm{a}_{n,l,d}^2 \right ]  - 2\E \left [ \gamma_{n,l,d} \right ] \E \left [  \widetilde{\bW}_{l,d}  \right ] \E \left [ \widetilde{\ba}_{n,l-1} \mathrm{a}_{n,l,d}  \right ]  \right)   \\
& -  \frac{1}{2} \frac{N}{|S|} \sum_{n\in S}  \sum_{l=1}^L \sum_{d=1}^{D_l} \E \left [ \frac {1}{\eta_{l,d}^{2} } \right ]  \E \left [ \gamma_{n,l,d}^2 \right ] \Tr \left(  \E \left [  \widetilde{\bW}_{l,d}^T  \widetilde{\bW}_{l,d} \right ]   \E \left [  \widetilde{\ba}_{n,l-1}\widetilde{\ba}_{n,l-1}^T  \right ] \right)   \\
&-\frac{N}{2} \sum_{d=1}^{D_{L+1}} \E \left [ \log  \eta_{L+1,d}^{2}   \right]   - \frac{N}{2} \sum_{l=1}^{L} \sum_{d=1}^{D_l} \E \left [ \log  \eta_{l,d}^{2}   \right ] + \frac{1}{2} \sum_{n=1}^{N}\sum_{l=1}^{L} \log(|  \bS_{n,l} |)  \\
&+  \frac{N}{|S|} \sum_{n\in S} \sum_{l=1}^{L} \sum_{d=1}^{D_l}\frac{1}{T} \left(\rho_{n,l,d} - \frac{1}{2} \right) \E \left [  \widetilde{\bW}_{l,d} \right ] \E \left [ \widetilde{\ba}_{n,l-1} \right] \\
& -\frac{1}{2T^2} \frac{N}{|S|} \sum_{n\in S} \sum_{l=1}^{L} \sum_{d=1}^{D_l} \E \left [ \omega_{n,l,d} \right ] \left ( \Tr \left( \E \left [\widetilde{\bW}_{l,d}^T\widetilde{\bW}_{l,d} \right ] \E \left [ \widetilde{\ba}_{n,l-1}\widetilde{\ba}_{n,l-1}^T\right]\right) \right )  \\
&- \frac{N}{|S|} \sum_{n\in S} \sum_{l=1}^{L} \sum_{d=1}^{D_l}  \rho_{n,l,d}  \log  \rho_{n,l,d} + (1- \rho_{n,l,d} )\log (1- \rho_{n,l,d}) \\
& +\frac{N}{|S|} \sum_{n\in S} \sum_{l=1}^{L} \sum_{d=1}^{D_l} \frac{\mathrm{a}_{n,l,d}^2}{2} \E \left [  \omega_{n,l,d}  \right ] - \log(\cosh(\frac{\mathrm{a}_{n,l,d}}{2}))  + C_a, 
\end{align*}\normalsize

where the normalizing constant is 

{\footnotesize\begin{align*}
    C_a =  - \frac{N D_{L+1}}{2} \log(2\pi)   - N \sum_{l=1}^{L} D_l \log(2).
\end{align*}\normalsize}

\end{itemize}

\item Find global variational parameters for which we recall the vector of natural parameters for $\left(\bW,\bb\right)$ is $(\bm{B}^{-1} \bm{m}^T, -\bm{B}^{-1}/2)$, and for $\bmeta^2$ that is $(-\bm{\alpha+1}, -\bm{\beta}^{-1})$. We are only updating the parameter $\bm{\alpha}$ in the first iteration of the algorithm as 
\begin{align*}
    & \alpha_{l,d}= \alpha^h_0+ \frac{N}{2}, \text{ for } l=1, \ldots, L, \; d=1, \ldots, D_l, \\
     &   \alpha_{L+1,d} = \alpha_0  + \frac{N}{2} \text{ for } d=1, \ldots, D_{L+1}.
\end{align*}
We then find $\bm{\beta}$ via the intermediate variable $\hat{\beta}_{l,d}$:
    \item For  $l=1, \ldots, L, \; d=1, \ldots, D_l$ set
 \begin{align*}
    & \hat{\beta}_{l,d} = \beta^h_0  + \frac{1}{2}\frac{N}{|S|}\sum_{n\in S}  \left(  \E \left [\mathrm{a}_{n,l,d} \right ] - \E \left [\gamma_{n,l,d} \right ]   \E \left [\widetilde{\bW}_{l,d} \right ]   \E \left [\widetilde{\ba}_{n,l-1} \right ]\right )^2  \\
&  +  \frac{1}{2}\frac{N}{|S|}\sum_{n\in S}\E \left [a^2_{n,l,d} \right ] -  \E \left [\mathrm{a}_{n,l,d} \right ]^2  + \E \left [\gamma_{n,l,d} \right ] \Tr \left ( \E \left [ \widetilde{\bW}_{l,d}^T \widetilde{\bW}_{l,d} \right ]  \E \left [\widetilde{\ba}_{n,l-1} \widetilde{\ba}_{n,l-1}^T \right ] \right ) \\
& -\frac{1}{2}\frac{N}{|S|}\sum_{n\in S} \E \left [\gamma_{n,l,d} \right ]^2  \Tr \left ( \E \left [\widetilde{\bW}_{l,d}^T \right ] \E \left [\widetilde{\bW}_{l,d} \right ]  \E \left [\widetilde{\ba}_{n,l-1} \right ] \E \left [ \widetilde{\ba}_{n,l-1}^T \right ]  \right ).
\end{align*}
And for the final layer $l = L+1$ and $d=1, \ldots, D_{L+1}$
 \begin{align*}
       & \hat{\beta}_{L+1,d}
  = \beta_0 +  \frac{1}{2}\frac{N}{|S|}\sum_{n\in S}  \left( y_{n,d} -  \E \left [  \widetilde{\bW}_{L+1,d} \right ]  \E [ \widetilde{\ba}_{n,L}] \right)^2    \\
    & +  \frac{1}{2}\frac{N}{|S|}\sum_{n\in S}  \Tr \left ( \E \left [  \widetilde{\bW}_{L+1,d}^T \widetilde{\bW}_{L+1,d} \right ] \E \left [ \widetilde{\ba}_{n,L} \widetilde{\ba}_{n,L}^T\right] \right)  \\
   &- \frac{1}{2} \frac{N}{|S|}\sum_{n\in S} \Tr \left (\E [  \widetilde{\bW}_{L+1,d}]^T \E[\widetilde{\bW}_{L+1,d} ] \E [ \widetilde{\ba}_{n,L}] \E[\widetilde{\ba}_{n,L}^T]\right ).
\end{align*} 
The update for $l = 1, \ldots, L+1, \; d=1, \ldots, D_l$ is given by 
    \begin{align*}
    \beta_{l,d}^{(t)} = \left( (1- \ell_{t})\times(\beta_{l,d}^{(t-1)})^{-1} + \ell_{t} \times \hat{\beta}_{l,d}^{-1} \right)^{-1}.
\end{align*}
 Similarly, the variational parameters $\bm{B}, \bm{m}$ of global variables $( \bb, \bW)$ are obtained as a reparametrized linear combination of previous and intermediate updates. 
Specifically, for  $l=1, \ldots, L, \; d=1, \ldots, D_l$ set 
 \begin{align*}
   & \hat{\bm{B}}_{l,d}^{-1} = \bm{D}^{-1}_{l,d}+\frac{N}{|S|}\sum_{n\in S}   \left ( \frac{1}{T^2} \E \left [\omega_{n,l,d}\right ] + \E \left [ \eta_{l,d}^{-2}\right ] \E \left [\gamma_{n,l,d}\right ]  \right) \E \left [\widetilde{\ba}_{n, l-1} \widetilde{\ba}_{n, l-1}^T \right ] ,  \\
&    \hat{\bm{B}}_{l,d}^{-1}\hat{\bm{m}}_{l,d}^T  =  \frac{N}{|S|}\sum_{n\in S} \E \left [  \eta_{l,d}^{-2} \right ] \E \left [   \gamma_{n,l,d}\right ] \E \left [ \mathrm{a}_{n,l,d} \widetilde{\ba}_{n,l-1}\right ] + \frac{1}{T} \E \left [\widetilde{\ba}_{n, l-1} \right ] \left ( \E \left [\gamma_{n,l,d} \right ]  - \frac{ 1}{2}\right).\\
    \end{align*}
 For the final layer $l = L+1$ and $d=1, \ldots, D_{L+1}$ set 
   \begin{align*}
    &  \hat{\bm{B}}^{-1}_{L+1,d}=\bD^{-1}_{L+1,d} +\E \left [\eta_{L+1,d}^{-2} \right ] \frac{N}{|S|}\sum_{n\in S} \E \left [\widetilde{\ba}_{n,L+1} \widetilde{\ba}_{n,L+1}^T \right], \\
     & \hat{\bm{B}}_{L+1,d}^{-1} \hat{\bm{m}}_{L+1,d}^T  =\E \left [\eta_{L+1,d}^{-2} \right ] \left ( \frac{N}{|S|}\sum_{n\in S} y_n \E \left [\widetilde{\ba}_{n,L+1} \right]  \right ),
\end{align*}
where for $l=1, \ldots, L+1$ and $d=1, \ldots, D_{l}$,
\begin{align*}
\bD_{l,d}^{-1} &= \diag \left ( s_0^{-2}, \E \left [\tau_l^{-1}\right ] \E \left [ \psi_{l,d,1}^{-1} \right ], \ldots, \E \left [\tau_l^{-1}  \right ] \E \left [ \psi_{l,d,D_{l-1}}^{-1} \right ]\right ).
\end{align*}
Then the updates  $l=1, \ldots, L+1$ and $d=1, \ldots, D_{l}$, are given by
\begin{align*}
&\bm{B}_{l,d}^{(t)}  = \left( (1- \ell_{t})\times (\bm{B}_{l,d}^{(t-1)})^{-1} + \ell_{t} \times \hat{\bm{B}}_{l,d}^{-1} \right)^{-1} , \\
& \bm{m}_{l,d}^{(t)}  =( (1- \ell_{t}) \bm{B}_{l,d}^{(t)}(\bm{B}_{l,d}^{(t-1)})^{-1}(\bm{m}_{l,d}^{(t-1)})^T + \ell_{t}  \bm{B}_{l,d}^{(t)}\hat{\bm{B}}_{l,d}^{-1}\hat{\bm{m}}_{l,d}^T)^{T}.
\end{align*}
\end{enumerate}
We monitor the noisy estimate of the ELBO which is computed as in \cref{appendix:ELBO} but with sums over $n =1, \ldots, N$ replaced with the scaled sums over $n \in S$. }

\section{Experiments} \label{sec:experimentsap}

\subsection{Initialization schemes} \label{sec:initschemes}
Initialization plays an important role in the ability of Bayesian inference algorithms to effectively approximate the posterior. This is especially true in variational schemes for complex posteriors (such as for BNNs), which are only guaranteed to converge to local optimum. We design two possible variations of random yet effective initialization schemes. To simplify the exposition, we describe the procedure in the case of Inverse Gamma shrinkage priors, for which $\lambda = 0$ and the selection of the scale parameters $\delta$ determines the level of shrinkage. Note that during the training step, we employ the expectation-maximization algorithm to set an optimal $\delta_{\glob}$, whilst the value of $\delta_{loc,l}$ remains fixed. To encourage more shrinkage for larger depth, we assume  $\delta_{\glob} \propto 1/\sqrt{L}$, and to encourage shrinkage for larger width set $\delta_{\loc, l} \propto 1/\sqrt{D_l}$. Given  specified values of  $\nu_{\loc}, \nu_{\glob}, \delta_{\loc},  \delta_{\glob}, \alpha^h_0,  \alpha_{0},  \beta^{h}_{0},  \beta_{0}$, we first re-scale the shrinkage parameters to  scale appropriately $$\delta_{\glob} =\frac{\delta_{\glob}}{\sqrt{L}}, \delta_{\loc, l} =\frac{\delta_{\loc}}{\sqrt{D_{l-1}}},\nu_{\loc, l} =  \nu_{\loc},  $$ and the initialization steps are:
\begin{enumerate}
    \item Covariance for biases and weights: $\bB_{l,d} = 0.01 \mathbf{I}_{D_{l-1} + 1}$ for $l = 1, \ldots, L+1, d = 1, \ldots D_l$.
    \item Covariance for stochastic activation: $\bS_{n,l} = 0.01  \mathbf{I}_{D_{l}}$ for $n = 1, \ldots, N, \; l = 1, \ldots, L.$
    \item Variational parameters for $\bmeta$: Set  $\alpha_{L+1,d} = \alpha_{0}, \alpha_{l,d} = \alpha_{0}^h $ and  $\beta_{L+1,d} = \beta_{0}, \beta_{l,d} = \beta_{0}^h.$ 
     \item Variational parameters for $\btau, \bpsi$:   
    \begin{align*}
    \nu_{\loc, l,d,d'} & =  \nu_{\loc, l},  \;  \nu_{\glob, l} =  \nu_{\glob}, \\
        \delta_{\glob, l} &\sim \sqrt{2(\nu_{\glob, l} -1)\IG\left (\nu_{\glob, l},\delta_{\glob}\right )}, \\
          \delta_{\loc, l,d,d'}& \sim \sqrt{2(\nu_{\loc, l,d,d'} -1)\IG \left(\nu_{\loc, l,d,d'}, \delta_{\loc, l} \right)}.
    \end{align*}

    \item Use \cref{alg:laplace} to initialize the variational means of the weights and biases for all intermediate layers, and the variational means of the stochastic activations and the variational parameters of the binary activations. 
%     $\mathbf{m}_{l,d} \in \R^{D_{l-1}+1}$ for all $l = 1, \ldots, L$, specifically: 
% $\mathbf{m}_{l,d} = \left (m^b_{l,d}, \left(\mathbf{m}^W_{l,d}\right) ^T  \right )^T$
    \item Variational mean of the weights and biases for the last layer $\mathbf{m}_{L+1}$ is obtained as a solution of fitting $D_{y}$ ridge regressions with inputs $\mathbf{z}_{L} $ and outputs  $\by_{d}$.

\end{enumerate}

 \begin{algorithm}
\caption{Initialization} \label{alg:laplace}
\begin{algorithmic}
\REQUIRE   Training inputs $\bx_{n}$, $n=1,\ldots, N$; choice of mode \textit{laplace} or \textit{spike-slab}
\STATE{$\bz_{n,0} =\bx_{n} $} 
\FOR{$l = 1 \ldots L,$}
\STATE{set $\Delta = 0.05*(\max(\mathbf{z}_{n,l-1}) -  \min(\mathbf{z}_{n,l-1}))$}
\FOR{$d = 1 \ldots D_{l}$}
\IF{\textit{laplace}}
\STATE{\begin{align*}
   & m_{l,d, d'}^W \sim \text{Laplace}\left(0, \sqrt{\frac{2}{D_{l-1}}}\right), 
\end{align*}}
\ENDIF.
\IF{\textit{spike-slab}}
\STATE{
\begin{align*}
   % & m_{l,d, d'}^W \sim \Norm \left(0, 1\right) \times \Bern \left (\frac{1}{1 +D_{l-1}} \right)  \\
     & m_{l,d, d'}^W \sim \pi \Norm \left(0, \frac{2}{\sqrt{D_{l-1}}}\right) + (1-\pi) \delta_0, \text{ where } \pi = \frac{1}{1+\sqrt{D_{l-1}}},
\end{align*}}
\ENDIF.
\STATE{\begin{align*}
   % &s_{d'}^S \sim \text{Cat}(\mathcal{S}, \mathcal{P}), \; \hat{\mathcal{S}} = \hat{\mathcal{S}} \cup \{s_{d'}^S \} \SW{\text{Maybe we don't need this step, instead the following}} \\
    &\mathbf{s} = (s_{1}, \ldots, s_{D_{l-1}}), \text{ where } s_{d'} \sim \text{Unif}([\min(z_{n,l-1,d'}) - \Delta_{d'}, \max(z_{n,l-1,d'})  + \Delta_{d'}]),  \\
  &  m_{l,d}^b =  - \mathbf{m}_{l,d}^W \mathbf{s}, \; \mathbf{m}_{l,d} = \left( m_{l,d}^b, \mathbf{m}_{l,d}^W\right), 
  % &  \mathcal{S} = \mathcal{S} \setminus s_{d'}^S, \; \mathcal{P} = \{ \min(\{\text{dist}(s - \hat{s}) \; |\; \hat{s} \in \hat{\mathcal{S}} \}) \; | \; s \in \mathcal{S}\} \SW{\text{Maybe we don't need this step, instead the following}}
\end{align*}}
\ENDFOR.
% \SW{Check these equations, missing variables...}
\STATE{\begin{align*}
   & \rho_{n,l,d} = \sigma \left ( \frac{m_{l,d}^b + \mathbf{m}_{l,d}^W \bz_{n,l-1}}{T}\right) \quad d = 1, \ldots, D_{l},\\
  &  \bM_{n,l} = \mathbf{m}_{l}^W \odot  \mathbf{\rho}_{n,l}\mathbf{1}^T_{D_{l}}, \text{ where by } \mathbf{1} \text{ we denote a vector of ones,} \\
  & \bt_{n,l} = m_{l}^b \odot \mathbf{\rho}_{n,l}, \\
  &\bz_{n,l} = \bM_{n,l} \bz_{n,l-1} + \bt_{n,l},
\end{align*}}
\ENDFOR.
\ENSURE $ \bM_{n,l},  \bt_{n,l}, \mathbf{m}_{l,d}$ for $l = 1, \ldots, L, d = 1, \ldots, D_{l}$ and $\mathbf{z}_{L}$.
\end{algorithmic}
\end{algorithm}

% In the variational distribution of weights and biases $(\bW, \bb)$, the mean $\mathbf{m}_{l,d} \in \R^{D_{l-1}+1}$ is a row vector and the covariance $\mathbf{B}_{l,d}$ is a $\R^{D_{l-1}+1} \times \R^{D_{l-1}+1}$ matrix for all $l = 1, \ldots, L+1$, specifically: 
% \begin{align*}
%     & \mathbf{m}_{l,d} = \left (m^b_{l,d}, \left(\mathbf{m}^W_{l,d}\right) ^T  \right )^T, \\
%     & \mathbf{B}_{l,d} = \begin{pmatrix} B_{l,d}^{b} &  \left(\mathbf{B}_{l,d}^{bW}\right)^T \\ \mathbf{B}_{l,d}^{bW}& \mathbf{B}_{l,d}^{W} \end{pmatrix}.
% \end{align*}

\subsection{Implementation details}

When comparing the performance of our method to already existing ones we implement the following model in \cmsspy{Numpyro}: 
\begin{align*}
    &\by \sim \Norm \left(\bW_{L+1} \text{ReLU}(\bz_{L})+ \bb_L, \bSigma \right),  \text{ where } \quad \bSigma \sim \IG(2, \sigma_y) \mathbf{I}_{D_y}, \\
\bz_{l} &=  \bW_l  \text{ReLU}(\bz_{l-1}) + \bb_l, \; W_{l, d, d'} \sim \Norm \left(0, \frac{\sigma_W^2 \gamma }{\sqrt{D_{l-1}}}\right), \quad b_{l, d} \sim \Norm(0, \sigma_b^2\gamma), 
\end{align*}
  where $\bz_{n,0} =\bx_{n}, \;  \gamma \sim \IG(2, 1) \text{ and } l=1,\ldots, L, \; d =1, \ldots D_{l}, \; d' = 1, \ldots, D_{l-1}. $ The choice of $\sigma_y, \sigma_W $ and $ \sigma_b$ is made in accordance with $\alpha_0, s_0$ and 
$\delta_{\loc, l}$, respectively. For experiments with \blu{mfVI} we use \cmsspy{Adam} optimizer with learning rate set to 0.001 and maximum number of iterations varying from 5000 to 20000 depending on the dataset and depth of the network. Additionally, we consider the Bayes by Backprop model of \citep{blundell2015weight} and adapt its \cmsspy{Pytorch} implementation from the publicly available repository \citep{BNNBBBgithub}. For all experiments with BBB we set the learning rate to 0.01 and maximum number of epochs varies from 500 to 1000. 

In all examples, we normalize the input but do not re-scale the output. 
Suppose that the data on which we evaluate the predictive performance consists of $N$ points and the true target is $\by^{*}$, then recorded evaluation metrics are RMSE, NLL and EC and are computed as follows: 
\begin{align*}
       \text{RMSE} & = \sqrt{\frac{1}{N} \sum_{n}^N \left [(y^{*}_{n} - \E [y^{o}_{n}])^2\right ]},\\
        \text{NLL} & = \frac{1}{N} \sum_{n}^N \log \Norm (y^{*}_{n} \; | \; \E[y^{o}_{n}], \Var(y^{o}_{n})) \\
    \text{EC}& =\frac{ \# \{ \by^{*} \in [q^{o}_{0.025}, q^{o}_{0.975}]\}}{N}.  
\end{align*}
where the predicted observations are $\by^{o}$ and the corresponding quantiles are denoted as $q^{o}$. When computing quantiles to obtain empirical coverage and illustrating the uncertainty in \cref{sec:experiments} and below in \cref{sec:supplfig}, we rely on the Gaussian approximation. 

\subsection{Supplementary material to the diabetes example.} \label{sec:supplfig}
\begin{figure}[ht]
    \centering
       \includegraphics[width=0.55\linewidth]{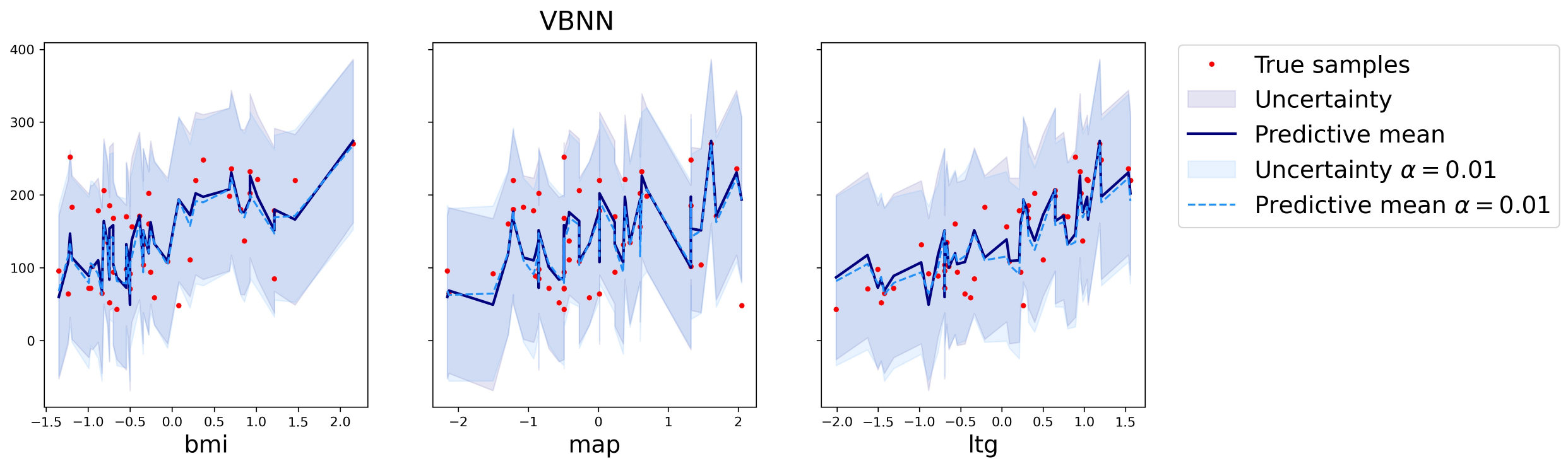}\\
    \includegraphics[width=0.48\linewidth]{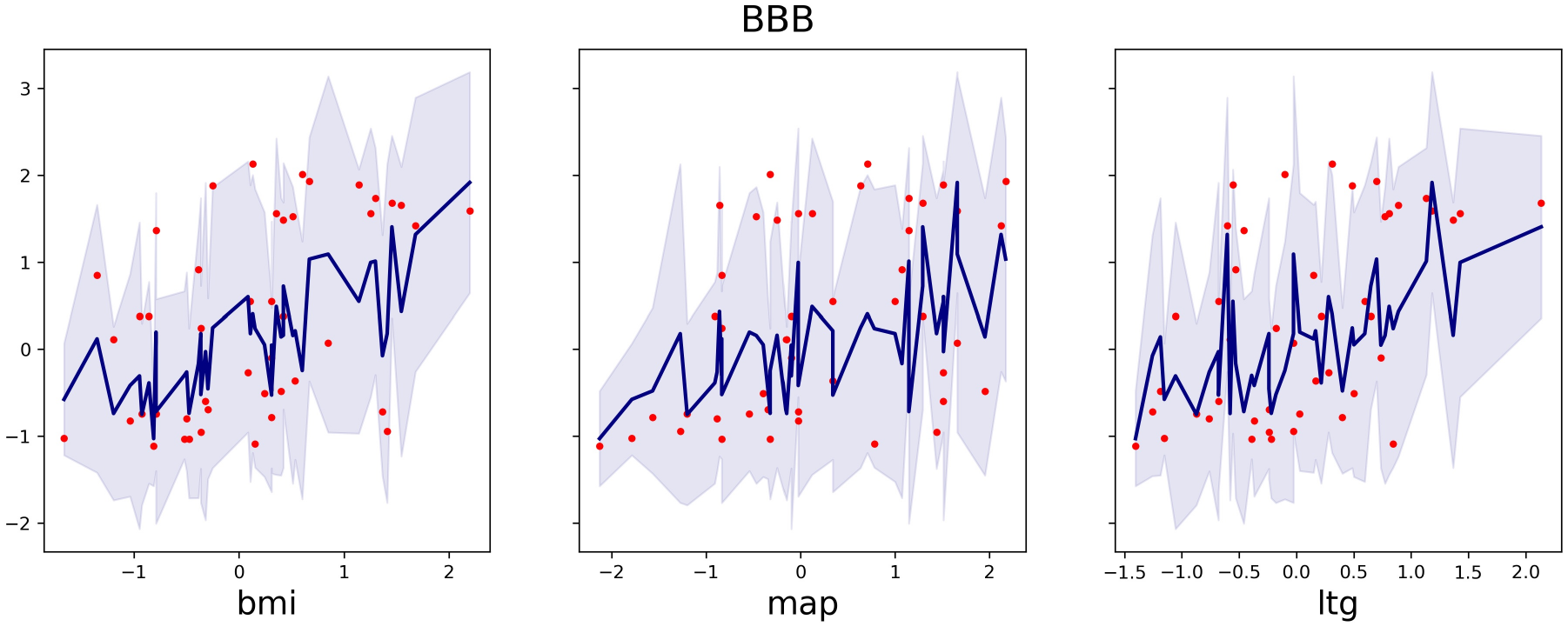}
    \includegraphics[width=0.48\linewidth]{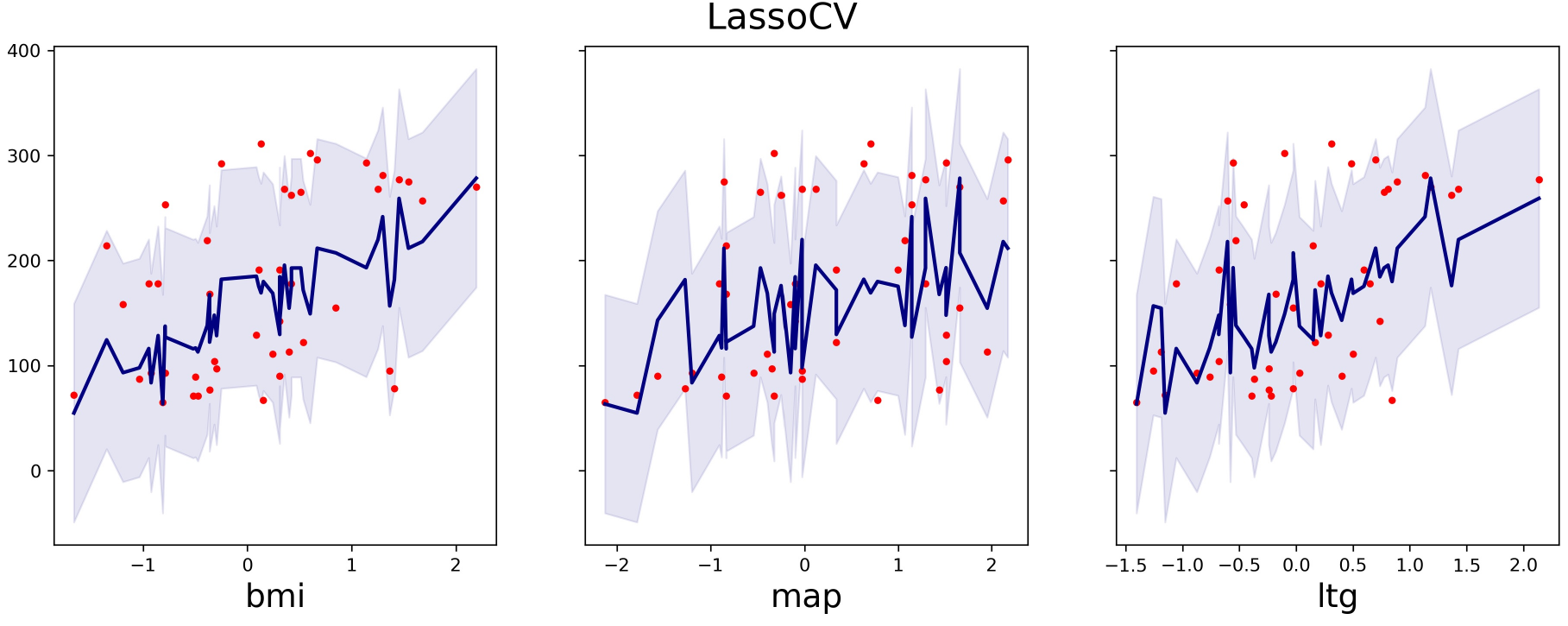}
    \includegraphics[width=0.48\linewidth]{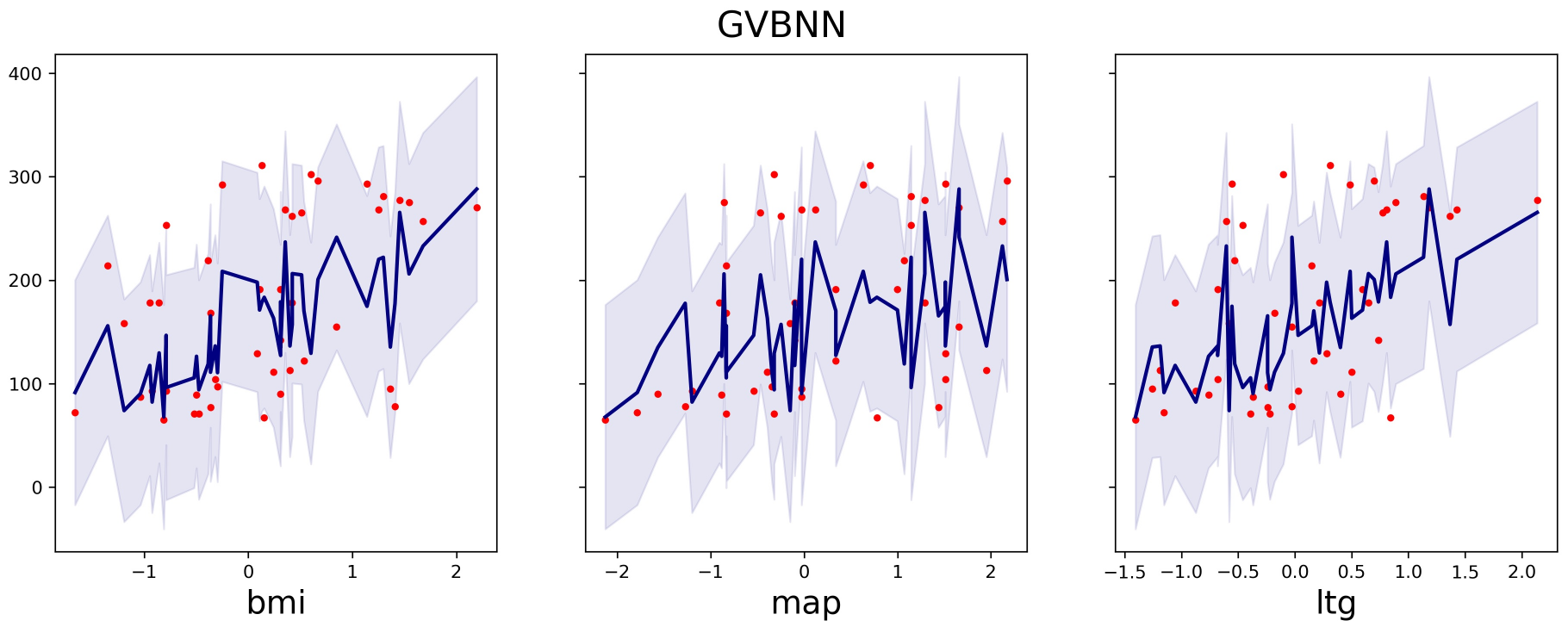}
    \includegraphics[width=0.48\linewidth]{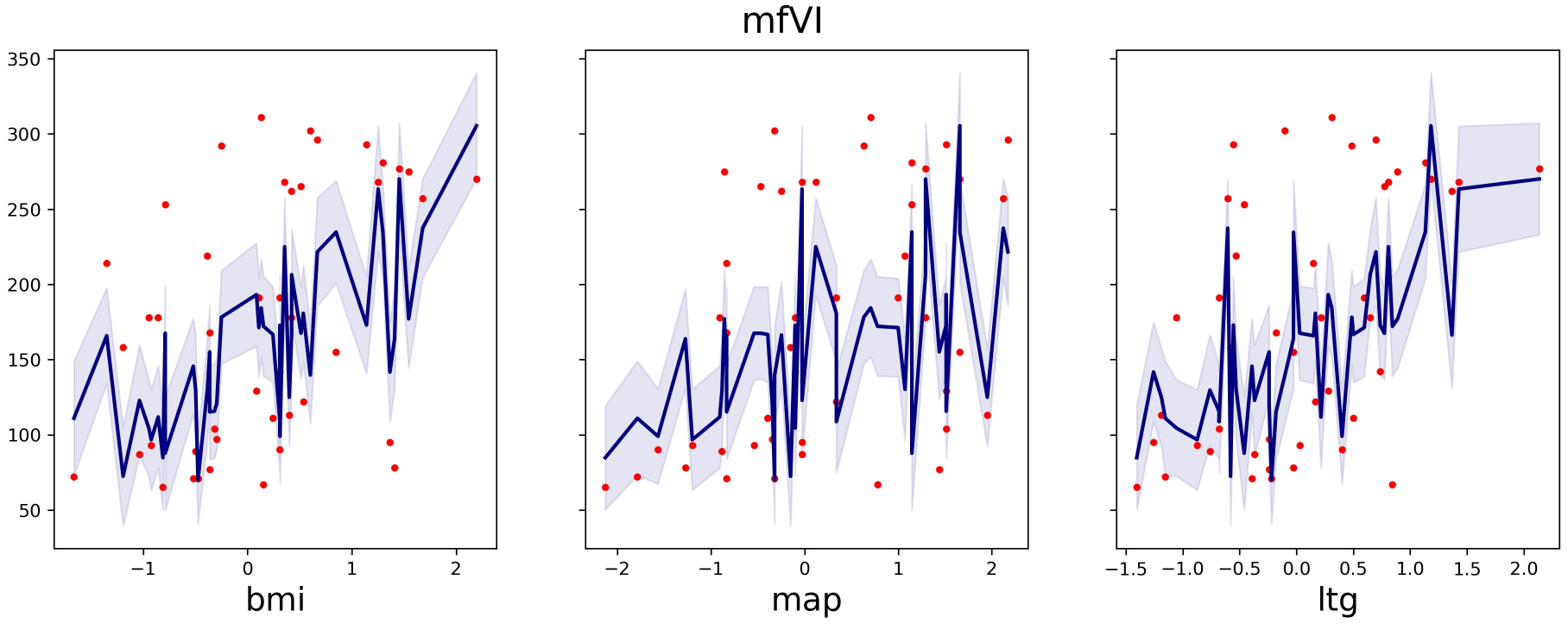}
    \includegraphics[width=0.48\linewidth]{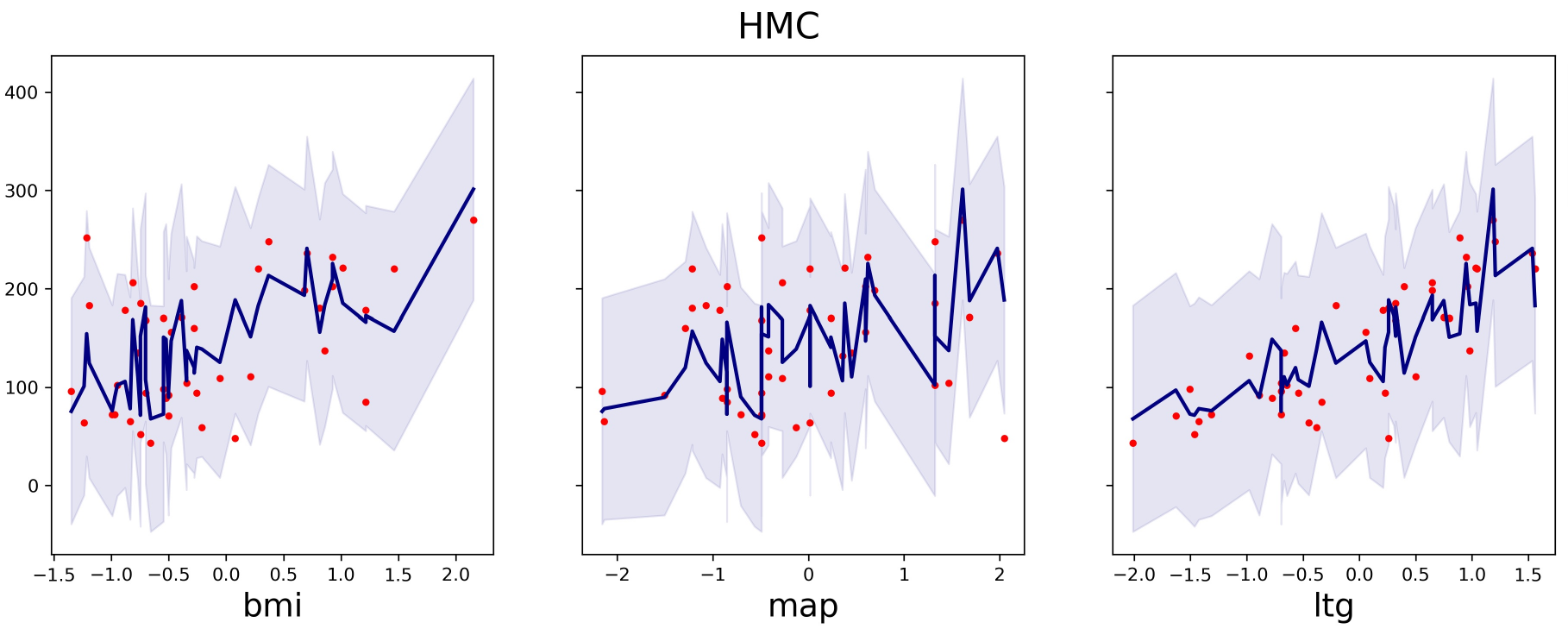}
    \includegraphics[width=0.48\linewidth]{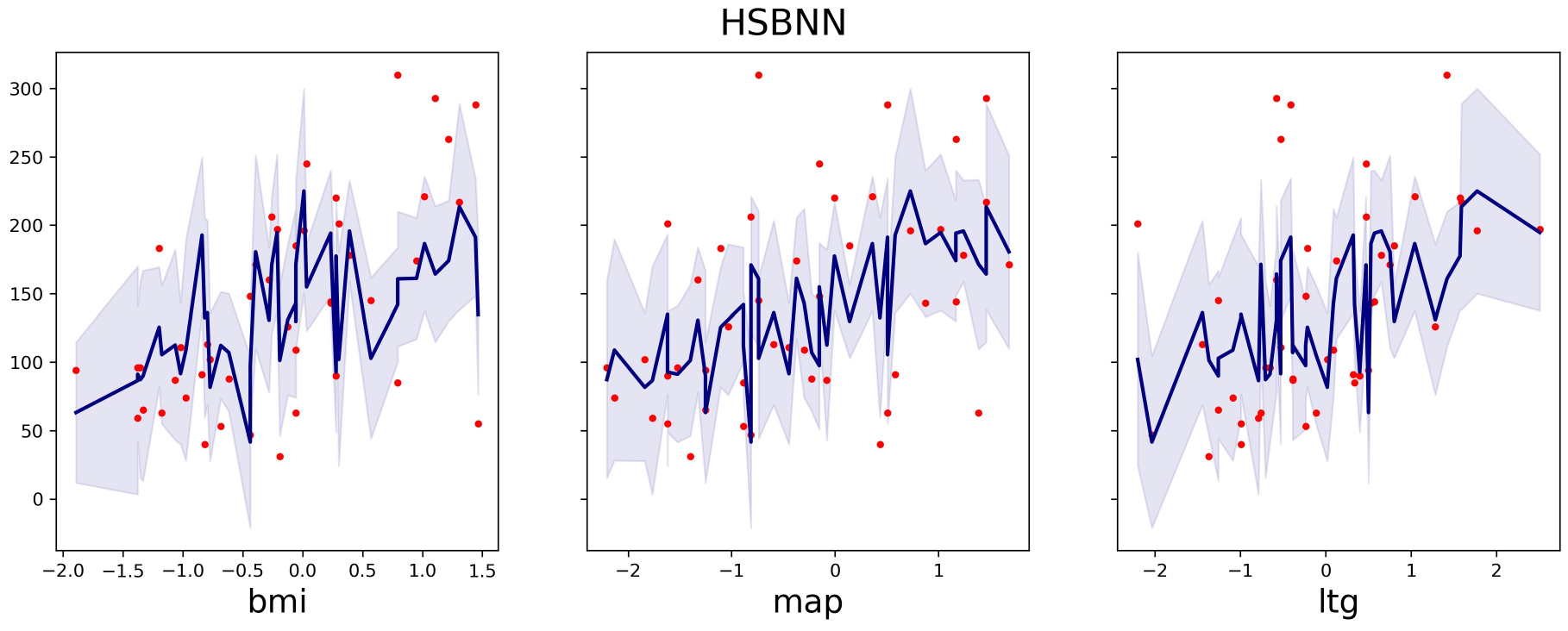}
    \caption{Predictive mean and the uncertainty estimates for the observations for three of the predictors with considerable contribution.}
    \label{fig:maindiabetespredictors}
\end{figure}

\cref{fig:maindiabetespredictors} supplements \cref{tab:diabetespreformance} and the diabetes example in \cref{sec:diabetesex}. Here, in the case of VBNN, BBB and \blu{mfVI} models we provide the uncertainty of the observations and in the case of the LassoCV we provide residual standard deviation.  Additionally, we illustrate the sparse prediction and the uncertainty obtained from sparse weights of the VBNN, which largely coincide with the original prediction and uncertainty estimates. Whilst the coverage estimates for observations of VBNN and BBB are comparable, the \blu{mfVI} underestimates the uncertainty and provides a dramatically lower coverage for observations. 

\subsection{Supplementary information on the datasets}\label{sec:experimentsdata}

\cmsspy{Boston housing} \citep{bostobhousing}: $n =506, p = 13$, the predictors are per capita crime rate by town, the proportion of residential land zoned for lots over 25,000 sq.ft., the proportion of non-retail business acres per town, Charles River dummy variable, nitrite oxides concentration, average number of rooms per dwelling, the proportion of owner-occupied, units built before 1940, weighted distances to five Boston employment centres, index of accessibility to radial highways, full-value property-tax rate, the pupil-teacher ratio by town, the quantitative measure of systemic racism as a factor in house pricing, lower status of the population; the response of interest is the median value of owner-occupied homes. The \cmsspy{Boston housing} dataset is among the most popular pip available datasets, and with respect to variable selection, it was considered in e.g. \citep{schafer2013sequential}. 

\cmsspy{Energy} \citep{misc_energy_efficiency_242}: $n =768, p = 8$, the predictors are
relative compactness, surface area, wall area, roof area, overall height, orientation, glazing area, and glazing area distribution, and the task is to predict the heating load of residential buildings. 

\cmsspy{Yacht dynamics} \citep{misc_yacht_hydrodynamics_243}: $n =308, p = 6$, 
the predictors are long position, prismatic coefficient, length-displacement ratio, bean-draught ratio, length-bean ratio and froude number, and the task is to model the residuary resistance per unit weight of displacement for a yacht hull.

\cmsspy{Concrete compressive strength} \citep{misc_concrete_compressive_strength_165}: $n =1030, p = 8$, the predictors are cement, furnace slag, fly ash, water, superplasticizer, coarse aggregate, fine aggregate and the age of testing, and the response variable is the compressive strength of concrete. This is also considered from the variable selection perspective in several works, including \citep{schafer2013sequential, griffin2024expressing}.

\cmsspy{Concrete slump test} \citep{misc_concrete_slump_test_182}: $n =103, p = 7$, the predictors are concrete ingredients, namely cement, furnace slag, fly ash, water, superplasticizer, coarse aggregate, and fine aggregate, and the task is to predict the slump of concrete.

\subsection{Supplementary material to the UCI datasets experiments} \label{sec:monstertableforUCI}
\cref{tab:metricsUCI} supplements \cref{fig:Monster}, \cref{fig:Slump} and the experiments described in \cref{sec:UCIdatasetsap}. 
\begin{table}[t]
  \caption{\blu{RMSE, NLL and Coverage for UCI datasets.}}
     \label{tab:metricsUCI}
     \centering\blu{
\begin{tabular}{ccccccc}
 &  & \multicolumn{5}{c}{Dataset} \\ \cline{3-7} 
Metric & Method & Slump & Yacht & Boston & Energy & Concrete \\ \hline
\multirow{9}{*}{RMSE} & 4SVIBNN & $7.15\pm 1.5$ & $3.57\pm .8$ & $3.38\pm .9$ & $ 1.62\pm .2$ & $ 7.15 \pm .6 $ \\
& 4VBNN & $7.01\pm 1.2$ & $1.23\pm .3$ & $3.21\pm .6$ & $ 1.1\pm .2$ & $ 6.71 \pm .6 $ \\
 & SVBNN & $7.36\pm 1.62$ & $5.57\pm 1.14$ & $3.76\pm .92$ & $ 2.41\pm .49$ & $ 8.43\pm .75 $ \\
% & e2VBNN & $7.07\pm 1.1$ & $1.56\pm .5$ & $3.28\pm .7$ & $ 1.17\pm .2$ & $ 7.01 \pm .7 $ \\
& VBNN & $7.37\pm1.4 $ & $2.47\pm 1.1 $ & $3.47\pm .8$ & $ 1.37\pm .3$ & $ 7.67 \pm .9 $ \\
& GVBNN & $7.64\pm 1.21$ & $4.88\pm 2.66$ & $4.02\pm .88$ & $ 2.5\pm .42$ & $ 7.84 \pm .68$ \\
 & \blu{mfVI} & $ 7.9 \pm 1.7 $ & $ 1.61\pm .36 $ & $ 3.29 \pm .6 $ & $ 2.27 \pm .25 $ & $ 6.11 \pm .5 $ \\
 & BBB & $7.33\pm 1.94 $ & $1.45\pm .6$ & $3.46\pm .96$ & $2.65\pm .26$ & $6.48\pm .61 $ \\
 & HMC & $ 7.\pm 1.24$ & $.56\pm .13$ & $2.42\pm .47$ & $.3\pm.07 $ & $4.01\pm.78 $ \\ 
 & HSBNN & $ 6.41\pm 1.28$ & $1.2\pm .23$ & $2.92\pm .55$ & $.6\pm.09 $ & $5.21\pm.56$ \\ \hline
\multirow{9}{*}{NLL}
 & 4SVBNN & $3.42\pm 0.2$ & $2.74\pm .2$ & $2.61\pm .2$ & $ 1.94\pm .1$ & $ 3.38  \pm .07 $ \\
 & 4VBNN & $3.39\pm .2$ & $1.97 \pm.2 $ & $2.57 \pm .13$ & $1.63 \pm .16 $ & $ 3.33 \pm .06 $ \\
 & SVBNN & $3.47\pm .29$ & $3.13\pm .27$ & $2.78\pm .31$ & $ 2.31\pm .22$ & $ 3.56\pm 0.11$ \\
% & e2VBNN & $3.4 \pm 0.2 $ & $1.93\pm 0.2 $ & $2.59\pm .1$ & $ 1.67\pm .1$ & $ 3.37 \pm .1 $ \\
& VBNN & $3.46\pm0.2$ & $2.25\pm0.4 $ & $2.69\pm .26$ & $ 1.75\pm .21$ & $ 3.5 \pm .12 $ \\
& GVBNN & $3.47\pm 0.17$ & $2.86\pm .59$ & $2.82\pm .22$ & $ 2.35\pm .15$ & $ 3.48 \pm .09 $ \\
 & \blu{mfVI} & $ 3.77\pm .5 $ & $1.96 \pm .09 $ & $ 2.61\pm .22$ & $ 2.28\pm .09 $ & $ 3.57 \pm .2 $ \\
 & BBB & $6.23\pm 2.76$ & $1.69\pm .14 $ & $2.47\pm .16 $ & $2.08\pm .15$ & $ 3.21\pm 0.13 $ \\
 & HMC & $3.41\pm .24 $ & $.87\pm .1$ & $2.28\pm.18 $ & $.23\pm.44 $ & $2.74\pm.27 $ \\ 
 & HSBNN & $ 5.02\pm 1.79$ & $1.31\pm .18$ & $5.01\pm 1.24$ & $1.08\pm.2 $ & $4.32\pm.69 $ \\ 
\hline
\multirow{9}{*}{Coverage} 
 & 4SVBNN & $.95 \pm .06$ & $.98\pm .03$ & $.97\pm .03$ & $ .98\pm .01$ & $ .97 \pm .01 $ \\
& 4VBNN & $.94\pm .04$ & $ .99 \pm .02$ & $ .97\pm .02$ & $.99\pm .0 $ & $ .97 \pm .02 $ \\
 & SVBNN & $.91 \pm .08$ & $.93\pm .02$ & $.94\pm .03$ & $ .91\pm .04$ & $ .93 \pm .03 $ \\
% & e2VBNN & $.95\pm.04 $ & $.98\pm .02 $ & $.96\pm .02$ & $ .99\pm .01$ & $ .96 \pm .02 $ \\
& VBNN & $.92\pm.06$ & $.96\pm .01$ & $.95\pm .03$ & $ .98\pm .02$ & $ .94 \pm .02$ \\   
& GVBNN & $.96\pm .04$ & $.95\pm .04$ & $.96\pm .02 $ & $ .93\pm .04$ & $.95\pm .02$ \\
 & \blu{mfVI} & $ .78 \pm .1$ & $.96 \pm .03$ & $ .96 \pm .01$ & $.95 \pm .03$ & $ .8 \pm .04$ \\
 & BBB & $.75 \pm .12 $ & $1. \pm .0$ & $.97\pm .02 $ & $.99\pm .0$ & $.97\pm .02 $ \\
 & HMC & $.9\pm .08 $ & $.98\pm.02 $ & $.96\pm .02$ & $.95\pm.03 $ & $.94\pm.03 $ \\
  & HSBNN & $ .67\pm .14$ & $.94\pm .05$ & $0.62\pm .08$ & $.89\pm.04 $ & $.71\pm.06 $ \\ 
\end{tabular}}
\end{table}

\subsection{Supplementary material to the SVI experiments} \label{ap:SVIsupexperiment}
\begin{figure}[t]
    \centering
    \includegraphics[width=0.9\linewidth]{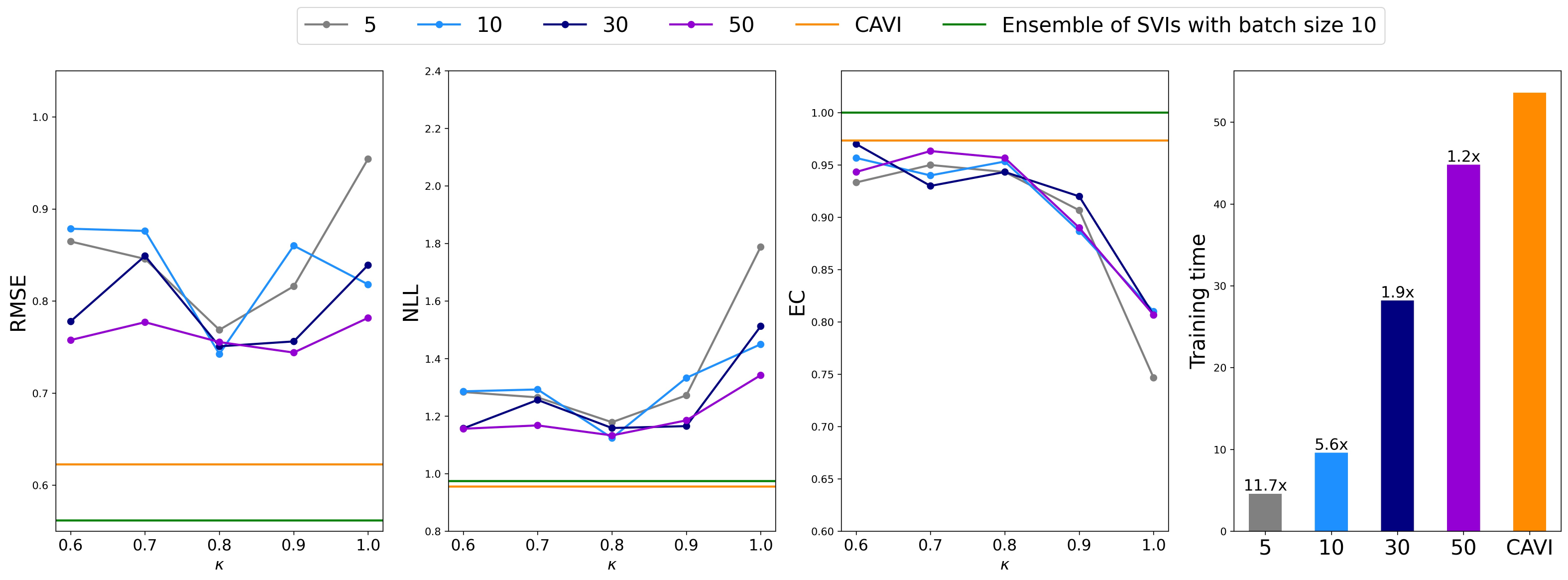}
    \caption{RMSE, NLL, EC of SVI compared to CAVI and plotted for various mini-batch sizes as a function of the forgetting rate $k$; the most right plot compares training times, where bar labels indicate the scale of computational gains.}
    \label{fig:sviexperiment}
\end{figure}
We use a synthetic dataset generated in \cref{sec:toy} to investigate the performance of VBNN trained with stochastic variational inference in comparison to VBNN trained with classical coordinate ascent. We consider a single-layer VBNN with the number of hidden units set to $D_H=20$ and evaluate the performance of the SVI compared to CAVI for various step sizes (defined in \cref{eq:stepsizeforSVI}) and mini-batch sizes. As before, we use 90$\%$ of the data to train the neural network and the rest is used for evaluation; when computing evaluation metrics, we average the results obtained in 10 different random runs. Similar to previous experiments, the recorded metrics are RMSE, NLL and the empirical coverage for the observations. 
\begin{figure}[t]
    \centering
    \includegraphics[width=0.8\linewidth]{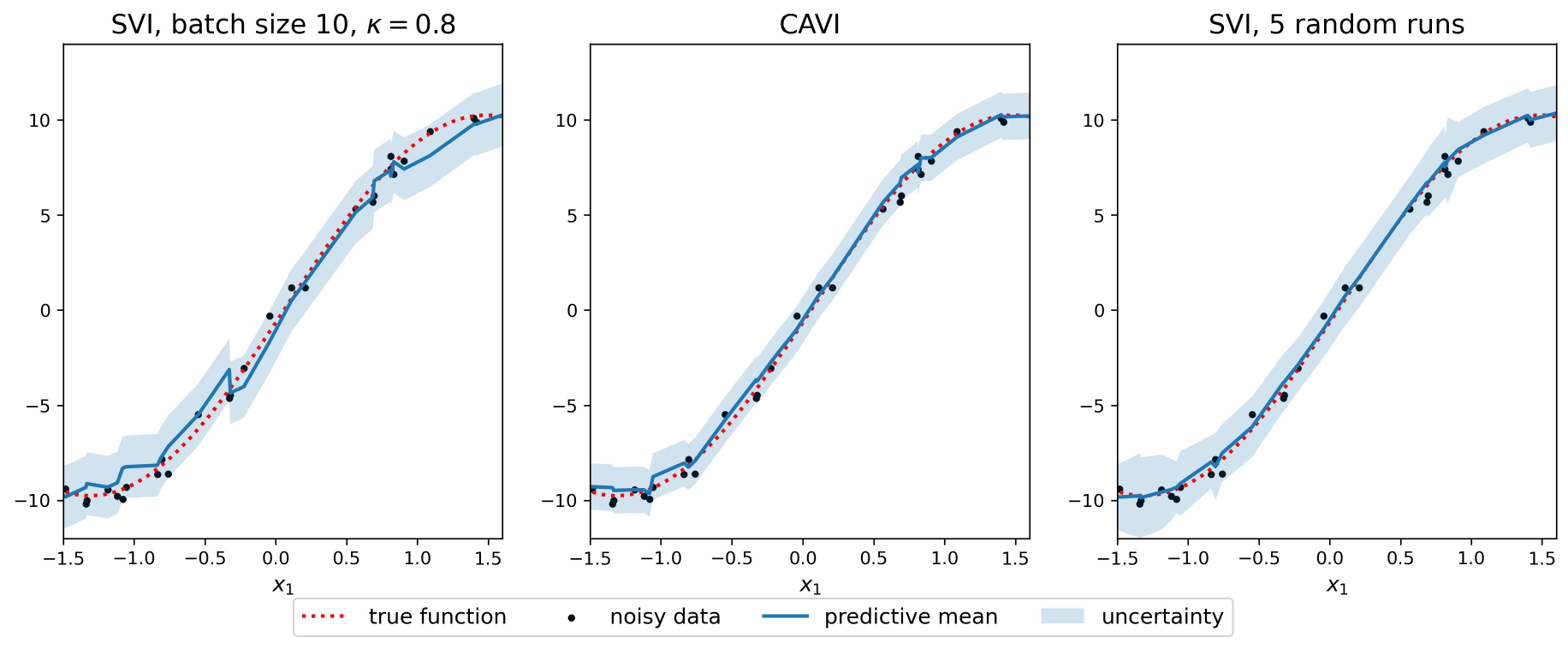}
    \caption{The predictive means and uncertainty estimate as a function of the first coordinate for SVI, CAVI and ensembles of SVI approximations.}
    \label{fig:SVIfunctionexample}
\end{figure}

The computational gains obtained by utilising stochastic gradients and subsampling would motivate further research along the lines of model combination.  Indeed, SVI makes ensembling techniques particularly appealing, in our experiment with a relatively small synthetic dataset, an ensemble of 5 SVI approximations with a mini-batch size of 10 performs comparable to CAVI while being more than 5 times faster than CAVI (see \cref{fig:sviexperiment} and \cref{fig:SVIfunctionexample}).

\section{Review of relevant distributions}\label{appendix:distrib}

\subsection{Generalized Inverse Gaussian}

The Generalized Inverse Gaussian has density:
 \begin{align*}
 p(x \mid \nu, \delta, \lambda ) = \frac{(\lambda/\delta)^\nu}{2K_\nu(\lambda \delta)} x^{\nu-1} \exp \left( -\frac{1}{2} (\delta^2/x + \lambda^2 x) \right),
 \end{align*}
 where $K_\nu()$ is the modified Bessel function of the second kind. 
The GIG prior requires $\nu>0$ if $\delta= 0$ and $\nu<0$ if $\lambda=0$ for a proper prior. Then the expectations arising in computations throughout this paper are: 
  \begin{align*}
  &\E \left [ x\right ] = \frac{\delta K_{\nu+1}(\lambda \delta)}{\lambda K_{\nu}(\lambda \delta)},\\
    &\E \left [ \frac{1}{x}\right ] = \frac{\lambda K_{\nu+1}(\lambda \delta)}{\delta K_{\nu}(\lambda \delta)} - \frac{2\nu}{\delta^2}.
    % & \E \left [ \log x\right ] = \frac{ d K_{\nu}(\lambda \delta)/d\nu}{\delta K_{\nu}(\lambda \delta)} - \frac{1}{2}\frac{\lambda^2}{\delta^2},
 \end{align*}
Often, it is sensible to consider special cases of the GIG, which include:
\begin{enumerate}
    \item Inverse Gamma: when $\lambda= 0$, the GIG reduces to the IG with density:
     \begin{align*}
 p(x \mid \nu, \delta) = \frac{2^\nu}{\delta^{2\nu}\Gamma(-\nu)} (1/x)^{-\nu+1} \exp \left( -\frac{\delta^2}{2x}  \right),
 \end{align*}
 where $\nu <0$ and $\delta>0$. This can also be rewritten in terms of the more standard parametrization of the IG:
   \begin{align*}
 p(x \mid \alpha, \beta) = \frac{\beta^\alpha}{\Gamma(\alpha)} (1/x)^{\alpha+1} \exp \left( -\frac{\beta}{x}  \right),
 \end{align*}
 where $\alpha = - \nu >0$ and $\beta = \delta^2/2>0$. 
 Note that if $w \sim \Norm(0, \tau)$ and $\tau \sim \IG(\alpha, \beta)$,  this implies a marginal student t-prior on $w$ with degrees of freedom $\text{dof}=2 \alpha = -2\nu $ and scale $s = \sqrt{\beta/\alpha} = \delta / \sqrt{-2\nu}$. For example, setting $\nu=-1.5$ would correspond to $\text{dof}=3$ and $\nu=-2.5$ is equivalent to $ \text{dof}=5$.
%  The selection of the scale parameters, $s$ or $\delta$, is critical for the performance of Bayesian shrinkage so that it does not over-shrink or under-shrink. Choosing an excessively large scale parameter weakens the shrinkage effects; thus, it might fail to shrink
% the weights towards 0. On the other hand, choosing a scale parameter that is too small may
% cause many weights to be aggressively shrunk to 0, which might accidentally wipe out the
% effects of the important hidden nodes. In the setting of Bayesian variable selection, \citep{song2017nearly} show nearly optimal contraction rate and selection consistency when setting $s^2 \approx \delta^2 \approx 1/[N \log(p) p ^{-2 \gamma}]$ for sufficiently large $\gamma$, with $N$ and $p$ denoting the sample size and number of predictors/variables, respectively. In their experiments, they fix the degrees of freedom to $3$, i.e. $\nu =-1.5$ and explore $\gamma$ in the range of $-1/4$ to $1.1$. 
%  In this direction, to encourage more shrinkage for larger depth, we could assume  $\delta \propto 1/\sqrt{L}$ for $\tau$ (or more generally  $\delta \propto 1/[\log(L)L^{-2\gamma}]$), and more shrinkage for larger width by setting $\delta \propto 1/\sqrt{D_l D_{l-1}}$ (or more generally  $\delta \propto 1/[\log(D_l D_{l-1})(D_l D_{l-1})^{-2\gamma}]$). 
 
 The relevant expectations for the VI updates and ELBO computation include:
 \begin{align*}
     &\mathbb{E}\left[ x \right] = \frac{\beta}{\alpha -1} = \frac{-\delta^2}{2\nu + 2}, \\
     &\mathbb{E}\left[ \frac{1}{x} \right] = \frac{\alpha}{\beta} = \frac{-2 \nu}{\delta^2}, 
% &\E \left [\log x \right ]= \log \left (\beta \right) - \psi \left(\alpha \right) = \log \left (\frac{\delta^2}{2} \right) - \psi \left(-\nu \right),
 \end{align*}
 where $\psi$ is the logarithmic derivative of the gamma function (a.k.a. digamma function).
 \item Gamma: when $\delta^2= 0$, the GIG reduces to the Gamma with density:
 \begin{align*}
     p(x \mid \nu,\lambda) =\frac{\lambda^{2\nu}}{2^{\nu}\Gamma( \nu )} x^{ \nu -1} \exp(-\frac{\lambda^2}{2} x),
 \end{align*}
 where $\nu > 0$, rewriting in the standard parametrization with $\alpha = \nu$ and $\beta = \lambda^2/2$:
 \begin{align*}
     p(x \mid \alpha,\beta) = \beta^\alpha \frac 1 {\Gamma(\alpha)} x^{\alpha-1} \exp(-\beta x),
 \end{align*}
 where $\alpha = \nu > 0$ and $\beta = \lambda^2/2>0$.
 Similarly,  the relevant expectations are:
 \begin{align*}  
    &\E \left [ x \right ] = \frac{\alpha }{\beta} = \frac{2 \nu}{\lambda^2}, \\
    &\E \left [ \frac{1}{x}\right ] = \frac{\beta}{\alpha - 1} = \frac{\lambda^2}{2 \left (\nu - 1 \right )}. \\
    % & \E \left [ \log x \right ] = \psi(\alpha) - \log(\beta) = \psi (\nu) - \log(\frac{\lambda^2}{2})
 \end{align*}
 Note that if $w \sim \Norm(0, \tau)$ and $\tau \sim \Gam(1, \beta)$,  this implies a marginal Laplace prior on $w$ (i.e. Bayesian Lasso \citep{parkcasellalasso})  with  scale $s = 1/\sqrt{2\beta} = 1/\lambda$.
 
% \begin{figure}
%     \centering
%     \includegraphics[scale = 0.3]{pictures/gammamix.pdf}
%     \includegraphics[scale = 0.3]{pictures/gammamixdifflayers.pdf}
%     \caption{Density estimates for the probability distribution of weights when priors are Gamma.}
%     \label{fig:gammapriors}
%     \Description[Global-local shrinkage priors have substantial mass near zero and heavy tails]{ Two-dimensional contour plots in the case of Gamma prior look similar to a rectangle.}
% \end{figure}

\item Inverse Gaussian (IGaus): when $\nu=-1/2$, the GIG reduces to the Inverse Gaussian with density: 
   \begin{align*}
     p(x \mid \delta,\lambda) =  \frac{\delta}{\sqrt{2 \pi x^3}} \exp\left( -\frac{( \lambda x-\delta)^2}{2x}\right ),
 \end{align*}
 where setting $\alpha=\delta /\lambda>0$ and $\beta = \delta^2 >0$ we derive
   \begin{align*}
     p(x \mid \alpha,\beta) =  \left(\frac{\beta}{2 \pi x^3}\right )^{\frac{1}{2}} \exp\left( \frac{-\beta (x-\alpha)^2}{2 \alpha^2 x}\right ).
 \end{align*}
  The relevant expectations for the VI updates and ELBO computation include:
 \begin{align*}
 &  \E \left [ x \right ] = \alpha= \frac{\delta}{\lambda}, \\
   &  \E \left [ \frac{1}{x}\right ] = \frac{1}{\alpha} + \frac{1}{\beta} = \frac{\lambda}{\delta} + \frac{1}{\delta^2}.
 \end{align*}
 Note that if $w \sim \Norm(0, \tau)$ and $\tau \sim \text{IGaus}(\alpha, \beta)$,  the marginal distribution is of the form \citep{caronsparse}:
\begin{align*}
           p(w_k) &= \frac{1}{\pi \alpha}\left( \frac{\beta}{\beta + w_k^2}\right)^{\frac{1}{2}} \exp \left( \frac{\beta^{\frac{1}{2}}}{\alpha}\right) K_1 \left( \frac{(\beta + w_k^2)^{\frac{1}{2}}}{\alpha}\right)  \\
           & = \frac{\lambda}{\pi} \exp(\lambda) \left( \delta^2 + w_k^2 \right)^{-\frac{1}{2}} K_1\left ( \frac{\lambda}{\delta}\left( \delta^2 + w_k^2\right)^{\frac{1}{2}}\right).
\end{align*}

% \begin{figure}
%     \centering
%     \includegraphics[scale = 0.3]{pictures/igaussmixsame.pdf}
%     \includegraphics[scale = 0.3]{pictures/igauusmix.pdf}
%     \caption{Density estimates for the probability distribution of weights when priors are Inverse Gaussian.}
%     \label{fig:igausspriors}
%     \Description[Global-local shrinkage priors have substantial mass near zero and heavy tails]{ Two-dimensional contour plots in the case of Inverse Gaussian prior look similar to an ellipse.}
% \end{figure}
\end{enumerate}
\subsection{EM update for different cases of global-local priors} \label{sec:emappendix}
As discussed above, the special cases of the GIG include Inverse Gamma, Gamma and Inverse Gaussian distributions, we derive the EM updates in each of the special cases of priors:
\begin{enumerate}
\item Inverse Gamma: when the global shrinkage parameter has an Inverse Gamma distribution, then 
\begin{align*}
\delta_{\glob} & = \argmax\left(\delta_{\glob}^2\sum_{l=1}^{L+1}\frac{ \nu_{\glob, l}}{\delta_{\glob, l}^2}  - 2(L+1)\nu_{\glob}\log(\delta_{\glob}) \right), \\
  \delta_{\glob} & = \left( (L+1) \nu_{\glob} \right)^{\frac{1}{2}} \left(\sum_{l=1}^{L+1}\frac{ \nu_{\glob, l}}{\delta_{\glob, l}^2}\right)^{ - \frac{1}{2}}. 
\end{align*}
\item Gamma: similarly, when global shrinkage parameter is Gamma:
\begin{align*}
    \lambda_{\glob} & =\argmax \left(4(L+1) \nu_{\glob} \log(\lambda_{\glob}) - \lambda_{\glob}^2 \sum_{l=1}^{L+1} \frac{\delta_{\glob,l} K_{\nu_{\glob,l}+1}(\lambda_{\glob,l}\delta_{\glob,l})}{\lambda_{\glob,l} K_{\nu_{\glob,l}}(\lambda_{\glob,l} \delta_{\glob,l})}\right), \\
    \lambda_{\glob} & = \left( 2(L+1) \nu_{\glob}\right)^{\frac{1}{2}} \left(\sum_{l=1}^{L+1} \frac{\delta_{\glob,l} K_{\nu_{\glob,l}+1}(\lambda_{\glob,l}\delta_{\glob,l})}{\lambda_{\glob,l} K_{\nu_{\glob,l}}(\lambda_{\glob,l} \delta_{\glob,l})}\right)^{-\frac{1}{2}}. 
\end{align*}
\item Inverse Gaussian:  if the global shrinkage parameter is Inverse Gaussian, then
\begin{align*}
\lambda_{\glob} & = \argmax\left( 2(L+1)\lambda_{\glob}\delta_{\glob}
 - \lambda_{\glob}^2\sum_{l=1}^{L+1} 
\frac{\delta_{\glob,l} K_{\nu_{\glob,l}+1}(\lambda_{\glob,l}\delta_{\glob,l})}{ \nu_{\glob, l}K_{\nu_{\glob,l}}(\lambda_{\glob,l} \delta_{\glob,l})}  \right), \\
\lambda_{\glob} & = 2(L+1)\delta_{\glob}\left(\sum_{l=1}^{L+1} 
\frac{\delta_{\glob,l} K_{\nu_{\glob,l}+1}(\lambda_{\glob,l}\delta_{\glob,l})}{ \nu_{\glob, l}K_{\nu_{\glob,l}}(\lambda_{\glob,l} \delta_{\glob,l})}\right)^{-1}.
\end{align*}
\end{enumerate}

\end{document}